\documentclass[preprint,authoryear,3p]{elsarticle}

\usepackage{booktabs}

\usepackage{amssymb,amsthm,amsmath, amsfonts}
\usepackage{graphics}
\usepackage{graphicx}
\usepackage{float}
\usepackage{multirow}
\usepackage{longtable,lscape}
\usepackage{cases}
\usepackage{hyperref}
\usepackage{url}
\usepackage{caption}
\usepackage{subcaption}
\usepackage{color}
\usepackage{algpseudocode,algorithm,algorithmicx}
\usepackage[pagewise]{lineno}
\usepackage{mathtools}

\usepackage{tikz}
\usetikzlibrary{arrows.meta,positioning,calc}
\definecolor{ourblue}{RGB}{63,105,170} % fallback if not defined

\usepackage{pgfplots}
\pgfplotsset{compat=1.18}
\usetikzlibrary{positioning}
\definecolor{ourblue}{RGB}{60,100,220}

\usepackage{fancyhdr}
\usepackage{nomencl}
\makenomenclature

\usepackage{etoolbox}
\renewcommand\nomgroup[1]{%
 \item[\bfseries
  \ifstrequal{#1}{A}{Reinforcement Learning}{%
  \ifstrequal{#1}{B}{Traffic Signal Control}{%
  \ifstrequal{#1}{C}{Other Notations}{}}}%
]}

\usepackage{multirow}
\usepackage{multicol}
\usepackage{comment}
\usepackage{graphicx}
\usepackage{soul}
\usepackage{pdflscape}
\usepackage{booktabs,caption}
\usepackage{threeparttable} %[flushleft]
\usepackage{arydshln}
\usepackage{enumerate}
\usepackage{tabularx,makecell,array}
\setcellgapes{4pt}%parameter for the spacing
\usepackage[export]{adjustbox}
\usepackage{fullpage}

\usepackage{rotating}

\theoremstyle {plain}% default

\newcounter{x}

\theoremstyle{definition}

\theoremstyle{remark}

\usepackage{soul}
\usepackage{titlesec}
\titleformat{\paragraph}
{\normalfont\normalsize\bfseries}{\theparagraph}{1em}{}
\titlespacing*{\paragraph}
{0pt}{3.25ex plus 1ex minus .2ex}{1.5ex plus .2ex}

\usepackage{color}

\makeatletter
\def\ps@pprintTitle{%
	\let\@oddhead\@empty
	\let\@evenhead\@empty
	\def\@oddfoot{\reset@font\hfil\thepage\hfil}
	\let\@evenfoot\@oddfoot
}
\makeatother

\numberwithin{equation}{section}
\usepackage{enumitem}

\newcolumntype{Y}{>{\raggedright\arraybackslash}X}
\newcolumntype{C}{>{\centering\arraybackslash}X}

\usepackage{booktabs,tabularx,array}
\newcolumntype{L}{>{\raggedright\arraybackslash}X}
\newcolumntype{S}[1]{>{\raggedright\arraybackslash}p{#1}}

\usepackage{xcolor} % if not already loaded
\newcommand{\mname}[1]{\textbf{\textsf{#1}}} % method/paper short-name styling
\newcommand{\mcite}[1]{{\color{gray!60}\scriptsize\citep{#1}}} % subdued, small citation

\biboptions{authoryear}                 % enables \citep / \citet
\journal{Transportation Research Part C Special Issue ``Foundation Models and Large Language Models in Urban Mobility"}

\begin{document}

\begin{frontmatter} 
		
		
		
\title{
Do Math‑Reasoning LLMs Help Predict the Impact of Public Transit Events?
}
		
\date{\today}
		
		
\author[cu]{Bowen Fang}
\author[cu-cs]{Ruijian Zha}
\author[cu,dsi]{Xuan Di\corref{cor}}
% \ead{sharon.di@columbia.edu}

\cortext[cor]{Corresponding author. Tel.: +1 212 853 0435;}
\address[cu]{Department of Civil Engineering and Engineering Mechanics, Columbia University}
\address[cu-cs]{Department of Computer Science, Columbia University}
\address[dsi]{Data Science Institute, Columbia University}

\begin{abstract}
Predicting public transit incident duration from unstructured text alerts is a critical but challenging task. Addressing the domain sparsity of transit operations with standard Supervised Fine-Tuning (SFT) is difficult, as the task involves noisy, continuous labels and lacks reliable expert demonstrations for reasoning. While Reinforcement Learning from Verifiable Rewards (RLVR) excels at tasks with binary correctness, like mathematics, its applicability to noisy, continuous forecasting is an open question. This work, to our knowledge, is the first to bridge the gap between RLVR LLM training with the critical, real-world forecasting challenges in public transit operations. We adapt RLVR to this task by introducing a tolerance-based, shaped reward function that grants partial credit within a continuous error margin, rather than demanding a single correct answer. We systematically evaluate this framework on a curated dataset of NYC MTA service alerts. Our findings show that general-purpose, instruction-tuned LLMs significantly outperform specialized math-reasoning models, which struggle with the ambiguous, real-world text. We empirically demonstrate that the binary reward is unstable and degrades performance, whereas our shaped reward design is critical and allows our model to dominate on the most challenging metrics. While classical regressors are superior at minimizing overall MAE or MSE, our RLVR approach achieved a 35\% relative improvement in 5-minute accuracy (Acc@5) over the strongest baseline. This demonstrates that RLVR can be successfully adapted to real-world, noisy forecasting, but requires a verifier design that reflects the continuous nature of the problem.


\end{abstract}
		
\end{frontmatter}

\section{Introduction}

% \sd{Major issues:
% 1. read every single sentence and make sure it is meaningful, not some vague, high-level, non-human like nonsense written by chatGPT.\\
% 2. Too many sub(sub)sections which makes the organizations and paper structure unclear. Need to re-orgnize and combine some subsections. 
% Readers should know what messages each sub(sub)section conveys by reading the captions, how each sub(sub)section is related to one another and to the high-level section.\\
% 3. Do not include "takeaways" in incomplete sentences. You are writing an academic paper not making slides. All the sentences should be complete and formal, not in informal formats. 
% If you insist on including them, write complete sentences in a professional way at least. Not something that nobody understands with only phrases and words (which is normally explained during a presentation). This paper will be read by humans not machines.}

\subsection{Motivation}
\begin{figure*}[t]
\centering
\includegraphics[width=1\linewidth]{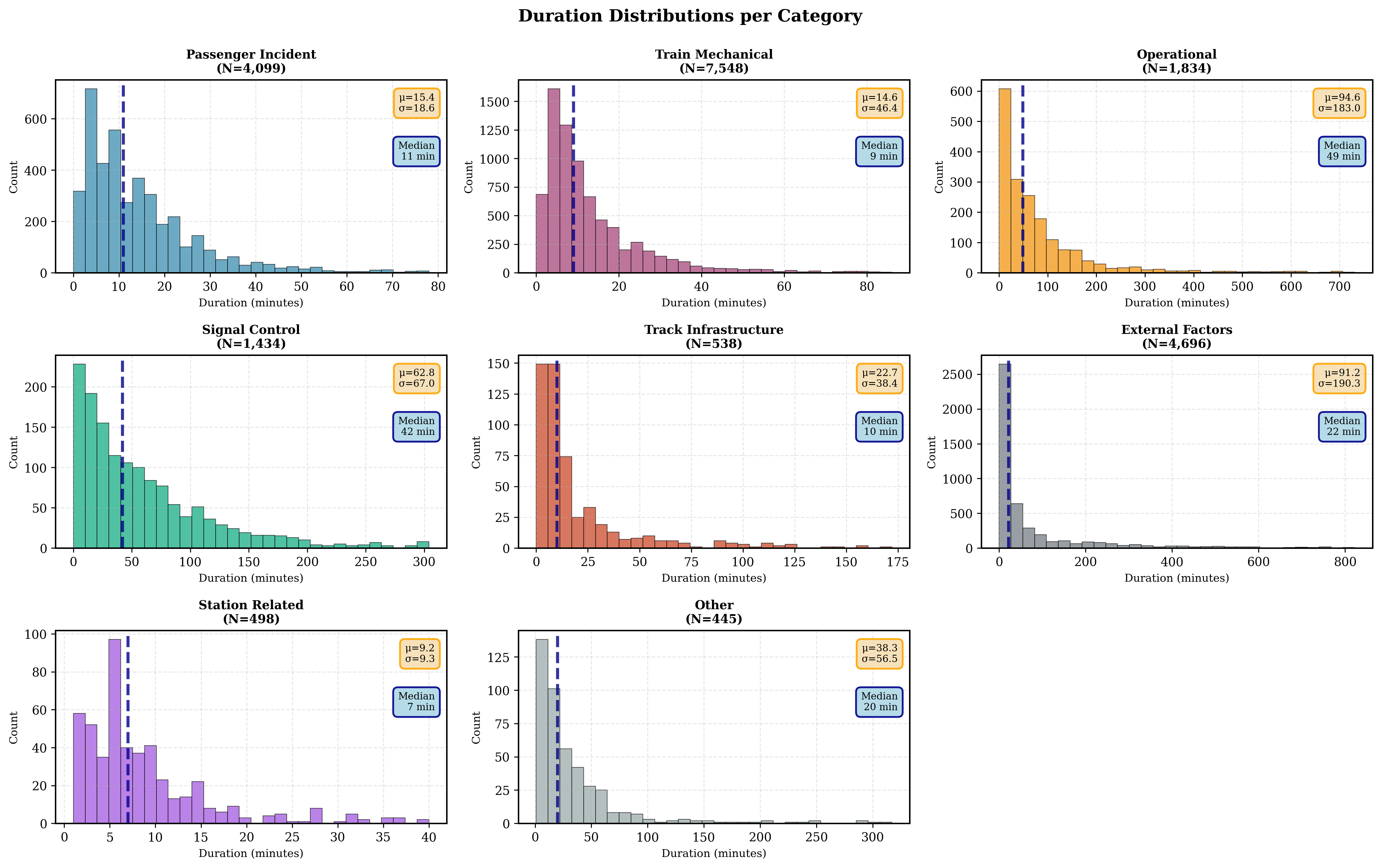}
\caption{\textbf{Transit disruption duration distributions per category.} Duration extracted from NYC MTA service alerts starting from April 28, 2020. Categories are extracted from GTFS-rt alerts using LLM, detailed in \S\ref{sec:eda}.
Histograms reveal distinct patterns based on incident type. Passenger incidents and train mechanical issues are heavily right-skewed, with most resolving quickly (modes at 5–15 min). In contrast, external factors and operational events display broader distributions with longer tails, while operational incidents show the highest variance. Blue dashed lines indicate the median duration for each category.} \label{fig:duration_histograms}
\end{figure*}

Public transit disruptions are a common and costly part of urban life, yet their impact, especially the \emph{incident duration}, remains difficult to predict from the earliest signals. Without reliable, real-time duration estimates, transit agencies and riders have to respond reactively, leading to inefficient resource allocation and frustrating travel experiences. While the progression of any single event appears complex, clear patterns exist across large datasets. In subway systems, for example, incidents caused by signal control often last substantially longer than those by passengers. Our data show a median duration of 42 minutes for signal control versus just 11 minutes for passenger incident (see Figure~\ref{fig:duration_histograms}. The passenger incident is in the first row first column, while signal control is in the second row first column). 
Such critical information revealing root cause, geographic locations, affected lines, and provisional mitigation, frequently appears first in free-form, unstructured service alert texts. 

This textual data is distributed through the General Transit Feed Specification (GTFS), an open standard for public transit information adopted by more than 10,000 public agencies from over 100 countries. It consists of two types of formats, namely, static GTFS schedule and up-to-date real time information (i.e., GTFS-rt). While GTFS offers standard timetables of each route, GTFS-rt provides live arrival time to each station/stop along the route. In addition, GTFS-rt service alerts contain rich information about the cause and the duration of the delay, despite that they are underexploited due to its inconsistent formats. As such, the text alerts within GTFS-rt, though often underexploited due to inconsistent formats, hold significant, untapped potential for understanding and modeling transit system performance. 

Interpreting and forecasting the duration of a disruption from text alerts poses challenges. When an initial alert does specify an end time, it typically represents a plan or an estimate, not the realized duration. Take New York City GTFS-rt for example, on March~23,~2023, a 34th~St roadwork detour was announced at 1{:}08~AM and scheduled to last “until 5~AM”. The actual all-clear message, however, was not issued until 9{:}23~AM. Our forecasting target is this \emph{realized duration}, and we aim to predict it as early as possible from the initial alerts. 
This textual data stream presents a natural opportunity for Large Language Models (LLMs), which excel at parsing and interpreting natural language. 

We would like to note that, issuing text alerts precede our model prediction, meaning that we do not intend to make prediction when an event happens, but when a text alert associated with that event is issued. 
We aim to make accurate predictions at early stages. 
Such forecast is meaningful for various downstream applications: operation teams can better dispatch alternative buses and optimize crew shifts; planning teams can quantify these gaps to improve future predictions; and trip planners can provide realistic ETAs and suggest alternative routes.

\subsection{Challenges}

Forecasting durations of transit disruptions leveraging LLM present challenges, in data representation, data quality, and methodological transfer.

First, the \textbf{domain sparsity} of transit operations is a critical hurdle. The specialized terminology and operational context in GTFS-rt alerts (e.g., “signal malfunction,” “switch problem,” “unauthorized person on the tracks”) are underrepresented in the generic corpora used to pre-train LLMs. This limits zero-shot performance and necessitates domain adaptation. 
Further, the unstructured nature of the alerts requires robust parsing, event linking, and temporal alignment before learning can even begin.

The most straightforward solution to domain sparsity is \textbf{Supervised Fine-Tuning (SFT)} on a domain-specific dataset. This process takes a large, general-purpose pre-trained model, and further trains 
the model % \sd{it (what does "it" refer to?)} 
on a smaller, curated dataset of high-quality examples from the target domain~\citep{ouyang2022training, wei2021finetuned}
% \sd{[Ref]}
. These examples are typically formatted as instruction-response pairs (e.g., a specific prompt and its ideal answer). During SFT, the model's weights are updated to predict the ``correct" response for each prompt, effectively teaching 
the model% \sd{it (what does "it" refer to?)} 
the specific knowledge, style, and formatting of the new domain.
However, this leads to the second major challenge: noisy and uncertain labels. The ``ground truth" realized durations are themselves heuristics, derived from edited and 
processed % \sd{deduplicated (is this a real word?)} 
alert sequences with ambiguous timestamps. An SFT model trained to minimize loss against these targets may learn to precisely replicate the noise in data, rather than learning the underlying patterns of the incident itself. The model must learn under significant uncertainty in both inputs and targets.

This limitation of SFT motivates exploring more recent reward-based training paradigms. Recent advances in verifier-driven training, particularly \textbf{Reinforcement Learning from Verifiable Rewards (RLVR)}, have substantially improved performance on domains with near-binary correctness (e.g., mathematics and programming), where a model’s response is rewarded when 
the response % \sd{it (what does "it" refer to?)} 
passes a deterministic checker. This success raises our central research question: \emph{Could the mathematical and logical reasoning capabilities that help LLMs solve verifiable tasks be transferred to the noisy, uncertain, and continuous task of incident duration prediction?} 

Transferring RLVR from math and coding to transit forecasting is non-trivial. The target is a continuous duration, not a binary correct answer. An exact-match verifier does not apply; multiple predictions may be acceptable if they are sufficiently close to the realized duration. Therefore, a successful adaptation requires shifting the training and evaluation from valuing exactness to closeness.

\subsection{Contributions}

To the best of our knowledge, this work is the first to bridge RLVR LLM training with the critical, real-world forecasting challenges in public transit operations. We introduce a data-driven framework for alert-driven incident-duration prediction. 
Our methodological contributions include:

\textbf{(1) Dataset.} We curate and release a high-quality dataset linking GTFS-rt service alerts to realized incident duration. This includes robust data-processing pipelines for event deduplication, temporal alignment of alert sequences, and standardized splits suitable for LLM fine-tuning and benchmarking LLMs.\footnote{Publicly available at \href{https://huggingface.co/datasets/bf2504/transit-incident-duration}{Hugging Face Datasets (bf2504/transit-incident-duration)}.}

% \sd{Do you have a github link to this dataset to show that you are sincere to release the dataset?}

\textbf{(2) RLVR adaptation to continuous, noisy targets.} We propose a \emph{tolerance-based, shaped reward} to adapt verifier-driven methods to this forecasting task. Instead of requiring an exact match, our approach grants partial or full credit to predictions within a predefined error margin, which stabilizes training and encourages the LLM to learn accurate, well-calibrated duration distributions. 

\textbf{(3) A systematic study of LLMs for incident duration forecasting.} We benchmark models with varying math-reasoning strengths, test prompting strategies that provide different levels of domain knowledge, and compare RLVR algorithms with clear ablations of our reward design. We also compare against a family of non-LLM regressors. Across all experiments, our findings demonstrate that strong reasoning ability alone is insufficient; high performance is only achieved by grounding the model in domain knowledge and applying task-oriented RLVR.

% =========================
% Table 1 (top row + caption fixed)
% =========================
\begin{table*}[t]
\centering
\footnotesize
\setlength{\tabcolsep}{5pt}
\renewcommand{\arraystretch}{1.1}
\caption{Representative transportation-focused foundation-model works, positioned by problem and capability. “Foundation-model” includes LLMs, VLMs, agent frameworks, and traffic foundation models.}
\begin{tabularx}{\textwidth}{S{0.20\textwidth} L L}
\toprule
\textbf{Reference} & \textbf{Transport Problem} & \textbf{Foundational Model Perspectives} \\
\midrule
\textbf{Ours} & Early \emph{incident duration} from alert text & \textbf{RLVR} (verifier-shaped tolerance reward); reasoning \\

TRIP \citep{liu2025trip} & ITS via dual (semantic/physical) state spaces; hierarchical decision & hierarchical RL; cross-modal alignment \\
TR-Agent \citep{guo2025automating} & Automating refinement of traffic models & agentic operations \\
Traffic-IT \citep{kuang2025traffic} & Traffic scene understanding & VLM VQA \\
TrafficGPT \citep{zhang2024trafficgpt} & Urban traffic management assistant & LLM tool-use; interactive dialogue \\
TransitGPT \citep{devunuri2025transitgpt} & GTFS analytics & LLM prompt-based code generation; tool-use \\
TraveLLM \citep{fang2024travellm} & Disruption-aware public transit routing from multimodal user queries & LLM prompting-based reasoning \\
DelayPTC-LLM \citep{chen2024delayptc} & Passenger travel choice under metro delays & LLM prompting with rationalized predictions; few-/zero-shot \\
Time-LLM \citep{jin2023time} & Generic time-series forecasting & LLM Prompt-as-Prefix (PaP) \\
News$\rightarrow$Forecast \citep{wang2024news} & Event-aware forecasting across news + time series & agentic operations; SFT \\
ST-LLM \citep{liu2024spatial} & Traffic prediction (spatio-temporal) & LLM with spatio-temporal tokenization \\
\bottomrule
\end{tabularx}

\label{tab:rw-summary-transport-fm}
\end{table*}

% =========================
% Figure 2 (with minor label fix)
% =========================
\begin{figure}[t]
\centering
\begin{tikzpicture}
\begin{axis}[
  width=0.9\linewidth,
  height=12.0cm,
  xmin=-0.5, xmax=3.5, ymin=-0.5, ymax=4.5,
  axis lines=left,
  xtick={0,1,2,3},
  xticklabels={non-LLM/VLM, Prompting, SFT, RLVR},
  ytick={0,1,2,3},
  yticklabels={Interfaces, Agentic ops, Urban ST, Incident duration}, % <— was "Incident Dur."
  grid=both, minor grid style={gray!15}, major grid style={gray!25},
  tick label style={font=\scriptsize},
  label style={font=\footnotesize},
  clip=true
]

% compact node styles to prevent overflow
\tikzset{
  bubble/.style={draw,rounded corners=2pt,fill=gray!10,align=left,inner sep=2.5pt,text width=2.7cm,font=\footnotesize},
  ourbox/.style={draw=ourblue,very thick,rounded corners=2pt,fill=ourblue!10,align=left,inner sep=3pt,text width=3cm,font=\footnotesize}
}

% ---- Nodes (x = distillation stage, y = task band) ----

% y=0 Interfaces (Prompting)
\node[bubble] at (axis cs:1,0.1) {
  \scriptsize \mname{TrafficGPT}~\mcite{zhang2024trafficgpt};\\
  \scriptsize \mname{TransitGPT}~\mcite{devunuri2025transitgpt};\\
  \scriptsize \mname{TraveLLM}~\mcite{fang2024travellm}
};

% y=1 Agentic ops (Prompting)
\node[bubble] at (axis cs:1,1) {
  \scriptsize \mname{TR-Agent}~\mcite{guo2025automating}
};

% y=2 Urban ST (SFT etc.)
\node[bubble] at (axis cs:2,2) {
  \scriptsize \mname{Time-LLM}~\mcite{jin2023time};\; \\
  \mname{ST-LLM}~\mcite{liu2024spatial};\; \\
  \mname{News$\to$Forecast}~\mcite{wang2024news}
};

% y=2 Urban ST (Prompting) — mobility foundation model
\node[bubble] at (axis cs:1,2) {
  \scriptsize \mname{MobiFuse}~\mcite{ma5218098learning};\; \\
  \mname{UniST}~\mcite{yuan2024unist}
};

% y=3 Incident duration (non-LLM baselines at left)
\node[bubble] at (axis cs:0,3) {
  \scriptsize \mname{Hazard-Based Model}~\mcite{kalair2021dynamic};\\
  \scriptsize \mname{Probabilistic Topic Model}~\mcite{zhao2024predicting};\\
  \scriptsize \mname{Passenger Trajectories}~\mcite{krishnakumari2020estimation};\\
  \scriptsize \mname{Disruption Exposure/Impact}~\mcite{yap2021predicting}
};

% ---- OUR PAPER — TOP ROW (RLVR column) ----
\node[ourbox,anchor=north east] at (axis cs:3.5,3.5) {\textbf{Ours}\\[-1pt]
  \scriptsize \mname{RLVR} of math reasoning LLMs (verifier-shaped distribution)};

\end{axis}
\end{tikzpicture}

\caption{%
\textbf{Map of related work in LLMs for transit and urban tasks.} The figure positions our research relative to existing literature along two primary axes.
\textbf{Horizontal Axis (Training/Adaptation Method):} This axis represents the method used to adapt the LLM to domain knowledge, progressing from non-LLM/VLM baselines and inference-time Prompting, to SFT (Supervised Fine-Tuning) on demonstrations, and finally to RLVR (Reinforcement Learning from Verifiable Rewards).
\textbf{Vertical Axis (Task Family):} This axis categorizes the application domain, including user-facing Interfaces, Agentic ops (agent-based operations), Urban ST (urban spatio-temporal forecasting), and our specific focus, Incident duration prediction.
While prior work has applied prompting and SFT to related urban tasks, our contribution (highlighted in the top-right) is the first to adapt RLVR for text-grounded incident duration forecasting.
} 

\label{fig:rw-map-pgfplots}

\end{figure}
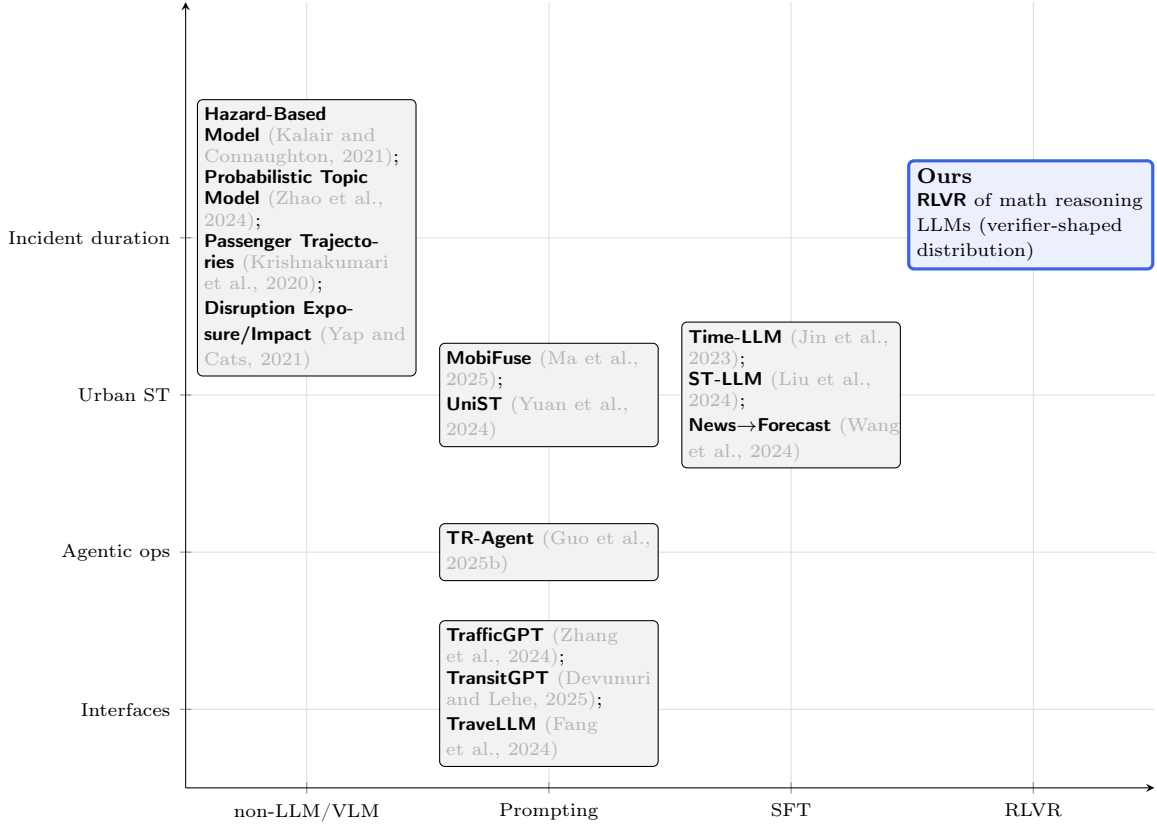

% === Framework overview ===
\begin{figure*}[t]
  \centering
  \includegraphics[width=\textwidth]{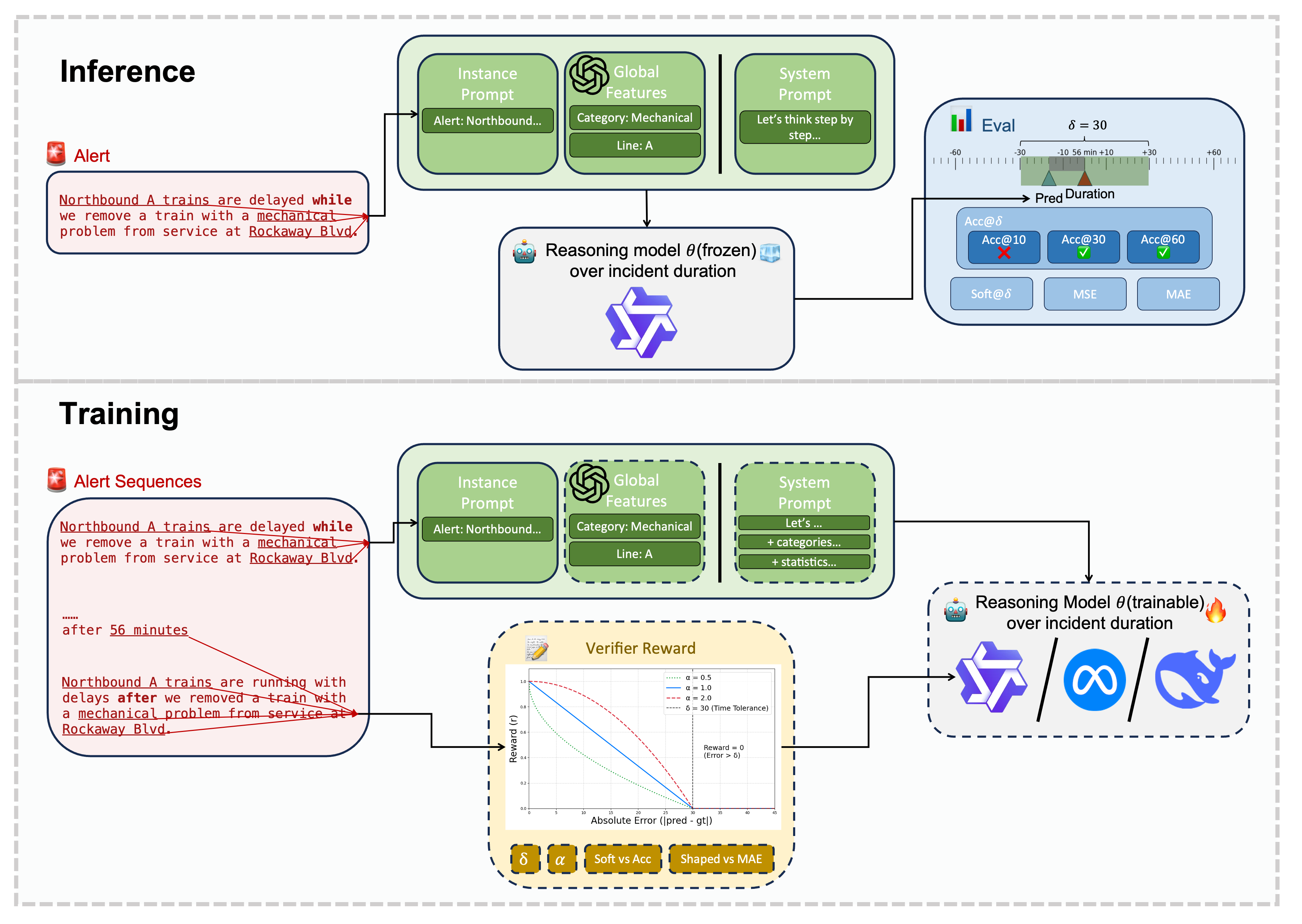} 
  \caption{%
\textbf{Framework overview} for incident-duration modeling from service alerts. 
\textbf{Top (Inference):} An incoming alert serves as the \textit{Instance Prompt}. It is combined with offline-extracted \textit{Global Features} and a \textit{System Prompt} to elicit reasoning. These inputs condition a frozen model ($\theta$) to produce a duration estimate. The \textbf{Evaluation} panel shows how this prediction is scored using a tolerance ruler (Acc@$\,\delta$, Soft@$\,\delta$) and standard metrics (MAE, MSE). 
\textbf{Bottom (Training):} The model is trained using RLVR. A ground-truth duration is derived from the full Alert Sequence. The \textit{Verifier Reward} function compares the model's prediction to this ground truth, assigning a reward based on tolerance bands ($\delta$) and optional shaping (e.g., soft vs. hard, $\alpha$ scaling). This reward signal updates the reasoning model ($\theta$).
\textbf{Figure Legend:} \textit{Solid} outlines denote the fixed inference path; \textit{dashed} outlines represent ablated design choices (prompting, reward variants, backbones). Colors indicate component roles: green for prompting context, yellow for the verifier, gray for the model, and blue for evaluation. 
Logos indicate interchangeable model backbones. 
}
\label{fig:framework-overview}
\end{figure*}

\section{Related Work}

\subsection{LLMs in Transportation}
\label{subsec:llms-transport}
The integration of large (multi)modal models into transportation research is gaining momentum. Recent articles aim to establish foundation pipelines that organize domain priors and task curricula (e.g., TRIP~\citep{liu2025trip}), or embed agentic loops that help update or audit traditional traffic models \citep{liu2025trip,guo2025automating}.  
% In traffic perception, multimodal LLMs/VLMs \sd{there are a lot more literature in VLM for autonomous driving. I suggest not to mention any here, only focus on LLM.} are adapted to road scenes \citep{kuang2025traffic}, 
Meanwhile, GNN-based models are applied for network-scale learning
% \sd{surrogates and reviews  remain complementary(?)} for network-scale learning 
\citep{xue2025data,narayanan2024graph}.

In user-facing applications (termed 'Interfaces' in Figure~\ref{fig:rw-map-pgfplots}), LLMs act as front-ends to connect natural language queries with structured transit/traffic data.
% On the \sd{interface (should we say "when it comes to the QA task", a more common term?)} side, \sd{LLM front-end?} connects natural language to structured transit/traffic data. 
\emph{TrafficGPT} demonstrates tool-augmented analysis for traffic management~\citep{zhang2024trafficgpt}, and \emph{TransitGPT} compiles user intents into executable code for GTFS queries and visual analytics \citep{devunuri2025transitgpt}. Application-specific systems increasingly consider disruptions, such as passenger choice under delays and re-routing in face of emergencies \citep{chen2024delayptc,fang2024travellm}. As positioned in Figure~\ref{fig:rw-map-pgfplots}, these systems predominantly occupy the “Prompting/Supervised Fine-Tuning (SFT)” columns, while our work targets RLVR with continuous targets. Table~\ref{tab:rw-summary-transport-fm} summaries representative problems and foundation model capabilities.

Beyond interfaces, a growing line of work shows that LLMs can act on spatio-temporal data. Representative work includes \emph{Spatial-Temporal LLMs}~\citep{liu2024spatial}, \emph{UniST}~\citep{yuan2024unist}; pretraining for traffic series~\citep{jin2021trafficbert}, 
and general LLM-as-forecaster methods~\citep{gruver2023large,jin2023time}. Event-aware forecasting further integrates textual signals, such as news, into quantitative forecasts~\citep{wang2024news}. More recently, foundation models for human mobility are  explored for cross-domain fusion \citep{ma5218098learning}. 
% \sd{shouldn't this go to the paragraph about satptial-temporal LLM?}.
Our contribution is complementary: we use free-form service-alert text to reason and forecast \emph{incident duration}, with feedback signals for fine-tuning, rather than focusing on dialog, classification, or explanation alone.

%While our ultimate goal is to forecast the duration of transit disruption, rather than offering route planning, we reference classical and modern transit routing and graph methods: 
There is also a body of literature on transit routing on graphs, including: bicriterion and hyper-partitioned transit routing~\citep{agarwal2024scalable,delling2017faster,witt2016trip}; transfer patterns and real-time pathfinding~\citep{bast2010fast,jariyasunant2011algorithm}; and fast path estimation in large graphs~\citep{gubichev2010fast}. These studies primarily emphasizes search efficiency and operational constraints, whereas we target early incident-duration prediction from unstructured text.

\subsection{Transit delay and incident-duration prediction}
\label{subsec:delay-duration}

Non-LLM models have demonstrated that delay and incident duration are predictable from structured signals, with recent studies spanning rail~\citep{sarhani2024prediction} and ferry~\citep{sarhani2025predicting} systems. 
% \sd{separate rail and ferry for each citation}.
Features derived from unstructured text logs have also been fused 
% \sd{Short-text features (? what features?)} have also been fused 
with structured covariates to improve metro incident-duration models~\citep{zhao2024predicting}. Hazard-based models, including dynamic formulations, are used to model clearing processes
% and \sd{dynamic formulations ? (what formulations?)} are used to model clearing processes
~\citep{li2018overview,kalair2021dynamic}. Trajectory- and stop-level 
analyses % \sd{formulations ? (what formulations?)} 
quantify delay propagation to riders~\citep{krishnakumari2020estimation,yap2021predicting}. In contrast to these studies (left column of Figure~\ref{fig:rw-map-pgfplots}), we focus on text grounded, early prediction of \emph{incident duration}, producing calibrated distributions.

\section{Preliminaries}

% \sd{Here you should introduce prompting, SFT/RLVR, and training algorithms related to LLM...}
\subsection{Post-training for distilling domain knowledge}
\label{subsec:rlvr}
% \sd{From this subsection onward, it should go to Sec. 3 Preliminary (preliminary knowledge needed to prepare to understand your method). Not related work.}

A simple way to inject domain knowledge is \emph{prompting}: conditioning the base model $\pi_{ref}(y\!\mid\!x)$ on problem context, tool specs, or examples so that the response distribution is steered without changing parameters \citep{brown2020language,liu2023pre}. Stronger adaptation comes from \emph{instruction/SFT}, where curated input–output pairs teach style and content directly \citep{wei2021finetuned,ouyang2022training}. Both methods 
% \sd{routes (what routes? methods or directions?)}
distill knowledge into the model—prompting at inference, SFT via gradient updates on demonstrations.

\subsubsection{A unifying post-training view: KL-regularized objectives.}
Preference-based post-training (e.g., RLHF) can be written, per input $x$, as the KL-regularized optimization
\begin{align}
\label{eq:kl_obj}
\max_{\pi(\cdot\mid x)} \
\mathbb{E}_{y\sim \pi(\cdot\mid x)}[\,r_\phi(x,y)\,]
\;-\;\tfrac{1}{\beta}\,
\mathrm{KL}\!\left(\pi(\cdot\mid x)\,\Vert\,\pi_{ref}(\cdot\mid x)\right),
\end{align}
with reference model $\pi_{ref}$ (often SFT), reward model $r_\phi$, and trade-off $\beta{>}0$ \citep{ziegler2019fine,ouyang2022training}. This yields the closed-form optimum
\begin{align}
\label{eq:boltzmann}
\pi^\star(y\mid x)
\;=\;
\frac{1}{Z_\beta(x)}\;\pi_{ref}(y\mid x)\,\exp\!\big(\beta\,r_\phi(x,y)\big),
\end{align}
i.e., an \emph{exponentially tilted} $\pi_{ref}$ by the reward landscape \citep{christiano2017deep,ouyang2022training}. Because building $r_\phi$ is costly, \emph{Direct Preference Optimization} (DPO) learns the policy directly from pairwise preferences implied by \eqref{eq:boltzmann} \citep{rafailov2023direct}.

\subsubsection{RLVR: replacing human rewards with verifiers.}
Recent work replaces human rewards with \emph{verifiable} signals: unit tests, symbolic solvers, or executable checks \citep{cobbe2021training,lewkowycz2022solving,lightman2023let}. Open efforts such as \emph{GRPO}/\emph{DAPO} couple verifier signals with group-relative objectives \citep{shao2024deepseekmath,zheng2025group,yao2025group,yu2025dapo,guo2025deepseek,yue2025does,shao2025spurious,wu2025reasoning}. Tooling like \emph{Math-Verify} and test-time scaling (self-consistency/tree search) further improve pass rates \citep{Kydlicek_Math-Verify_Math_Verification,muennighoff2025s1,jurayj-etal-2025-final,wang2022self,yao2023tree}.

\subsubsection{Our setting: continuous, noisy targets.}

Exact-match verifiers are ill-posed for \emph{incident duration}. We instantiate \eqref{eq:kl_obj} with \emph{tolerance-bounded, shaped rewards} tied to proper-scoring objectives, linking to RL with imperfect verifiers and risk-aware exploration \citep{cai2025reinforcement,wen2025reinforcement,li2025steering,cui2025entropy}. Concretely, the reward grants partial/full credit within a tolerance band and smoothly penalizes deviations, so the model learns distributions that are accurate for early incident duration.

% This is the exact message and lesson and what the reader should know, a TL;DR; shall not be removed!

% \vspace{2pt}
% \noindent\textit{Takeaway.} Along \emph{Prompting}~$\rightarrow$~\emph{SFT}~$\rightarrow$~\emph{RLHF/DPO}~$\rightarrow$~\emph{RLVR}, domain knowledge moves from handcrafted prompts and human labels toward \emph{programmatic verifiers}. 

\subsection{GRPO (Group Relative Policy Optimization).}
Given an input prompt $x$ from the training distribution $\mathcal{D}$ and a group of $G$ responses $\{y_i\}_{i=1}^G$ sampled from the current policy $\pi_{\theta_{\text{old}}}(\cdot|x)$, GRPO optimizes a clipped PPO-style objective. Its key innovation is the use of \emph{group-normalized} advantages, which stabilizes training by evaluating each response relative to others in the same batch. The objective is:
\begin{equation}\label{eq:grpo}
\mathcal{J}_{\mathrm{GRPO}}(\theta)
=
\mathbb{E}\!\left[
\frac{1}{G}\sum_{i=1}^{G}\frac{1}{|y_i|}\sum_{t=1}^{|y_i|}
\min\bigl(w_{i,t}(\theta)\,\widehat{A}_{i},
\mathrm{clip}\!\bigl(w_{i,t}(\theta),1-\varepsilon,1+\varepsilon\bigr)\,\widehat{A}_{i}\bigr)
\right],
\end{equation}
with the per-token probability ratio $w_{i,t}(\theta)$ and the standardized group advantage $\widehat{A}_{i}$ defined as:
\begin{equation}\label{eq:grpo-adv}
w_{i,t}(\theta)=
\frac{\pi_{\theta}(y_{i,t}\!\mid\! x,y_{i,<t})}{\pi_{\theta_{\mathrm{old}}}(y_{i,t}\!\mid\! x,y_{i,<t})},
\qquad
\widehat{A}_{i}=\frac{R_i-\mathrm{mean}(\{R_j\}_{j=1}^G)}{\mathrm{std}(\{R_j\}_{j=1}^G) + \epsilon_{\text{norm}}}.
\end{equation}
All tokens within a single response $y_i$ share the same advantage $\widehat{A}_i$, and the normalization by response length $|y_i|$ mitigates potential bias towards shorter or longer outputs.

\subsection{DAPO (Decouple Clip \& Dynamic sAmpling Policy Optimization).}
DAPO enhances group-based objectives with two key modifications. First, it decouples the clipping bounds $(\varepsilon_{\mathrm{low}}, \varepsilon_{\mathrm{high}})$ to allow for more aggressive updates on high-reward responses while maintaining conservative updates on low-reward ones. Second, it employs a dynamic sampling strategy to ensure that each training batch contains a mix of both high- and low-reward outputs relative to the ground truth $a=y_e$. Given a question $q$ and a sampled group of outputs $\{o_i\}_{i=1}^G \sim \pi_{\theta_{\text{old}}}(\cdot|q)$, the objective is:
\begin{equation}\label{eq:dapo}
\mathcal{J}_{\mathrm{DAPO}}(\theta)=
\mathbb{E}\!\left[
\frac{1}{\sum_{i=1}^G |o_i|}
\sum_{i=1}^G \sum_{t=1}^{|o_i|}
\min\bigl(r_{i,t}(\theta)\,\widehat{A}_{i},\
\mathrm{clip}\!\bigl(r_{i,t}(\theta),1-\varepsilon_{\mathrm{low}},1+\varepsilon_{\mathrm{high}}\bigr)\,\widehat{A}_{i}\bigr)
\right],
\end{equation}
This expectation is computed over batches that satisfy the condition $0 < |\{o_i: \texttt{is\_equivalent}(a, o_i)\}| < G$, where equivalence is determined by our tolerance-based reward: $\texttt{is\_equivalent}(a, o_i) = \mathbb{I}\{|\mathrm{parse}(o_i) - a| \le \delta\}$. The per-token ratio $r_{i,t}(\theta)$ and group advantage $\widehat{A}_{i}$ are defined analogously to GRPO. This combination of asymmetric clipping and balanced sampling focuses the training process on the most informative examples, improving learning efficiency.

% ============================
% METHOD
% ============================

\section{Methodology}\label{sec:methodology}

\subsection{Problem statement}   
% verbally describe the task
The task is to predict the total duration of a public transit incident from text alerts issued in the format of GTFS-rt, such as a bus delay, truck maintenance, or station closure, in real-time. We are given a stream of unstructured text alerts as they are reported by a transit authority. These alerts arrive sequentially, with each new alert providing updated information on an ongoing disruption. At any point during the incident, our model uses the sequence of alerts received up to that moment to predict the final duration of the disruption. To add more context of our task, Table~\ref{tab:alert-text-examples} offers some example text alerts.

\begin{table}[!htbp]
\centering
\small
\setlength{\tabcolsep}{6pt}
\renewcommand{\arraystretch}{1.15}
\caption{Examples of raw service-alert text and the corresponding incident category.}
\label{tab:alert-text-examples}
\begin{tabularx}{\textwidth}{@{}>{\raggedright\arraybackslash}X>{\raggedright\arraybackslash}p{0.28\textwidth}@{}}
\toprule
\textbf{Alert text} & \textbf{Category} (extracted by LLM; defined in \S\ref{sec:eda}) \\
\midrule
Southbound 2 3 trains are delayed while we request NYPD assistance for people being disruptive on a train at Hoyt St. & disruptive person \\
Northbound R trains are delayed while we address a train with a mechanical problem at Elmhurst Av. & mechanical problem \\
Northbound D trains are delayed while EMS responds to a person in need of medical help at 34 St-Herald Sq. & medical assistance \\
6 trains are running with delays in both directions while we address a signal problem near Brooklyn Bridge-City Hall. & signal problem \\
Northbound 1 trains are delayed while we remove debris from the track at 191 St. & debris on tracks \\
\bottomrule
\end{tabularx}
\end{table}

% \sd{provide some examples of text alerts here. maybe one from each category. So far it's still unclear what text alerts are like, except one example in Fig. 3}

\subsection{Problem formulation}
Let $\mathcal{A}$ be the universe of all text alerts and $\mathcal{E}$ be the set of distinct transit events. Each event $e \in \mathcal{E}$ is defined as a time-ordered sequence of one or more related alerts, representing the lifecycle of a single disruption from its initial report to its resolution. 
We write this sequence as 
\[
\mathbf{a}_e \;\coloneqq\; \{a_{e,1},\ldots,a_{e,m_e}\},\qquad
\text{where}\quad a_{e,j}=(t_{e,j},x_{e,j}),
\]
where $t_{e,j} \in \mathbb{R}_+$ is the timestamp of the $j$-th alert for event $e$, and $x_{e,j}$ is its raw text content. We partition the raw alert stream $\mathcal{A}$ into these event-specific sequences. 

For event $e$, we define its start time, end time, and total duration in minutes:
\[
t_s(e)\coloneqq \min_j t_{e,j},\qquad
t_f(e)\coloneqq \max_j t_{e,j},\qquad
y_e \coloneqq \frac{t_f(e)-t_s(e)}{60}\in\mathbb{R}_+.
\]
The ground-truth duration $y_e$ is the prediction target. 
We also define a discrete set of incident categories $\mathcal{C}$ derived from the data. 

For each event $e$, we may infer a category $c_e \in \mathcal{C}$ from the alert text to provide a prior for its likely duration. 

The task is to predict the total duration $y_e$ of event $e$, given only the alerts available up to a forecast time $\tau$ with $t_s(e) \le \tau \le t_f(e)$. 
In a real-time setting, $\tau$ is typically the timestamp of the most recent alert, $t_{e,m_e}$. The model $f_\theta$ produces a prediction $\hat{y}_e$ from this information: 
\[
\hat{y}_e \;=\; f_\theta\bigl(\mathbf{a}_e^{(\le\tau)}\bigr), \qquad \text{where}\quad
\mathbf{a}_e^{(\le\tau)}\coloneqq\{a_{e,j}:t_{e,j}\le \tau\}.
\]
To ensure a clean evaluation, we partition events into disjoint training and testing folds, $\mathcal{E}_{\mathrm{tr}}$ and $\mathcal{E}_{\mathrm{te}}$. Any \emph{global} statistics used to inform prompts (e.g., a table of by-category average durations) are computed only on $\mathcal{E}_{\mathrm{tr}}$ to prevent data leakage. 

% -------------------------
% Evaluation metrics: tolerance sweep
% -------------------------
\subsubsection{Evaluation metrics: tolerance sweep}\label{sec:c-sweep-metrics}
In addition to MAE/MSE, we report tolerance-based metrics over a grid $\mathcal{T}$ (e.g., $\{5,10,30,60,120\}$ minutes), which capture usefulness at different error budgets and align with our RL rewards (see \S\ref{sec:rlvr}). % 

For any $c \in \mathcal{T}$ and predictions $\{\hat{y}_e\}_{e\in\mathcal{E}_{\mathrm{te}}}$:
\begin{align}
\mathrm{acc}_c &\;=\; \frac{1}{|\mathcal{E}_{\mathrm{te}}|}\sum_{e\in\mathcal{E}_{\mathrm{te}}}\mathbb{I}\!\left(|\hat{y}_e-y_e|\le c\right), \label{eq:acc_c}\\
\mathrm{soft}_c &\;=\; \frac{1}{|\mathcal{E}_{\mathrm{te}}|}\sum_{e\in\mathcal{E}_{\mathrm{te}}}\max\!\left(1-\frac{|\hat{y}_e-y_e|}{c},\,0\right). \label{eq:soft_c}
\end{align}
$\mathrm{acc}_c$ is the fraction of events where the absolute difference between the predicted time and the ground truth time is within $c$ minutes. $\mathrm{soft}_c\in[0,1]$ gives partial credit within the $c$-minute band (linearly decreasing to $0$ at $c$). We report both because $\mathrm{acc}_c$ is easy to read, while $\mathrm{soft}_c$ preserves ranking among near-correct predictions. 

% -------------------------
% Model class II 
% -------------------------
\subsection{Post-trained LLM via RLVR for Incident Duration}\label{sec:rlvr}
To overcome the limitations of the feature extraction baseline, we treat the LLM itself as the predictor and fine-tune it end-to-end using RLVR. This approach allows the model's reasoning process to be directly shaped by feedback on its prediction accuracy. The model takes a task instruction and the alert history as input and generates a natural language response containing a numeric prediction.

\subsubsection{Design of Incident Duration Prediction}
\paragraph{Policy and verifiable parsing.}
We model the LLM as a policy $\pi_\theta(\cdot\,|\,\mathbf{p}_e)$ that generates a sequence of tokens conditioned on a prompt $\mathbf{p}_e$, constructed from alerts $\mathbf{a}_e^{(\le\tau)}$ and, in some variants, dataset-level knowledge (see \S\ref{sec:prompts}). To extract a numerical value from free-text, we apply a deterministic parser $\mathrm{parse}(\cdot)$ adapted from robust verifiers used in mathematical reasoning (e.g., Math-Verify~\cite{Kydlicek_Math-Verify_Math_Verification}):
\[
\widehat{y}_e \;=\; \mathrm{parse}\!\left(\text{output from }\pi_\theta(\cdot\,|\,\mathbf{p}_e)\right)\;\in\;\mathbb{R}_+\cup\{\bot\}.
\]
If parsing fails ($\bot$), we set the reward to zero for tolerance-based rewards (R1/R2) and, for the MAE baseline (R0) defined below, we set the error to $e_e=\delta$. 

\paragraph{Tolerance-based reward design.}
Directly using negative MAE/MSE as a reward signal is numerically unstable for policy gradient methods in this task, as the large and unbounded range of durations can lead to high-variance gradients and training collapse. We therefore employ bounded rewards $r_e\in[0,1]$ governed by a tolerance $\delta>0$ and absolute error $e_e \coloneqq |\widehat{y}_e - y_e|$.
We define and compare three reward formulations used in this study:
% \sd{Below, we present three rewards used for this study.}

\emph{(R0) Negative-MAE} 
\begin{equation}\label{eq:r0}
r^{\mathrm{mae}}_e \;=\; -\,e_e,
\quad\text{with}\quad
e_e=\begin{cases}
|\widehat{y}_e-y_e|, & \text{if }\widehat{y}_e\in\mathbb{R}_+,\\
\delta, & \text{if parsing fails }(\widehat{y}_e=\bot).
\end{cases}
\end{equation}
We include R0 as an ablation for the bounded rewards designed below.

\emph{(R1) Binary-within-tolerance.}
\begin{equation}\label{eq:r1}
r^{\mathrm{bin}}_e \;=\; \mathbb{I}\{\,e_e < \delta\,\}.
\end{equation}

\emph{(R2) Shaped-within-tolerance.}
\begin{equation}\label{eq:r2}
r^{\mathrm{shp}}_e \;=\; \Bigl[\,1 - \Bigl(\tfrac{e_e}{\delta}\Bigr)^{\alpha}\Bigr]_+,
\qquad \alpha \ge 1,\quad [u]_+ \coloneqq \max\{u,0\}.
\end{equation}
The parameter $\alpha$ controls the steepness of the decay. We conduct a sensitivity analysis for both $\delta$ and $\alpha$ in our experiments. For all RL-based training, the final reward $R_i$ for a given response is set to $r_e \in [0,1]$, which is then used to compute the advantage for policy updates.

\paragraph{Relationship and properties.}

The shaped reward~\eqref{eq:r2} strictly generalizes the binary reward~\eqref{eq:r1} and 
% \sd{exposes?} 
is controlled via $(\delta,\alpha)$. 
Below, we present the properties of our proposed reward design.
% \sd{Below, we present the properties of our proposed reward design.}

% \sd{what does G represent? neither relationship nor properties...}
\textbf{(1) R2 $\Rightarrow$ R1 as a limit (exact).} For any fixed $\delta>0$,
\[
\lim_{\alpha\to\infty} r^{\mathrm{shp}}_e \;=\; \mathbb{I}\{\,e_e < \delta\,\} \;=\; r^{\mathrm{bin}}_e,
\]
since $(e_e/\delta)^\alpha \to 0$ for $e_e<\delta$ and $\to 1$ for $e_e\ge\delta$. Thus \eqref{eq:r2} recovers \eqref{eq:r1} as a limiting case while providing a smooth continuum for finite $\alpha$.

\textbf{(2) Monotonicity in error and tolerance.} On $0\le e_e < \delta$,
\[
\frac{\partial r^{\mathrm{shp}}_e}{\partial e_e} \;=\; -\frac{\alpha}{\delta}\Bigl(\tfrac{e_e}{\delta}\Bigr)^{\alpha-1}\le 0,
\qquad
\frac{\partial r^{\mathrm{shp}}_e}{\partial \delta} \;=\; \alpha\,\frac{e_e^{\alpha}}{\delta^{\alpha+1}}\;\begin{cases}
=\,0,& e_e=0,\\[2pt]
>\,0,& 0<e_e<\delta.
\end{cases}
\]
Globally, $r^{\mathrm{shp}}$ is $(\alpha/\delta)$-Lipschitz in $e_e$ and identically $0$ for $e_e\ge\delta$. \emph{This Lipschitz constant $L=\alpha/\delta$ controls the sensitivity of the GRPO/DAPO
normalized advantages; see (G6) for the explicit bound.}

\textbf{(3) Effect of shape $\alpha$.} For $0<e_e<\delta$,
\[
\frac{\partial r^{\mathrm{shp}}_e}{\partial \alpha} \;=\; -\Bigl(\tfrac{e_e}{\delta}\Bigr)^{\alpha}\log\!\Bigl(\tfrac{e_e}{\delta}\Bigr)\;>\;0,
\]
since $\log(e_e/\delta)<0$. Increasing $\alpha$ “hardens’’ the reward: it pushes $r^{\mathrm{shp}}$ closer to $1$ throughout the interior $e_e<\delta$ while steepening the drop near the boundary $e_e\approx\delta$ (slope magnitude at $e_e\!\uparrow\!\delta$ equals $\alpha/\delta$). Small $\alpha$ yields smoother shaping and finer discrimination among near-correct predictions.

\textbf{(4) Scale invariance.} The family depends on the \emph{relative} error: $r^{\mathrm{shp}}(e_e;\delta,\alpha)=g_\alpha(e_e/\delta)$ with $g_\alpha(z)=[1-z^\alpha]_+$. Consequently, choosing $\delta$ rescales the acceptance window without altering the functional form.

\textbf{(5) Limiting behaviors.} As $\delta\to 0^+$, $r^{\mathrm{shp}}_e\to 0$ for any fixed $e_e>0$; as $\delta\to\infty$, $r^{\mathrm{shp}}_e\to 1$. These limits formalize the tolerance–strictness trade-off.

\textbf{(6) Bounded sensitivity of normalized advantages.}
Let $L \coloneqq \alpha/\delta$. For a group of $G$ rollouts, the normalized advantage is $\widehat{A}_i$ as defined in Eq.~\ref{eq:grpo-adv}. Let $\mu = \mathrm{mean}(\{R_j\}_{j=1}^G$ and $\sigma = \mathrm{std}(\{R_j\}_{j=1}^G)$, with $R_i=r^{\mathrm{shp}}(e_i)\in[0,1]$. 
For a perturbed group $\{e'_j\}$, let $\Delta e_j \coloneqq e'_j-e_j$ and 
$\delta_e \coloneqq \max_j |\Delta e_j|$. 
Because $r^{\mathrm{shp}}$ is $L$-Lipschitz in $e$ (see (G2)), we have 
$|R'_j-R_j|\le L\,|\Delta e_j|$ for all $j$. 
Then the normalized advantages obey
\begin{equation}\label{eq:g6-main}
\bigl|\widehat{A}'_i-\widehat{A}_i\bigr|
\;\le\; \frac{L}{\sigma+\epsilon_{\text{norm}}}\Bigl(|\Delta e_i|+\delta_e\Bigr)
\;+\; \frac{2L\,\delta_e}{(\sigma+\epsilon_{\text{norm}})^2}\Bigl(\,|R_i-\mu| + L\bigl(|\Delta e_i|+\delta_e\bigr)\Bigr).
\end{equation}
Hence increasing $\alpha$ or decreasing $\delta$ (i.e., increasing $L=\alpha/\delta$) makes 
$\widehat{A}_i$ more sensitive, while larger within-group spread $\sigma$ damp changes. 

% -------------------------
% Prompt templates and knowledge injection
% -------------------------
\subsubsection{Prompt templates and knowledge injection}\label{sec:prompts}
% REVISION: describe templates and refer to the verbatim versions below; remove duplicated prompt text here.

% \sd{since this heavily relies on the definition of Category list and how you define them, should it be moved to after Sec. 5.2? Or move Sec. 5.2 up}

The performance of LLMs may be sensitive to prompt design. We study three \emph{prompt templates} that incrementally inject dataset-level knowledge (exact prompt content in \emph{Prompts (verbatim text)} below).

\begin{enumerate}[label=(P\arabic*)]
    \item \textbf{Reasoning add-on (CoT).} Appends a short instruction to encourage step-by-step reasoning, relying only on the instance text and pretrained knowledge. 
    \item \textbf{Category list.}
    % \sd{needs to highlight that this information is not directly available, we need to extract such categories using LLM...}
    Appends the fixed set of incident categories $\mathcal{C}$, which are derived from alert text using an LLM (as detailed in \S\ref{sec:eda}), and asks the model to first infer a likely category before predicting the duration. The intermediate category is a latent step (not directly supervised). 
    \item \textbf{Category statistics.} Appends $\mathcal{C}$ and a per-category summary table $\mathcal{S}$ (see \S\ref{sec:eda} for details on $\mathcal{C}$ and $\mathcal{S}$), computed only on $\mathcal{E}_{\mathrm{tr}}$ and held fixed during validation/test to avoid leakage:
    \[
    \mathcal{S}=\bigl\{(\text{count}_c,\mu_c,\sigma_c,q_{c,0.25},q_{c,0.50},q_{c,0.75},\text{min}_c,\text{max}_c):c\in\mathcal{C}\bigr\}.
    \]
\end{enumerate}

\paragraph{Why these designs differ.}
% REVISION: concise rationale and cross-reference to verbatim prompts.
Moving from (P1) to (P3) increases dataset-level guidance. (P2) supplies global structure unavailable from a single alert via $\mathcal{C}$; (P3) adds population-level priors via $\mathcal{S}$, which can improve calibration but may over-anchor the prediction. See \emph{Prompts (verbatim text)} for the exact instructions.

\paragraph{Prompts.} %(verbatim text)
We study the effectiveness of three prompt variants (P1–P3):
\begin{quote}\small
\textbf{Base instruction (all P1–P3):}\\
\texttt{Based on the following transit alert(s), predict the total duration of the delay in minutes.}\\[2pt]
\textbf{P1 (reasoning add-on):}\\
\texttt{Let's think step by step.}\\[2pt]
\textbf{P2 (categories add-on):}\\
\texttt{First infer the incident category from the list \{\,$\mathcal{C}$\,\}, then predict the duration.}\\[2pt]
\textbf{P3 (categories + statistics add-on):}\\
\texttt{You may consult the per-category statistics table \{\,$\mathcal{S}$\,\} to calibrate your prediction.}
\end{quote}

\section{Experiments}
% TODO: describe the experiment settings,
% TODO: it appears with smaller c (stricter), the advantage of LLMs is larger
% Author by Ruijian
% revised by Bowen
\subsection{Data}\label{subsec:data-intro}
% TODO: add GTFS-RT 
We study NYC MTA service alerts from \textbf{April 28, 2020} onward, yielding \textbf{21{,}092 events} with verified temporal boundaries: start $t_s$, end $t_f$, and duration $y=t_f-t_s$. Because events themselves do not state their duration, we infer boundaries from the alert sequence: for each event we feed all related alerts (in order) to \texttt{DeepSeek-R1-Distill-Qwen-7B} and ask whether any alert explicitly concludes the incident (e.g., “delays cleared,” “service has resumed”). 
% \sd{give more examples to illustrate the data format, how to extract information...}

\begin{table}[!htbp]
\centering
\small
\setlength{\tabcolsep}{4pt}
\renewcommand{\arraystretch}{1.08}
\caption{Duration statistics by event category for NYC MTA transit events dataset (N=21{,}092). Duration measured in minutes.}
\label{tab:duration_stats}
\begin{tabular}{l|rr|rrrr}
\toprule
\textbf{Category} & \textbf{Count} & \textbf{\%} & \textbf{Mean} & \textbf{Median} & \textbf{Std} & \textbf{Range} \\
\midrule
External Factors & 4696 & 22.3\% & 91.2 & 22.0 & 190.3 & 0--3597 \\
Operational & 1834 & 8.7\% & 94.6 & 49.0 & 183.0 & 0--3646 \\
Other & 445 & 2.1\% & 38.3 & 20.0 & 56.5 & 1--392 \\
Passenger Incident & 4099 & 19.4\% & 15.4 & 11.0 & 18.6 & 0--551 \\
Signal Control & 1434 & 6.8\% & 62.8 & 42.0 & 67.0 & 0--636 \\
Station Related & 498 & 2.4\% & 9.2 & 7.0 & 9.3 & 1--86 \\
Track Infrastructure & 538 & 2.6\% & 22.7 & 10.0 & 38.4 & 0--513 \\
Train Mechanical & 7548 & 35.8\% & 14.6 & 9.0 & 46.4 & 0--3752 \\
\midrule
\textbf{Overall} & 21092 & 100.0\% & 42.6 & 13.0 & 116.1 & 0--3752 \\
\bottomrule
\end{tabular}
\end{table}

We generate 8 independent responses per event and apply majority voting to decide if a conclusive alert exists; when positive, $t_f$ is the timestamp of the voted terminal alert and $t_s$ is the timestamp of the event’s first alert. Events without a voted terminal alert are excluded from the final set. At corpus scale, $\sim$50{,}000 raw events are processed, yielding $\sim$400{,}000 model responses (8{,}096 max response length per prompt) and completes in $\sim$3 hours on a single H100 (80GB) GPU. We apply lightweight consistency checks (non-negative duration, monotone timestamps, de-duplication) to finalize boundaries. This LLM-assisted procedure is motivated by the high lexical variability of alert phrasing and the absence of structured “resolved” flags, which make rule-only extraction brittle while still allowing us to keep the inference path auditable and reproducible.

% redundant to the table tab:duration_stats
% \begin{figure}[t]
% \centering
% \includegraphics[width=1\linewidth]{Figs/eda_v2/fig1_category_distribution.png}
% \caption{\textbf{Distribution of transit event categories.} 
% Duration extracted from NYC MTA service alerts starting from April 28, 2020. Categories are extracted from GTFS-rt alerts using LLM, detailed in \S\ref{sec:eda}.
% Train mechanical issues (35.8\%, N=7,548) and external factors (22.3\%, N=4,696) constitute the majority of events, representing the most frequent disruption types in the NYC transit system. Eight distinct categories capture the full range of transit disruptions. 
% % \sd{Should we use this figure to replace Fig. 1? This already contains information about what you need to convey in Sec. 1.}
% }
% \label{fig:category_distribution}
% \end{figure}

\subsection{Exploratory data analysis (EDA): categories and durations}
\label{sec:eda}
\textbf{Feature scope.} Unless otherwise noted, \emph{all methods in \S\ref{sec:methodology} use the detailed categorical features produced by the frozen LLM in \S\ref{subsubsec:llm-extracter}} (fine-grained incident types). For exploratory summaries, we merge these into \textbf{eight mutually exclusive} macro–categories:
\emph{train mechanical}, \emph{passenger incident}, \emph{external factors}, \emph{operational}, \emph{signal control}, \emph{track infrastructure}, \emph{station related}, and \emph{other}.
The dataset is imbalanced, with \emph{train mechanical} (35.8\%) and \emph{external factors} (22.3\%) dominating (Table~\ref{tab:duration_stats}). Durations are heavy-tailed (overall median 13\,min; mean 42.6\,min; $\sigma{=}116.1$\,min).

\begin{figure}[!htbp]
\centering
\includegraphics[width=1\linewidth]{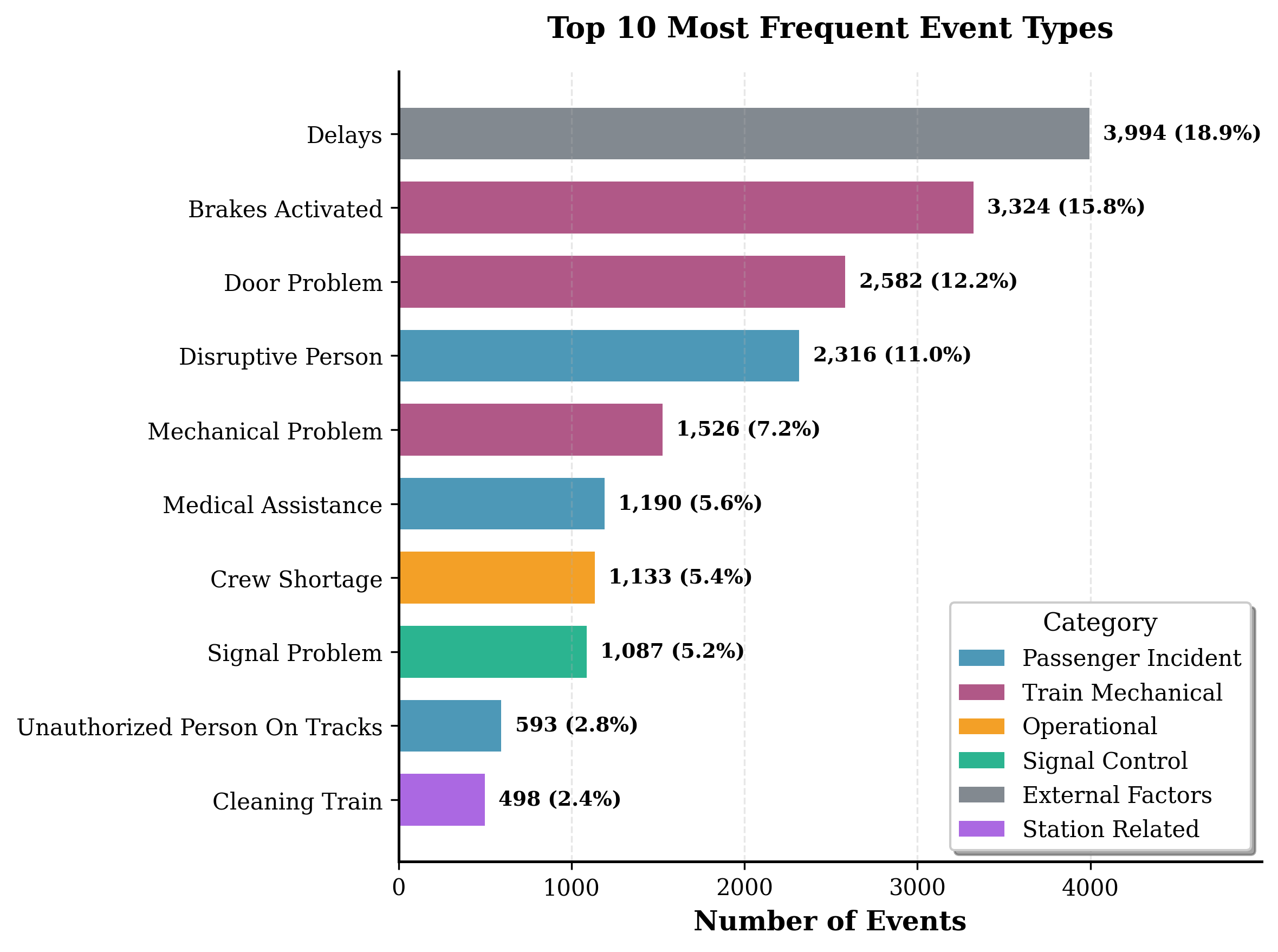}
\caption{\textbf{Top 10 most frequent event types.} Among 26 fine-grained event types, train service updates (13.2\%, N=2,784) and mechanical problems (12.8\%, N=2,702) are most common. Medical assistance is the leading passenger incident type (8.7\%, N=1,835). The top 10 types account for 67.3\% of all events.}
\label{fig:top_event_types}
\end{figure}

\textbf{Fine$\to$coarse mapping.}
For each event $e$, we prompt the frozen LLM with the full alert sequence (in order) and obtain a 26-way fine label. We generate 8 independent responses and assign the event label by majority vote; this single label is then deterministically merged into one of the eight macro classes for analysis (Table~\ref{tab:category_mapping}). Models in \S\ref{sec:methodology} always use the \emph{fine} labels; the macro labels are for Exploratory Data Analysis (EDA) 
% \sd{has its full name defined earlier?} 
only.

\begin{figure}[!htbp]
\centering
\includegraphics[width=1\linewidth]{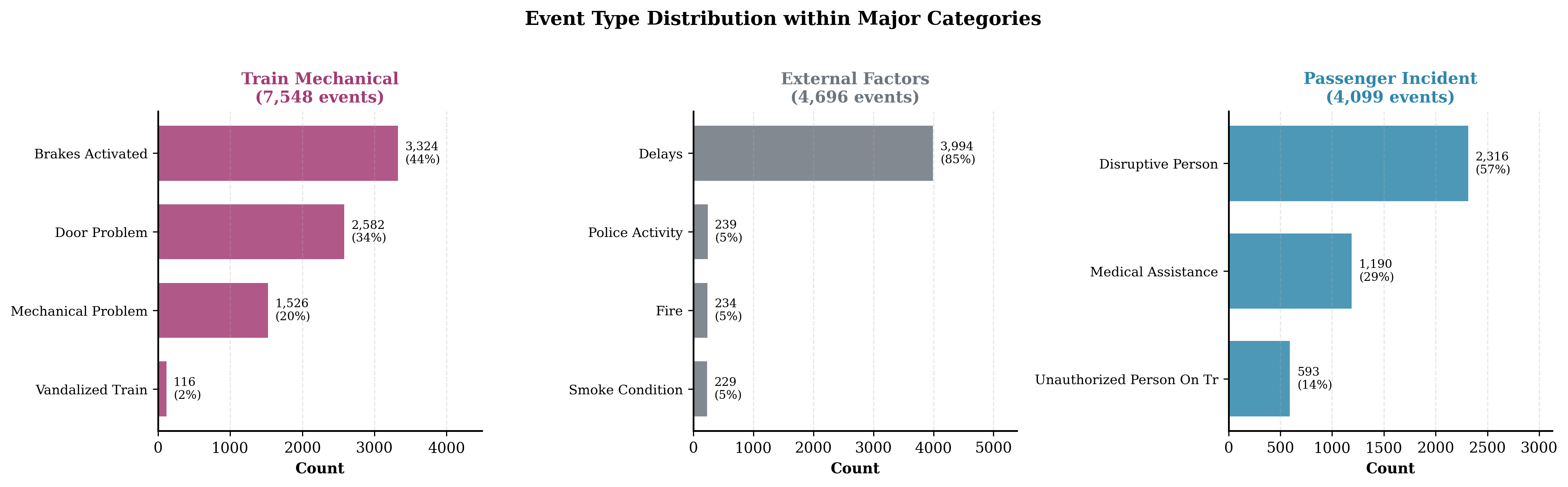}
\caption{\textbf{Event type distribution within major categories.} Internal composition of the three largest categories reveals concentrated patterns: Train Mechanical dominated by mechanical problems (35.8\%) and brake activations (23.9\%); External Factors primarily delays (60.4\%); Passenger Incidents led by medical assistance (44.8\%).}
\label{fig:category_breakdown}
\end{figure}

Beyond the eight categories, the \textbf{26 fine types} capture recurrent patterns (Figs.~\ref{fig:top_event_types}, \ref{fig:category_breakdown}). Per-category histograms reveal distinct shapes—sharp low-duration modes for \emph{passenger incident} and \emph{train mechanical}, and broader long-tailed profiles for \emph{external factors} and \emph{operational}.

\begin{table}[!htbp]
\centering
\small
\setlength{\tabcolsep}{6pt}
\renewcommand{\arraystretch}{1.08}
\caption{Deterministic merge from fine types (frozen LLM; \S\ref{subsubsec:llm-extracter}) to the eight macro categories used for exploratory analysis. Fine labels are shown verbatim (space-separated).}
\label{tab:category_mapping}
\begin{tabular}{ll}
\toprule
\textbf{Macro category} & \textbf{Fine types} \\
\midrule
\emph{Train mechanical} &
\texttt{mechanical problem}, \texttt{door problem}, \texttt{brakes activated}, \texttt{vandalized train} \\
\emph{Passenger incident} &
\texttt{medical assistance}, \texttt{disruptive person}, \texttt{unauthorized person on tracks} \\
\emph{External factors} &
\texttt{delays}, \texttt{police activity}, \texttt{fire}, \texttt{smoke condition}, \texttt{debris on tracks} \\
\emph{Operational} &
\texttt{train service update}, \texttt{bus service update}, \texttt{detour}, \texttt{detour cleared}, \\ & \texttt{residual delays}, \texttt{crew shortage}, \texttt{work train issue}, \texttt{ferry service issue} \\
\emph{Signal control} &
\texttt{signal problem}, \texttt{switch problem} \\
\emph{Track infrastructure} &
\texttt{track condition}, \texttt{power issue} \\
\emph{Station related} &
\texttt{cleaning train} \\
\emph{Other} &
\texttt{other} \\
\bottomrule
\end{tabular}
\end{table}

\subsection{Baselines}
\subsubsection{Model class 0: Classical baselines}\label{sec:baselines-non-llm}

\textbf{B0: Global-mean predictor.}
The global mean baseline is a sanity check that anchors evaluation with a trivial reference, where $\hat{y}_e=\mu_{\mathrm{tr}}$ and
$\mu_{\mathrm{tr}}=\frac{1}{|\mathcal{E}_{\mathrm{tr}}|}\sum_{i\in\mathcal{E}_{\mathrm{tr}}}y_i$. 

\textbf{B1: Tokenizer-hash bag-of-IDs.}
Given a tokenizer $T$ (used only to map text to integer token IDs), we build a fixed-size bag-of-IDs by hashing token IDs into $d$ buckets via modulo, 
\[
\phi^{(T,d)}_e[b]=\#\{i:\texttt{ID}_i \bmod d = b\},\qquad b=0,\ldots,d-1,
\]
then apply row-wise $\ell_2$ normalization. 
We sweep tokenizers $T$ and $d\in\{32,64,128,256,512\}$. 
This baseline is included because it is LLM-free, fast to compute, and controls dimensionality while still capturing coarse lexical variation. 

\begin{table}[t]
\centering
\small
\setlength{\tabcolsep}{4pt} % REVISION: tighter to avoid overflow
\renewcommand{\arraystretch}{1.1}
\caption{Tokenized alert length for train/test splits.}
\label{tab:tok-lengths}
\begin{threeparttable}
\begin{tabularx}{\textwidth}{l *{4}{>{\centering\arraybackslash}X}}
\toprule
\textbf{Tokenizer family} &
\makecell{\textbf{Mean}\\(train / test)} &
\makecell{\textbf{95th pctl.}\\(train / test)} &
\makecell{\textbf{99th pctl.}\\(train / test)} &
\makecell{\textbf{Max}\\(train / test)} \\
\midrule
Qwen\tnote{a}   & 28.36 / 28.48 & 57 / 57 & 78 / 82 & 334 / 271 \\
Llama-3.1-8B    & 27.40 / 27.50 & 54 / 54 & 73 / 79 & 309 / 247 \\
\bottomrule
\end{tabularx}
\begin{tablenotes}[flushleft]
\footnotesize
\item[a] Statistics are identical for \texttt{DeepSeek-R1-Distill-Qwen}, \texttt{Qwen2.5 Instruct}, and \texttt{Qwen2.5 Math}.
\end{tablenotes}
\end{threeparttable}
\end{table}

\emph{Choice of $d$.}
As Table~\ref{tab:tok-lengths} shows, inputs are short (means $\approx$27–28 tokens; 95th percentiles $\approx$54–57), and maximum length is around 300 for Qwen and Llama. Choosing $d\in[32,512]$ balances collision rate and computation without introducing very high-dimensional sparse features.

% -------------------------
% Model class I — LLM as feature extractor
% Change: any use of event type belongs here because \hat c_e is predicted by GPT-5-Pro.
% -------------------------
\subsubsection{Model class I: LLM as feature extractor}\label{subsubsec:llm-extracter}
% REVISION: simplify lead sentence; make “frozen at inference” explicit.
We use the LLM only as a \emph{feature extractor} and keep it frozen. 
A frozen LLM $g_\zeta$ (GPT-5-Pro) maps the alert history up to $\tau$ into a fixed vector $\mathbf{z}_e\in\mathbb{R}^p$, built from % REVISION: add “up to $\tau$” for consistency with the problem setup.
(i) a predicted incident category $\hat{c}_e\in\mathcal{C}$, (ii) affected entities (e.g., lines or stations), and (iii) simple temporal aggregates (e.g., number of alerts so far). % REVISION: clarify examples briefly
To stabilize $\hat{c}_e$, we use majority voting over 8 responses. % (kept)

% REVISION: add reason and leakage guard for category-mean baseline.
\textbf{Category-mean baseline.}
We also report a category-only predictor: $\hat{y}_e=\mu_{\mathrm{tr}}(\hat{c}_e)$, the training-set mean duration for the predicted category.
This isolates the value of category prediction itself. All means are computed on $\mathcal{E}_{\mathrm{tr}}$ to avoid leakage. % new explanatory sentence

\emph{Limitation.} The LLM parameters are not updated; features are fixed, so we cannot steer the model’s internal reasoning for this task. % REVISION: concise limitation statement

\subsubsection{Regression layer}
For B1 and Model class I (M1), we train the same downstream regressor on a generic feature vector $\mathbf{v}_e\in\mathbb{R}^q$, where
$\mathbf{v}_e=\phi^{(T,d)}_e$ for B1 and $\mathbf{v}_e=\mathbf{z}_e$ for M1.
The regressor $h_\beta:\mathbb{R}^q\!\to\!\mathbb{R}_+$ is fit with MSE:
\begin{equation}\label{eq:m0}
\mathcal{L}_{\text{reg}}(\beta)
=\frac{1}{|\mathcal{E}_{\mathrm{tr}}|}
\sum_{e\in\mathcal{E}_{\mathrm{tr}}}
\bigl(y_e-h_\beta(\mathbf{v}_e)\bigr)^2,
\qquad
\hat{y}_e=h_\beta(\mathbf{v}_e).
\end{equation}

We use the same set of shallow regressors (tree/boosting, $k$NN, SVR) from scikit-learn~\citep{scikit-learn} to decouple representation from predictor capacity. 

All models report MAE/MSE and the tolerance metrics in Eqs.~\eqref{eq:acc_c}–\eqref{eq:soft_c}. % (kept)

\subsection{Summary of variants}
% REVISION: Replace generic lead-in; align rows with B0, B1, M1, M2 and their prompt variants; remove unused columns.
Table~\ref{tab:methods} summarizes the non-LLM baselines (B0–B1), the frozen LLM feature extractor (M1), and the end-to-end RLVR model (M2) evaluated with prompt templates P1–P3 (\S\ref{sec:prompts}). No ground-truth info is used at test time; any category usage relies on the predicted $\hat{c}_e$.

\begin{table}[!htbp]
\centering
\small
\setlength{\tabcolsep}{5pt}
\renewcommand{\arraystretch}{1.12}
\caption{Methods summary 
% \sd{M0, B0, M1 are confusing. M1 are used in two meanings. I suggest deleting M0, M1, M2 in the title, or give M1: LLM features... a different shorthand name (M1-regressor...?)}
}
\label{tab:methods}
\begin{tabularx}{\textwidth}{l
>{\centering\arraybackslash}m{2cm}
>{\centering\arraybackslash}m{2cm}
>{\centering\arraybackslash}m{2.5cm}
X}
\toprule
\textbf{Method (variant)} &
\textbf{LLM role} &
\makecell{\textbf{Supervision} / \\ \textbf{reward}} &
\makecell{\textbf{Inference-time} \\ \textbf{global info}} &
\textbf{Input} \\
\midrule
\multicolumn{5}{l}{\emph{Baselines (no LLM generation)}} \\
B0: Global mean
& None
& None
& Train-set mean $\mu_{\mathrm{tr}}$
& $\hat{y}_e=\mu_{\mathrm{tr}}$ \\
B1: Tokenizer-hash + regressor
& \textit{LLM tokenizer} 
& MSE on $y$ (Eq.~\ref{eq:m0})
& None
& $\phi^{(T,d)}_e$ \\
\midrule
\multicolumn{5}{l}{\emph{LLM as feature extractor (frozen)}} \\
M1–Cat: Category-mean via $\hat{c}_e$
& Extractor
& Train-set per-category means
& $\mathcal{C}$
& $\hat{y}_e=\mu_{\mathrm{tr}}(\hat{c}_e)$ \\
M1: LLM features + regressor
& Extractor
& MSE on $y$ (Eq.~\ref{eq:m0})
& None 
& $\mathbf{z}_e$ \\
\midrule
\multicolumn{5}{l}{\emph{End-to-end RLVR (ours)}} \\
M2–P1: RLVR with reasoning prompt
& End-to-end
& R0, R1, R2
& None
& P1, $\mathbf{a}_e^{(\le\tau)}$ \\
M2–P2: RLVR + category list
& End-to-end
& R0, R1, R2
& $\mathcal{C}$
& P2, $\mathbf{a}_e^{(\le\tau)}$ \\
M2–P3: RLVR + categories \& statistics
& End-to-end
& R0, R1, R2
& $\mathcal{C}$, $\mathcal{S}$ (from $\mathcal{E}_{\mathrm{tr}}$)
& P3, $\mathbf{a}_e^{(\le\tau)}$ \\
\bottomrule
\end{tabularx}
\end{table}

\subsection{Setup}
We organize our study around a set of parameters of interest: algorithm, reward (R0/R1/R2; Eqs.~\ref{eq:r0}--\ref{eq:r2}), prompt (P1/P2/P3; Section~\ref{sec:prompts}), backbone, tolerance $\delta$, and steepness $\alpha$. The parameters and their candidate values are summarized in Table~\ref{tab:param-grid}. All models are trained with VeRL~\citep{yu2025dapo} on a Slurm cluster~\citep{bloom2025empire}, with a total of 80 training jobs and 80 evaluation jobs; each job runs on a single node with two H100 (80GB) GPUs, 16 CPU cores, and 128~GB RAM. We cap wall time at six hours for training and two hours for evaluation, yielding approximately $1{,}000$ GPU-hours in total. We shard with FSDP across the two local GPUs, with no tensor parallelism. Inputs use a left-truncated instance prompt capped at 2{,}048 tokens, and the maximum response length is 4{,}096 tokens for all models (effective maximum position embeddings of 6{,}144). We enable the overlong buffer in the verifier: a linear penalty ramps from 0 up to 1 over the next 4{,}096 tokens (penalizing overlong outputs that still fit within the hard context limit). Optimization follows GRPO-style advantage estimation with PPO clipping. For DAPO we use asymmetric ratio bounds \(\epsilon_{\text{low}}=0.20\) and \(\epsilon_{\text{high}}=0.28\) with constant \(c=10\) as used in Dual-clip PPO, while for GRPO we use a symmetric \(\epsilon=0.20\) in both directions.  No entropy regularization is applied, and gradient clipping is set to 1.0. We use AdamW with a learning rate of \(10^{-6}\), 10 warm-up steps, and weight decay of 0.1. For rollout, we draw 8 rollouts per prompt at temperature 1.0 with \emph{top-p} set to 1.0 and \emph{top-$k$} disabled. We use dynamic batch sizing and cap rollout GPU memory utilization at 0.6. Unless otherwise noted, the batch size is 64 and the mini-batch size is 64, which is equivalent to one gradient update per rollout step. For KL regularization, we follow the respective implementations, where KL regularization is disabled for DAPO, and enabled for GRPO with coefficient \(10^{-3}\) using the low-variance surrogate ~\citep{schulman2020approximating}. We train for a total of 100 training steps. The base models we used are: \texttt{DeepSeek-R1-Distill-Qwen-7B}, \texttt{Qwen2.5-7B-Instruct}, \texttt{Qwen2.5-Math-7B}, and \texttt{Llama-3.1-8B-Instruct}.

\begin{table*}[!htbp]
\centering
\footnotesize
\setlength{\tabcolsep}{6pt}
\renewcommand{\arraystretch}{1.12}
\caption{Experimental parameter grid. Each row lists a parameter and the range of values explored.}
\label{tab:param-grid}
\begin{tabularx}{\textwidth}{l X}
\toprule
\textbf{Parameter} & \textbf{Values} \\
\midrule
Algorithm & DAPO, GRPO \\
\addlinespace[2pt]
Reward & R0, R1, R2 defined in Eqs.~\ref{eq:r0}--\ref{eq:r2} \\
\addlinespace[2pt]
Prompt & P1, P2, P3 defined in Section~\ref{sec:prompts} \\
\addlinespace[2pt]
Backbone & DeepSeek-R1-Distill-Qwen-7B; Qwen2.5-7B-Instruct; Qwen2.5-Math-7B; Llama-3.1-8B-Instruct \\
\addlinespace[2pt]
Tolerance $\delta$ (min) & \{5, 10, 30, 60, 120\} in Eq.~\ref{eq:r2} \\
\addlinespace[2pt]
Steepness $\alpha$ & \{0.5, 1, 2\} in Eq.~\ref{eq:r2} \\
\bottomrule
\end{tabularx}
\end{table*}

\subsection{Results}\label{sec:exp-results}

Our primary metric is accuracy within a tolerance band, $\text{Acc}@\delta$ as defined in Eq.~\ref{eq:acc_c}. We report $\text{Acc}@\delta$ at $\delta\in\{5,10,30,60,120\}$ minutes. By construction, $\text{Acc}@\delta$ is non-decreasing in $\delta$ and thus becomes less discriminative for large $\delta$. 
We also report $\text{Soft Acc}@\delta$ (Eq.~\ref{eq:soft_c}), which linearly decreases from the band boundary to the ground truth; while less intuitive than $\text{Acc}@\delta$, it is more informative when $\text{Acc}@\delta$ values are close. Further, we include mean absolute error (MAE) and mean squared error (MSE) as well.

\begin{figure}[t]
\centering
\includegraphics[width=1\linewidth]{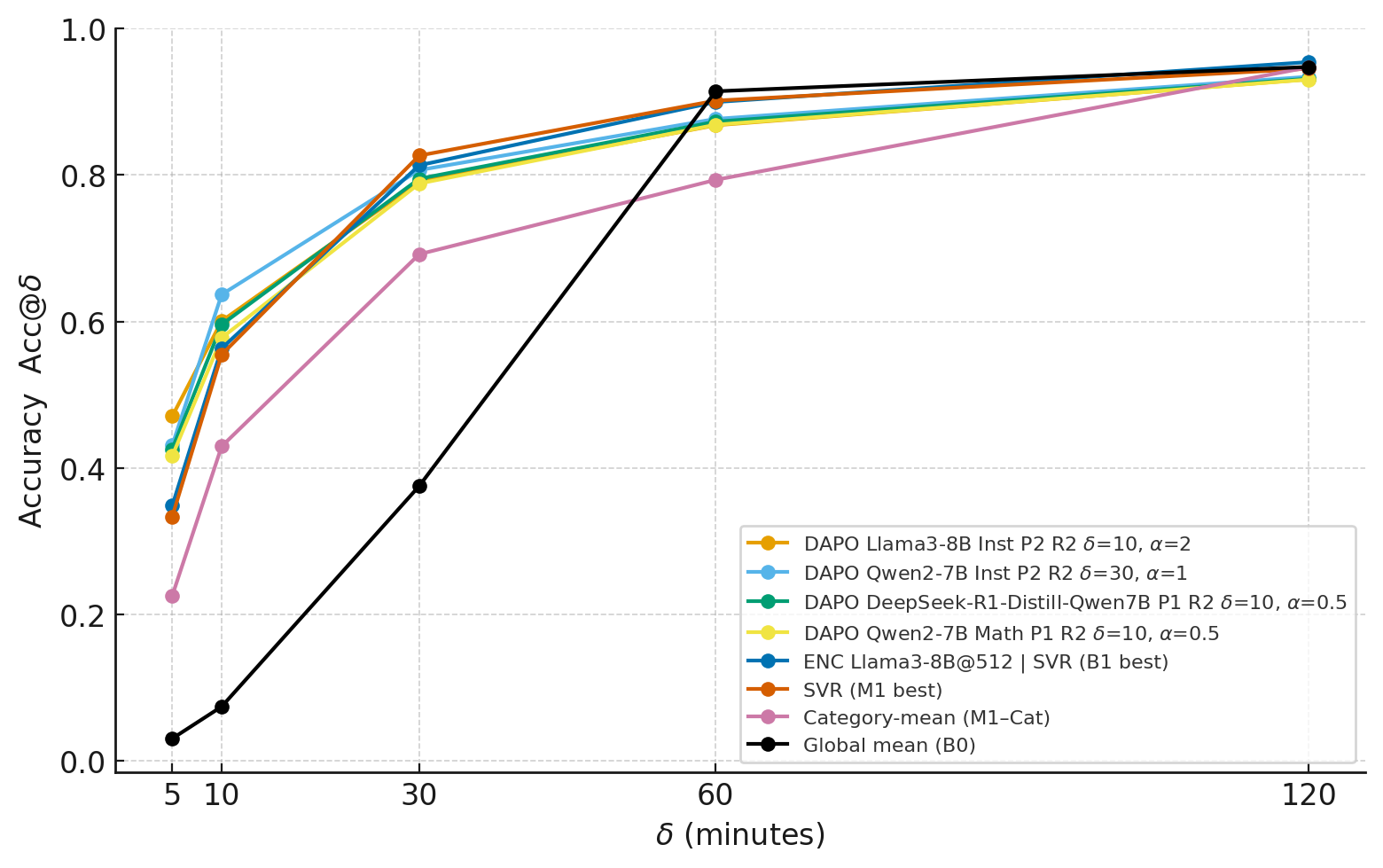}
\caption{The Acc@$\delta$ curves for methods summarized in Table~\ref{tab:methods}.}
\label{fig:main_results}
\end{figure}

Figure~\ref{fig:main_results} plots $\text{Acc}@\delta$ across $\delta$. We select, for each family, strong representatives: (i) the best $\text{Acc}@5$ run per model for RLVR and (ii) the strongest baselines. For RLVR models, the legend annotates the $(\delta,\alpha)$ if the reward R2 is used, which is distinct from the \emph{evaluation} $\delta$ on the $x$–axis. 

At tight bands ($\delta\le10$), RLVR dominates. The best RLVR run attains $\text{Acc}@5=\mathbf{0.471}$ (DAPO Llama3-8B Inst), compared to $0.349$ for the frozen-encoder+SVR baseline and $0.334$ for SVR (Table~\ref{tab:main-results-summary}). At $\delta=10$, the best RLVR reaches $\text{Acc}@10=\mathbf{0.637}$ (DAPO Qwen2-7B Inst), outperforming encoder+SVR at $0.563$ and SVR at $0.555$. These gaps indicate that under strict evaluation bands, RLVR extracts finer structure from the alert text than classical regressors. As $\delta$ widens, the method curves converge: by $\delta=30$, SVR slightly leads ($\text{Acc}@30=\mathbf{0.826}$) with RLVR close behind ($0.807$ at best). At hour-scale bands, differences are small for all learned methods, and even simple priors approach the top: at $\delta=120$, encoder+SVR is best ($\text{Acc}@120=\mathbf{0.954}$), while category means and the global mean both reach $0.947$.

Table~\ref{tab:main-results-summary} reports the full panel of metrics for the Fig.~\ref{fig:main_results} runs. The advantage for RLVR on tight-$\delta$ also appears in soft accuracy: the strongest RLVR achieves $\text{Soft Acc}@5=\mathbf{0.219}$ (vs.\ $0.183$ encoder+SVR; $0.168$ SVR) and $\text{Soft Acc}@10=\mathbf{0.377}$ (vs.\ $0.324$; $0.314$). In the mid-band, RLVR retains the best $\text{Soft Acc}@30=\mathbf{0.614}$ (DAPO Qwen2-7B Inst), whereas SVR leads $\text{Soft Acc}@60=\mathbf{0.733}$.

For MAE and MSE, classical regressors are preferable. The frozen-encoder+SVR baseline yields the lowest errors (MAE $\mathbf{30.93}$, MSE $\mathbf{13173.94}$), followed by SVR (MAE $35.68$, MSE $17724.31$). In comparison, all RLVR runs have higher errors (MAE $\approx38$–$40$ and MSE $\approx1.88$–$1.96\times10^4$). The metrics also reveal different error profiles: the category-mean prior, for example, shows relatively low MSE ($15369.64$) despite a high MAE ($41.72$). This suggests it has a broad average deviation but avoids the extreme misses that MSE penalizes heavily. Conversely, the high MSE of the RLVR runs implies they produce occasional large errors, even while achieving superior small-band accuracy.

The evaluation tolerance ($\delta$) reveals a clear performance trade-off. 
At strict evaluation tolerances ($\delta\le10$), RLVR (DAPO) outperforms classical baselines on both hard and soft accuracy. As tolerance widens ($\delta\ge30$), the accuracy curves cluster, and SVR becomes competitive or best on accuracy while clearly minimizing MAE/MSE. 
This suggests a practical split: for coarse service-level reporting (e.g., within an hour), classical regressors suffice; however, for early, tight-band decisions, RLVR provides the largest gains.

\begin{table}[!htbp]
\centering
\small
\setlength{\tabcolsep}{4pt}
\renewcommand{\arraystretch}{1.15}
\caption{Main results summary for runs used in Fig.~\ref{fig:main_results}. Best values are \textbf{bold} (min for MAE/MSE; max otherwise).}
\label{tab:main-results-summary}
% v-- Add this resizebox wrapper --v
\resizebox{\textwidth}{!}{%
\begin{tabular}{l r r r r r r r r r r r r}
\toprule
\textbf{Run} & \textbf{MAE} $\downarrow$ & \textbf{MSE} $\downarrow$ & \textbf{Acc@5}  & \textbf{Acc@10} & \textbf{Acc@30} & \textbf{Acc@60} & \textbf{Acc@120} & \textbf{Soft@5} & \textbf{Soft@10} & \textbf{Soft@30} & \textbf{Soft@60} & \textbf{Soft@120} \\
\midrule
DAPO Llama3-8B Inst P2 R2 $\delta$=10, $\alpha$=2 & 39.85 & 19443.43 & \textbf{0.471} & 0.601 & 0.791 & 0.868 & 0.931 & \textbf{0.218} & \textbf{0.377} & 0.600 & 0.716 & 0.810 \\
DAPO Qwen2-7B Inst P2 R2 $\delta$=30, $\alpha$=1 & 38.48 & 19110.25 & 0.431 & \textbf{0.637} & 0.807 & 0.876 & 0.934 & 0.193 & 0.364 & \textbf{0.614} & 0.728 & 0.819 \\
DAPO DeepSeek-R1-Distill-Qwen7B P1 R2 $\delta$=10, $\alpha$=0.5 & 38.43 & 18832.81 & 0.426 & 0.596 & 0.795 & 0.873 & 0.931 & 0.207 & 0.358 & 0.594 & 0.715 & 0.811 \\
DAPO Qwen2-7B Math P1 R2 $\delta$=10, $\alpha$=0.5 & 39.95 & 19637.65 & 0.417 & 0.578 & 0.788 & 0.869 & 0.930 & \textbf{0.219} & 0.356 & 0.585 & 0.709 & 0.807 \\
ENC Llama3-8B@512 \textbar{} SVR (B1 best) & \textbf{30.93} & \textbf{13173.94} & 0.349 & 0.563 & 0.813 & 0.899 & \textbf{0.954} & 0.183 & 0.324 & 0.592 & 0.729 & \textbf{0.830} \\
SVR (M1 best) & 35.68 & 17724.31 & 0.334 & 0.555 & \textbf{0.826} & 0.901 & 0.945 & 0.168 & 0.314 & 0.597 & \textbf{0.733} & 0.830 \\
Category-mean (M1--Cat) & 41.72 & 15369.64 & 0.226 & 0.430 & 0.692 & 0.793 & 0.947 & 0.114 & 0.226 & 0.483 & 0.616 & 0.755 \\
Global mean (B0) & 50.59 & 17630.94 & 0.031 & 0.074 & 0.376 & \textbf{0.914} & 0.947 & 0.016 & 0.034 & 0.143 & 0.467 & 0.700 \\
\bottomrule
\end{tabular}
} 
\end{table}

\subsection{Ablations}
To understand the contribution of different components in our framework, we conduct a series of ablation studies along three key directions: (1) the choice of the base model and RLVR algorithm, (2) the hyperparameters of our designed reward, and (3) the desgin of prompts and reward signals. We present the results in Table~\ref{tab:ablation-stacked}.

\begin{table}[!htbp]
\centering
\small
\setlength{\tabcolsep}{4pt}
\renewcommand{\arraystretch}{1.15}
\caption{Ablations along three directions. In each block, the first row shows absolute metrics; subsequent rows are relative ($\Delta$) \% changes vs the first row, and the left cell lists only the differing factor(s).}
\label{tab:ablation-stacked}
\resizebox{\textwidth}{!}{%
\begin{tabular}{l r r r r r r r r r r r r}
\toprule
\textbf{Run / Change} & \textbf{MAE} $\downarrow$ & \textbf{MSE} $\downarrow$ & \textbf{Acc@5} & \textbf{Acc@10} & \textbf{Acc@30} & \textbf{Acc@60} & \textbf{Acc@120} & \textbf{Soft@5} & \textbf{Soft@10} & \textbf{Soft@30} & \textbf{Soft@60} & \textbf{Soft@120} \\
\midrule
\multicolumn{13}{l}{\textit{Ablation: model and algorithm}} \\
\midrule
DAPO Qwen2-7B Inst P2 R2 $\delta$=30, $\alpha$=1 & 38.48 & 19110.93 & 0.431 & 0.637 & 0.807 & 0.876 & 0.934 & 0.193 & 0.364 & 0.614 & 0.728 & 0.819 \\
\midrule
\multicolumn{13}{c}{\emph{$\Delta$ metrics (relative \%)}} \\
\midrule
GRPO & $-0.7\%$ & $-0.3\%$ & $-2.4\%$ & $+0.3\%$ & $+0.2\%$ & $+0.3\%$ & $+0.0\%$ & $+3.9\%$ & $+0.4\%$ & $+0.3\%$ & $+0.3\%$ & $+0.2\%$ \\
Llama3-8B Inst & $-0.7\%$ & $-0.5\%$ & $-3.4\%$ & $+0.6\%$ & $+0.3\%$ & $+0.1\%$ & $+0.1\%$ & $-1.0\%$ & $-1.4\%$ & $+0.2\%$ & $+0.3\%$ & $+0.2\%$ \\
GRPO + Llama3-8B Inst & $-0.9\%$ & $-0.5\%$ & $-4.4\%$ & $+1.7\%$ & $+0.4\%$ & $+0.1\%$ & $-0.1\%$ & $-4.9\%$ & $-1.6\%$ & $+0.4\%$ & $+0.4\%$ & $+0.2\%$ \\
GRPO + DeepSeek-R1-Distill-Qwen7B & $+1.0\%$ & $-0.1\%$ & $-4.8\%$ & $-1.4\%$ & $-0.5\%$ & $-0.1\%$ & $-0.1\%$ & $-1.6\%$ & $-3.4\%$ & $-1.5\%$ & $-0.7\%$ & $-0.4\%$ \\
DeepSeek-R1-Distill-Qwen7B & $+1.1\%$ & $-0.0\%$ & $-4.9\%$ & $-2.3\%$ & $-0.7\%$ & $-0.1\%$ & $-0.1\%$ & $-1.1\%$ & $-3.4\%$ & $-1.7\%$ & $-0.8\%$ & $-0.4\%$ \\
Qwen2-7B Math & $+1.5\%$ & $+0.4\%$ & $-8.5\%$ & $-3.4\%$ & $-1.0\%$ & $-0.2\%$ & $-0.0\%$ & $-3.7\%$ & $-5.7\%$ & $-2.4\%$ & $-1.3\%$ & $-0.6\%$ \\
GRPO + Qwen2-7B Math & $+1.6\%$ & $-0.0\%$ & $-9.2\%$ & $-3.9\%$ & $-0.7\%$ & $-0.2\%$ & $-0.0\%$ & $-4.2\%$ & $-6.4\%$ & $-2.5\%$ & $-1.2\%$ & $-0.6\%$ \\
\midrule\midrule
\multicolumn{13}{l}{\textit{Ablation: hyper parameters of designed reward}} \\
\midrule
DAPO Llama3-8B Inst P2 R2 $\delta$=10, $\alpha$=2.0 & 39.85 & 19443.43 & 0.471 & 0.601 & 0.791 & 0.868 & 0.931 & 0.218 & 0.377 & 0.600 & 0.716 & 0.810 \\
\midrule
\multicolumn{13}{c}{\emph{$\Delta$ metrics (relative \%)}} \\
\midrule
Qwen2-7B Inst                                & $-3.5\%$ & $-1.7\%$ & $-1.6\%$ & $+6.0\%$ & $+2.0\%$ & $+1.0\%$ & $+0.3\%$ & $-11.6\%$ & $-3.5\%$ & $+2.3\%$ & $+1.7\%$ & $+1.1\%$ \\
$\alpha=1.0$                                 & $-3.9\%$ & $-1.9\%$ & $-3.7\%$ & $+5.0\%$ & $+1.5\%$ & $-0.1\%$ & $-0.1\%$ & $-12.2\%$ & $-4.8\%$ & $+2.6\%$ & $+2.0\%$ & $+1.2\%$ \\
$\alpha=0.5$                                 & $-9.2\%$ & $-1.8\%$ & $-7.4\%$ & $+3.2\%$ & $-0.8\%$ & $-2.1\%$ & $-0.2\%$ & $-23.2\%$ & $-16.5\%$ & $0.0\%$  & $+0.5\%$ & $-0.1\%$ \\
$\delta=60, \ \alpha=0.5$                    & $-9.2\%$ & $-1.9\%$ & $-7.8\%$ & $+2.1\%$ & $-0.9\%$ & $-2.2\%$ & $-0.1\%$ & $-23.6\%$ & $-16.8\%$ & $-0.7\%$ & $-0.2\%$ & $-0.3\%$ \\
Qwen2-7B Inst $+\ \delta=30, \ \alpha=1.0$   & $-3.4\%$ & $-1.7\%$ & $-8.7\%$ & $+5.8\%$ & $+2.0\%$ & $+1.0\%$ & $+0.3\%$ & $-11.3\%$ & $-3.6\%$ & $+2.3\%$ & $+1.8\%$ & $+1.1\%$ \\
DeepSeek\textendash R1\textendash Distill\textendash Qwen7B $+\ \alpha=0.5$ & $-3.6\%$ & $-3.1\%$ & $-9.5\%$ & $-0.8\%$ & $+0.5\%$ & $+0.0\%$ & $+0.1\%$ & $-5.1\%$  & $-5.0\%$ & $-1.0\%$ & $-0.1\%$ & $+0.1\%$ \\
$\delta=30$                                  & $-2.9\%$ & $-1.9\%$ & $-11.3\%$ & $+3.5\%$ & $+0.8\%$ & $-0.0\%$ & $-0.1\%$ & $-12.4\%$ & $-4.7\%$ & $+2.6\%$ & $+1.9\%$ & $+1.2\%$ \\
$\delta=30, \ \alpha=1.0$                    & $-4.1\%$ & $-1.9\%$ & $-11.6\%$ & $+5.0\%$ & $+2.0\%$ & $+0.8\%$ & $+0.2\%$ & $-12.8\%$ & $-4.8\%$ & $+2.7\%$ & $+2.0\%$ & $+1.2\%$ \\
\midrule\midrule
\multicolumn{13}{l}{\textit{Ablation: prompts and rewards}} \\
\midrule
DAPO Qwen2-7B Inst P2 R2 $\delta$=30, $\alpha$=1.0 & 38.48 & 19110.25 & 0.431 & 0.637 & 0.807 & 0.876 & 0.934 & 0.193 & 0.364 & 0.614 & 0.728 & 0.819 \\
\midrule
\multicolumn{13}{c}{\emph{$\Delta$ metrics (relative \%)}} \\
\midrule
GRPO & $-0.8\%$ & $-1.3\%$ & $+0.2\%$ & $-0.2\%$ & $-0.1\%$ & $+0.1\%$ & $-0.1\%$ & $+4.1\%$ & $-0.2\%$ & $+0.4\%$ & $+0.4\%$ & $+0.1\%$ \\
P1 & $-2.8\%$ & $-1.5\%$ & $-1.5\%$ & $-0.7\%$ & $+1.3\%$ & $+0.9\%$ & $+0.3\%$ & $+0.9\%$ & $-0.4\%$ & $+0.8\%$ & $+1.1\%$ & $+0.8\%$ \\
P3 & $-11.0\%$ & $-17.0\%$ & $-8.9\%$ & $-7.1\%$ & $+0.6\%$ & $+2.8\%$ & $+1.2\%$ & $-5.9\%$ & $-8.2\%$ & $-3.4\%$ & $+0.1\%$ & $+1.2\%$ \\
DeepSeek\textendash R1\textendash Distill\textendash Qwen7B & $-0.5\%$ & $-0.1\%$ & $-0.9\%$ & $-1.4\%$ & $-0.5\%$ & $-0.1\%$ & $-0.1\%$ & $-1.6\%$ & $-3.4\%$ & $-1.5\%$ & $-0.6\%$ & $-0.4\%$ \\
MAE & $+2.3\%$ & $+1.8\%$ & $-2.9\%$ & $-4.1\%$ & $-1.0\%$ & $-0.3\%$ & $-0.1\%$ & $-4.8\%$ & $-3.9\%$ & $-1.6\%$ & $-1.1\%$ & $-0.4\%$ \\
R1 & $+5.4\%$ & $+2.5\%$ & $-10.6\%$ & $-14.0\%$ & $-4.3\%$ & $-0.4\%$ & $+0.3\%$ & $-19.7\%$ & $-14.3\%$ & $-5.5\%$ & $-2.1\%$ & $-0.7\%$ \\
DeepSeek\textendash R1\textendash Distill\textendash Qwen7B + P1 & $+0.5\%$ & $-0.1\%$ & $-10.2\%$ & $-3.8\%$ & $-0.7\%$ & $-0.2\%$ & $-0.3\%$ & $-4.4\%$ & $-6.8\%$ & $-2.5\%$ & $-1.2\%$ & $-0.7\%$ \\
DeepSeek\textendash R1\textendash Distill\textendash Qwen7B + P3 & $-5.1\%$ & $-8.5\%$ & $-11.2\%$ & $-8.0\%$ & $+0.4\%$ & $+2.1\%$ & $+0.6\%$ & $-9.5\%$ & $-10.3\%$ & $-4.0\%$ & $-0.4\%$ & $+0.6\%$ \\
GRPO + DeepSeek\textendash R1\textendash Distill\textendash Qwen7B & $+1.0\%$ & $-0.1\%$ & $-4.8\%$ & $-2.3\%$ & $-0.4\%$ & $-0.1\%$ & $-0.1\%$ & $-1.6\%$ & $-3.4\%$ & $-1.5\%$ & $-0.7\%$ & $-0.4\%$ \\
GRPO + MAE & $+2.0\%$ & $+2.0\%$ & $-3.1\%$ & $-4.1\%$ & $-1.1\%$ & $-0.3\%$ & $-0.2\%$ & $-4.4\%$ & $-3.9\%$ & $-1.7\%$ & $-1.0\%$ & $-0.5\%$ \\
GRPO + R1 & $+5.6\%$ & $+2.6\%$ & $-13.2\%$ & $-14.8\%$ & $-4.5\%$ & $-0.5\%$ & $+0.3\%$ & $-20.2\%$ & $-15.0\%$ & $-5.5\%$ & $-2.3\%$ & $-0.8\%$ \\
DeepSeek\textendash R1\textendash Distill\textendash Qwen7B + MAE & $+2.6\%$ & $+2.3\%$ & $-2.7\%$ & $-4.1\%$ & $-0.9\%$ & $-0.2\%$ & $-0.1\%$ & $-3.9\%$ & $-4.0\%$ & $-1.8\%$ & $-1.1\%$ & $-0.6\%$ \\
\bottomrule
\end{tabular}
}
\end{table}

As discussed in \S\ref{sec:exp-results}, our primary evaluation metric is Acc@5, and Soft@5 is used as a secondary indicator, especially when Acc@5 scores are close. In each ablation block, the first row represents the configuration that achieved the highest Acc@5 accuracy within that specific comparison group.

\textbf{Ablation on Model and Algorithm}. In the first block, we compare Qwen2.5-7B-Instruct trained with DAPO, against several alternatives. Substituting Qwen2.5-7B-Inst with Llama3-8B Inst leads to a relative drop of $-3.4\%$ in Acc@5; and the instruction-following version outperforms Qwen2-7B Math ($-8.5\%$ Acc@5), and DeepSeek-R1-Distill-QWen ($-4.9\%$ Acc@5). And DAPO algorithm shows a slight advantage over GRPO ($-2.4\%$ Acc@5).

The results provide a clear answer to our research question: the skills for formal, logical problem-solving do not directly transfer to our real-world prediction task. Instead, the task benefits more from the robust, general-purpose instruction-following and nuanced natural language understanding capabilities of the Inst model, for both QWen and Llama. They appear better equipped to handle the "noisy" and "uncertain" nature of transit event descriptions.

\textbf{Ablation on Reward Hyperparameters}. The second block investigates the sensitivity to the reward hyperparameters $\delta$ and $\alpha$. A trade-off is observed between regression error (MAE) and accuracy (Acc@$\delta$). The baseline setting of $\delta=10, \alpha=2.0$ achieves the highest Acc@5 (0.471). Regarding the impact of $\alpha$, we found that decreasing $\alpha$ (e.g., to 0.5) improves MAE (a $-9.2\%$ relative change), but this comes at the cost to our primary objective, where Acc@5 drops by $-7.4\%$ and Soft@5 plummets by $-23.2\%$. For $\delta$ used in training, similarly, increasing $\delta$ from 10 to 30 improves MAE ($-2.9\%$) but severely degrades Acc@5 ($-11.3\%$). This demonstrates that there is some degree of freedom to optimize different objectives by varying the two hyperparameters. When the goal is to maximize the tolerance based accuracy, a smaller $\delta$ and and larger $\alpha$ helps.

\textbf{Ablation on Prompts and Rewards}. For the third block, we validate our design choices for the prompt and reward signal, comparing the instance trained with prompt P2 and reward R2 against alternatives. The ablation reveals that our R2 reward design is the most critical component for the success. Changing to the R1 reward is catastrophic, causing a $-10.6\%$ relative drop in Acc@5 and a $-14.0\%$ drop in Acc@10. Using a simple MAE (R0) reward signal is also suboptimal, degrading Acc@5 by $-2.9\%$, Acc@10 by $-4.1\%$. 
Prompt design is also important. Our P2 prompt is shown to be the most effective. Using the P1 prompt results in a minor performance drop ($-1.5\%$ Acc@5), but the P3 prompt is significantly worse, causing a large $-8.9\%$ drop in Acc@5.

In conclusion, our ablation studies highlight three critical design choices for our framework's success. The R2 reward formulation is identified as the single most critical component, with the R1 alternative causing a catastrophic performance drop ($-10.6\%$ Acc@5). Second, the choice of a general-purpose Inst model is crucial, as the specialized Math model failed to handle the task's noisy, real-world nature. Third, the P2 prompt design proved significantly better than its P3 counterpart. Finally, we showed that the $\delta$ and $\alpha$ hyperparameters are essential for correctly tuning the trade-off between regression error and our primary tolerance-based accuracy metrics.

\textbf{Ablation on M1 Baselines}. We present the M1 baseline ablations in Table.~\ref{tab:ablation-m1-delta}, comparing various regression models against the naive Global mean (B0) baseline. As expected, the B0 model performs poorly (0.031 Acc@5). Simply predicting the Category-mean (M1--Cat) is substantially better, improving Acc@5 by +632.3\%. The SVR model performed best, achieving the highest Acc@5 with a +983.1\% relative improvement over the global mean. The K-Neighbors Regressor also showed a strong improvement (+693.8\%). Notably, many other models (e.g., Random Forest, Linear Regression) performed identically to the M1--Cat baseline, indicating they failed to learn a more complex pattern, and simply predicting the category average.

\begin{table}[!htbp]
\centering
\small
\setlength{\tabcolsep}{4pt}
\renewcommand{\arraystretch}{1.15}
\caption{Ablation: \textbf{M1 baselines}. First row shows absolute metrics for the Global mean (B0); subsequent rows are relative ($\Delta$) \% changes vs Global mean, ordered by Acc@5 improvement (except the second row, which is the Category-mean (M1--Cat)).}
\label{tab:ablation-m1-delta}
\resizebox{\textwidth}{!}{%
\begin{tabular}{l r r r r r r r r r r r r}
\toprule
\textbf{Run / Change} & \textbf{MAE} $\downarrow$ & \textbf{MSE} $\downarrow$ & \textbf{Acc@5}  & \textbf{Acc@10}  & \textbf{Acc@30}  & \textbf{Acc@60}  & \textbf{Acc@120}  & \textbf{Soft@5}  & \textbf{Soft@10}  & \textbf{Soft@30}  & \textbf{Soft@60}  & \textbf{Soft@120}  \\
\midrule
Global mean (B0) & 50.59 & 17630.94 & 0.031 & 0.074 & 0.376 & 0.914 & 0.947 & 0.016 & 0.034 & 0.143 & 0.467 & 0.700 \\
\midrule
\multicolumn{13}{c}{\emph{$\Delta$ metrics (relative \% vs Global mean)}} \\
\midrule
Category-mean (M1--Cat) & $-17.5\%$ & $-12.8\%$ & $+632.3\%$ & $+477.1\%$ & $+83.9\%$ & $-0.1\%$ & $-0.1\%$ & $+595.7\%$ & $+573.5\%$ & $+236.9\%$ & $+31.9\%$ & $+7.8\%$ \\
SVR & $-29.5\%$ & $+0.5\%$ & $+983.1\%$ & $+645.2\%$ & $+120.1\%$ & $-1.4\%$ & $-0.3\%$ & $+921.8\%$ & $+835.1\%$ & $+316.1\%$ & $+57.0\%$ & $+18.5\%$ \\
K-Neighbors Regressor & $-20.4\%$ & $-3.0\%$ & $+693.8\%$ & $+503.8\%$ & $+95.0\%$ & $+0.3\%$ & $+0.3\%$ & $+586.1\%$ & $+586.2\%$ & $+237.0\%$ & $+38.6\%$ & $+11.8\%$ \\
Random Forest & $-17.5\%$ & $-12.9\%$ & $+632.3\%$ & $+477.1\%$ & $+84.3\%$ & $-0.1\%$ & $-0.1\%$ & $+596.5\%$ & $+574.2\%$ & $+237.1\%$ & $+32.0\%$ & $+7.7\%$ \\
Linear Regression & $-17.5\%$ & $-12.8\%$ & $+632.3\%$ & $+477.1\%$ & $+84.2\%$ & $-0.1\%$ & $-0.1\%$ & $+595.7\%$ & $+573.5\%$ & $+236.9\%$ & $+31.9\%$ & $+7.8\%$ \\
Decision Tree & $-17.5\%$ & $-12.8\%$ & $+632.3\%$ & $+477.1\%$ & $+84.2\%$ & $-0.1\%$ & $-0.1\%$ & $+595.7\%$ & $+573.5\%$ & $+236.9\%$ & $+31.9\%$ & $+7.8\%$ \\
Ridge & $-17.5\%$ & $-12.8\%$ & $+632.3\%$ & $+475.8\%$ & $+84.1\%$ & $-13.4\%$ & $-0.1\%$ & $+593.5\%$ & $+571.7\%$ & $+236.6\%$ & $+31.9\%$ & $+7.7\%$ \\
Gradient Boosting & $-17.3\%$ & $-12.6\%$ & $+558.5\%$ & $+434.4\%$ & $+93.9\%$ & $+0.0\%$ & $+0.3\%$ & $+476.4\%$ & $+484.2\%$ & $+228.3\%$ & $+31.1\%$ & $+7.6\%$ \\
Lasso & $-12.2\%$ & $-8.1\%$ & $+184.6\%$ & $+185.4\%$ & $+85.5\%$ & $-10.4\%$ & $+0.8\%$ & $+175.0\%$ & $+189.8\%$ & $+148.2\%$ & $+21.0\%$ & $+5.3\%$ \\
AdaBoost & $-7.7\%$ & $-8.0\%$ & $+176.9\%$ & $+131.2\%$ & $+94.4\%$ & $-1.0\%$ & $-0.2\%$ & $+140.5\%$ & $+145.4\%$ & $+135.1\%$ & $+17.8\%$ & $+2.4\%$ \\
ElasticNet & $-3.4\%$ & $-2.4\%$ & $+20.0\%$ & $+12.1\%$ & $+23.4\%$ & $+0.4\%$ & $+0.1\%$ & $+16.7\%$ & $+18.5\%$ & $+17.1\%$ & $+4.2\%$ & $+1.6\%$ \\
\bottomrule
\end{tabular}
}
\end{table}

\begin{figure}[!htbp]
\centering % Center all subfigures

% --- Row 1: First Subfigure ---
\begin{subfigure}[b]{0.4\linewidth}
    \centering
    \includegraphics[width=\linewidth]{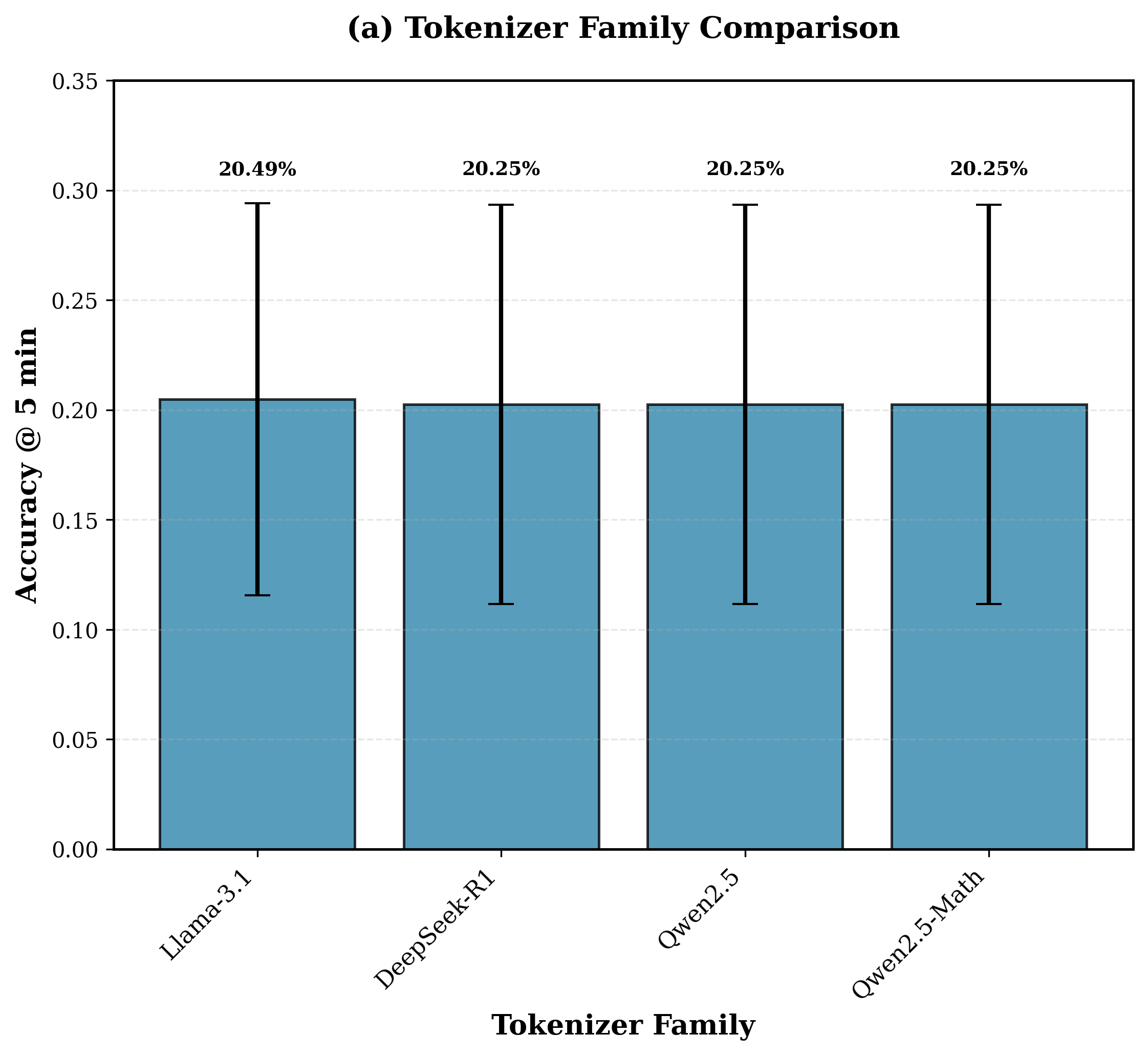}
    \caption{} % You can add a sub-caption here
    \label{fig:b1-ablation-tokenizer}
\end{subfigure}

% --- Row 2: Second and Third Subfigures ---
\begin{subfigure}[b]{0.48\linewidth}
    \centering
    \includegraphics[width=\linewidth]{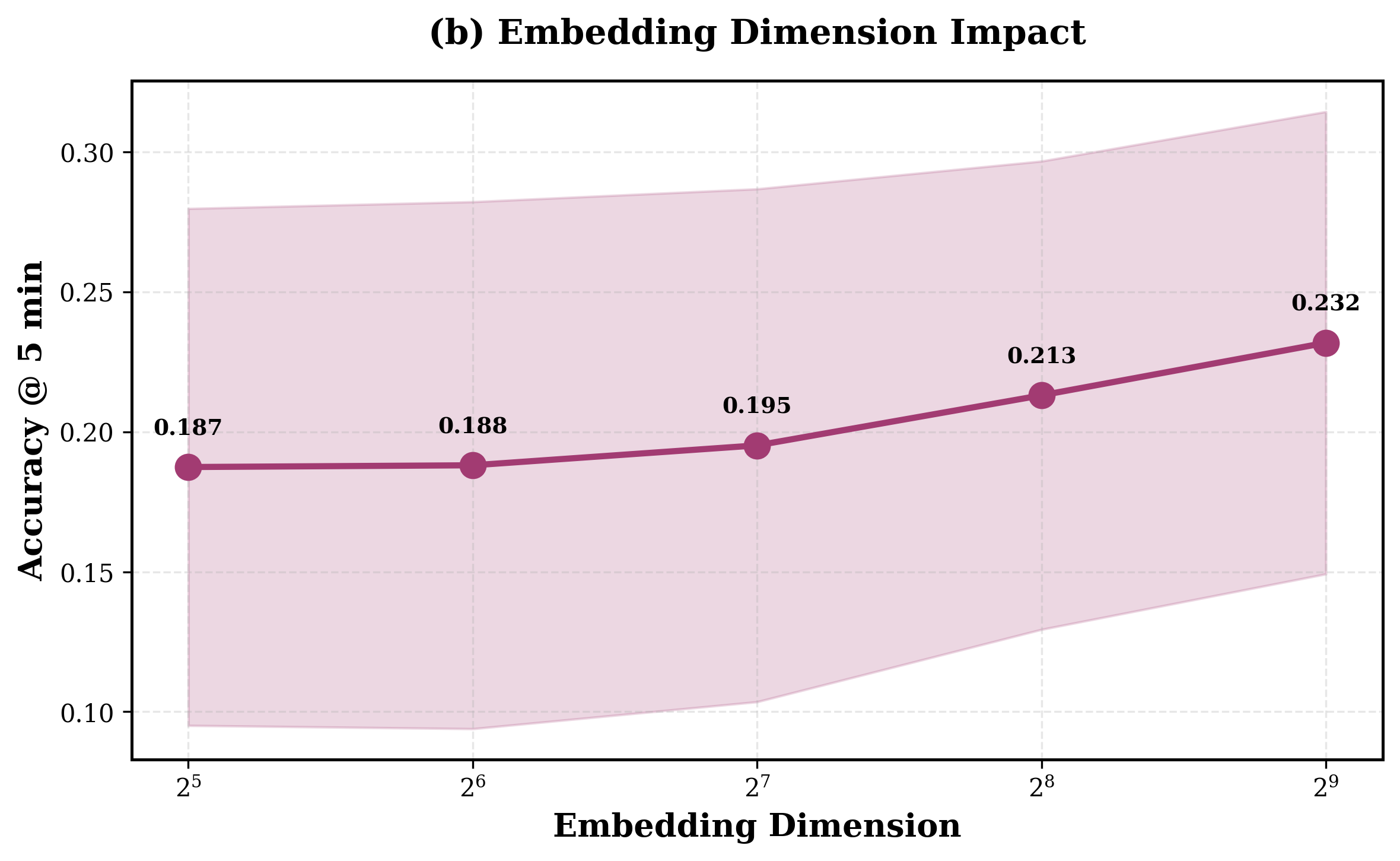}
    \caption{} % You can add a sub-caption here
    \label{fig:b1-ablation-embed_dim}
\end{subfigure}
\hfill % Adds horizontal space between figures
\begin{subfigure}[b]{0.48\linewidth}
    \centering
    \includegraphics[width=\linewidth]{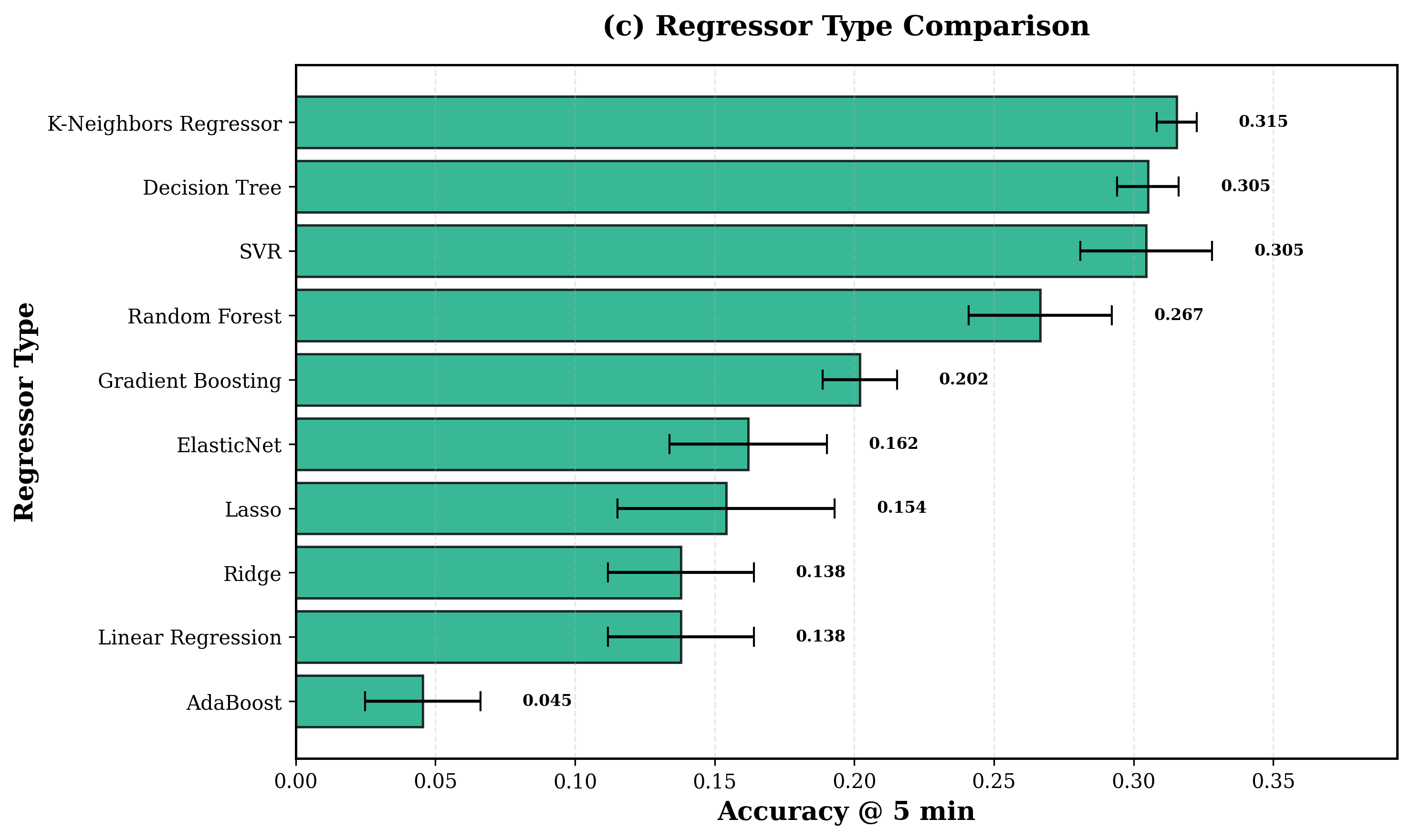}
    \caption{} % You can add a sub-caption here
    \label{fig:b1-ablation-regressor}
\end{subfigure}

% --- Main Caption ---
\caption{Ablation study on the B1 (Tokenizer-hash + regressor) baseline components. (a) Performance is largely insensitive to the tokenizer family. (b) Accuracy steadily improves as the hashing dimension d increases from 32 to 512. (c) The choice of regressor is the most critical factor, with K-Neighbors Regressor and SVR significantly outperforming other models.}
\label{fig:all-ablations} % A new, main label for the whole figure

\end{figure}

\textbf{Ablation on B1 Components}. We ablated the three key components of the B1 (Tokenizer-hash + regressor) baseline, with results shown in Figure~\ref{fig:all-ablations}. We found that the choice of regressor is the most critical factor; (c) shows that the K-Neighbors Regressor ($\text{Acc@5} \approx 0.325$) and SVR ($\text{Acc@5} \approx 0.301$) significantly outperform all other models. For the feature representation, (b) shows a clear trend: a larger hashing dimension d is better, with performance steadily increasing as d rises from 32 to 512. In contrast, (a) demonstrates that the choice of tokenizer family has a negligible impact on performance, with all four families yielding a nearly identical $\text{Acc@5} \approx 0.20$.

% Author by Ruijian, Bowen
\section{Discussion}

\subsection{Analysis}
\subsubsection{Entropy during training}
\label{sec:train-entropy}

\begin{figure}[!htbp]
\centering
\begin{subfigure}{0.48\textwidth}\centering\includegraphics[width=\linewidth]{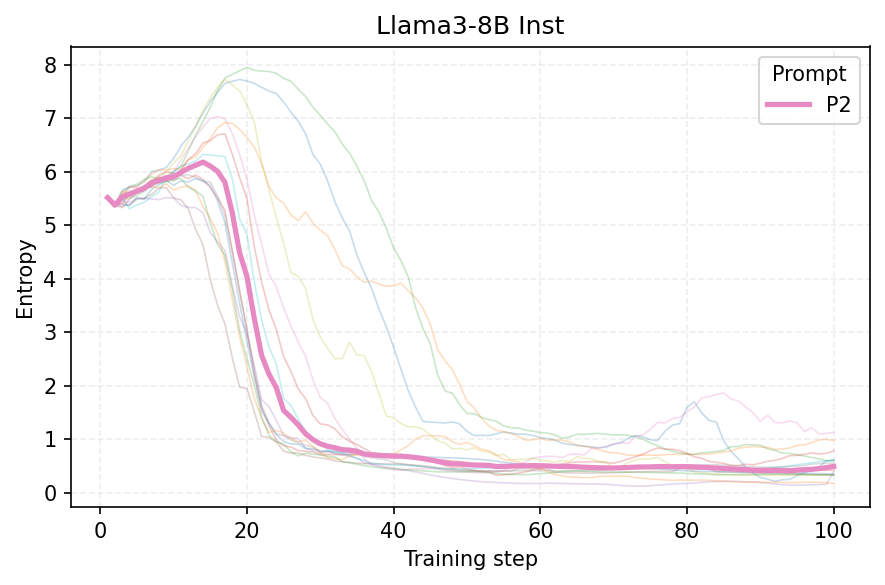}\vspace{2pt}\end{subfigure}
\begin{subfigure}{0.48\textwidth}\centering\includegraphics[width=\linewidth]{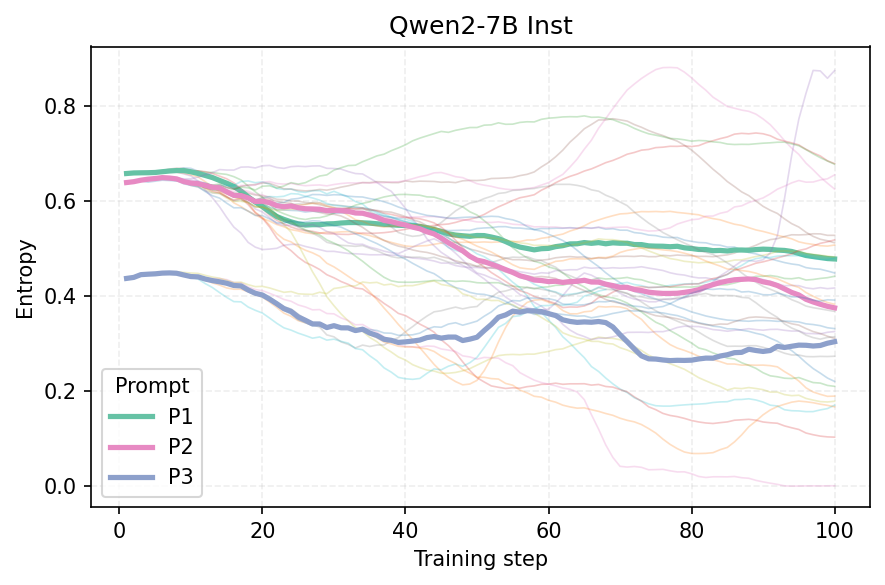}\vspace{2pt}\end{subfigure}\\[6pt]
\begin{subfigure}{0.48\textwidth}\centering\includegraphics[width=\linewidth]{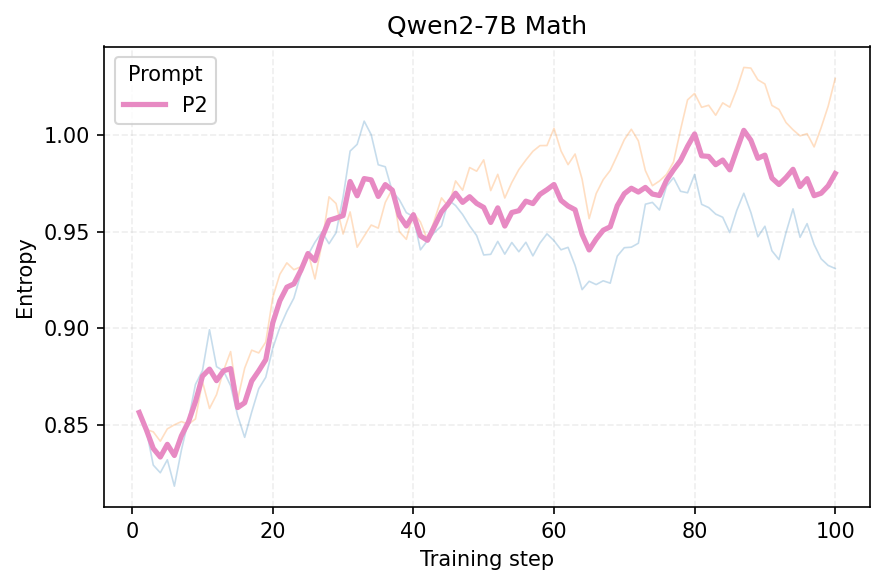}\vspace{2pt}\end{subfigure}
\begin{subfigure}{0.48\textwidth}\centering\includegraphics[width=\linewidth]{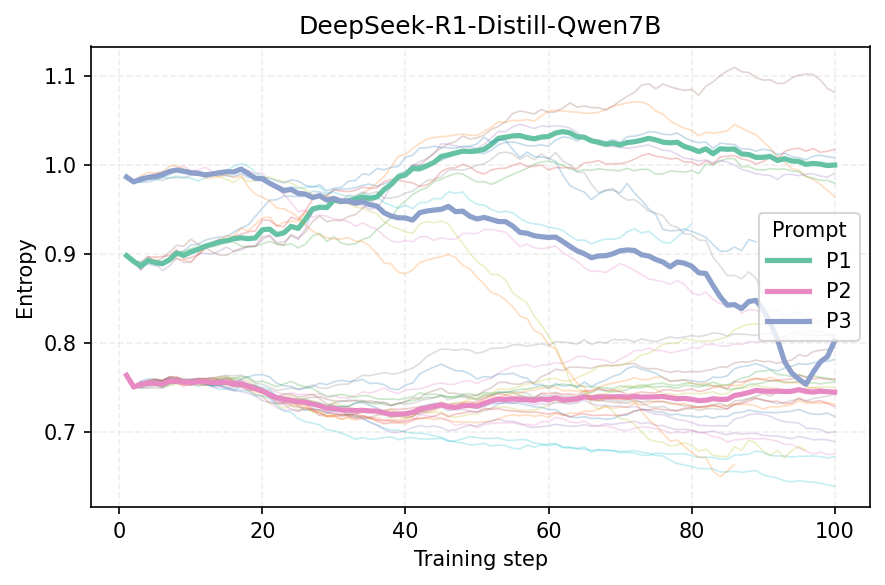}\vspace{2pt}\end{subfigure}
\caption{Training dynamics: predictive entropy vs.\ step. Each panel: thin lines show individual runs; thick lines show per-step medians. Prompts P1/P2/P3 are overlaid.}
\label{fig:train-entropy-2x2}
\end{figure}

We track the per-token predictive entropy of the policy on the training prompts over RLVR training steps, as shown in Figure~\ref{fig:train-entropy-2x2}. We plot separate panels for Llama and Qwen-family models; Qwen panels group three backbones: instruction-tuned (\texttt{Qwen2.5-7B-Inst}), math-specialized (\texttt{Qwen2.5-7B-Math}), and DeepSeek R1 distilled (\texttt{DeepSeek-R1-Distill-Qwen-7B}).

Llama starts substantially higher: most runs begin around \(5\!-\!6\), while Qwen-family models start below \(1\). Within the Qwen family the ordering at initialization is consistent: \texttt{Qwen-Inst} is the lowest entropy, \texttt{Qwen-Math} is higher, and \texttt{DeepSeek-R1-Distill} is the highest.

Llama almost always decays toward low entropy, typically ending below \(1\). Qwen shows mixed movement: roughly one-third of Qwen runs increase entropy; about half of \texttt{DeepSeek-R1-Distill} runs rise above their start, and \texttt{Qwen-Math} runs consistently increase relative to initialization. By contrast, instruction-tuned variants of both Llama and Qwen primarily decrease. 
Across models where \textbf{P3} is available (Qwen-Inst and DeepSeek-R1), we observe a clear entropy drop over steps.

\textbf{Interpretation.} Unlike math/coding, where entropy typically decreases over training, we observe a clear increase on our task for a substantial fraction of runs—especially for math-specialized backbones. RLVR can both concentrate and reopen distributional support depending on backbone and prompt: Llama begins broad and is driven to concentrate as training proceeds; within Qwen, \texttt{Math} and \texttt{R1} (higher than \texttt{Inst} at start but lower than Llama) tend to explore more under \textbf{P1}/\textbf{P2}. In these settings, there is no single dominant reasoning pattern (e.g., consistently higher advantage) and the support does not shrink systematically. By contrast, \textbf{P3} uniformly pushes toward concentration (entropy drops across models).

\textbf{Takeaway.} Entropy dynamics hinge on both specialization and prompt: (i) Llama moves from exploratory to decisive; (ii) within Qwen, \texttt{Inst} stays low or decreases, while \texttt{Math}/\texttt{R1} often increase unless using \textbf{P3}; (iii) \textbf{P3} acts as a strong concentrating signal across models. The richer domain cues in these prompts can potentially induce early over-concentration.

\subsubsection{Response length during training}
\label{sec:train-length}

\begin{figure}[!htbp]
\centering
\begin{subfigure}{0.48\textwidth}\centering\includegraphics[width=\linewidth]{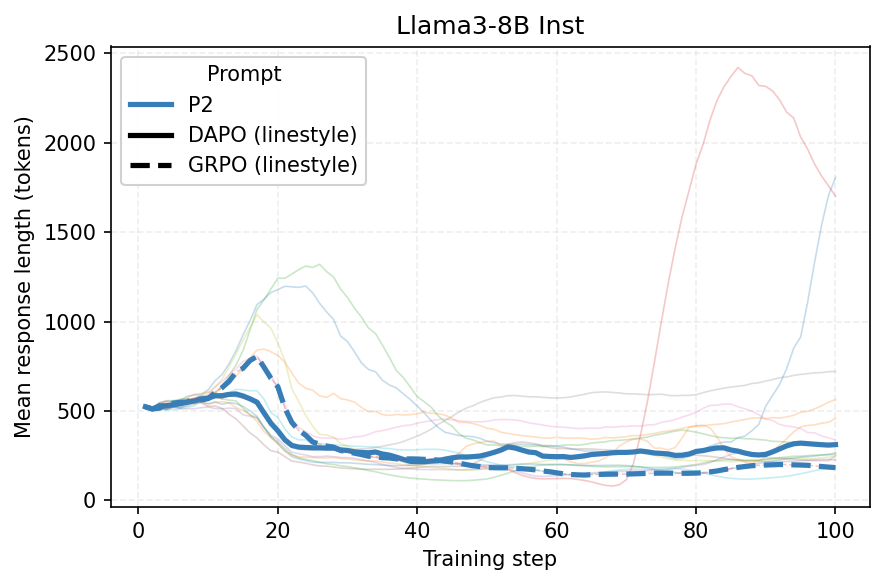}\vspace{2pt}\end{subfigure}
\begin{subfigure}{0.48\textwidth}\centering\includegraphics[width=\linewidth]{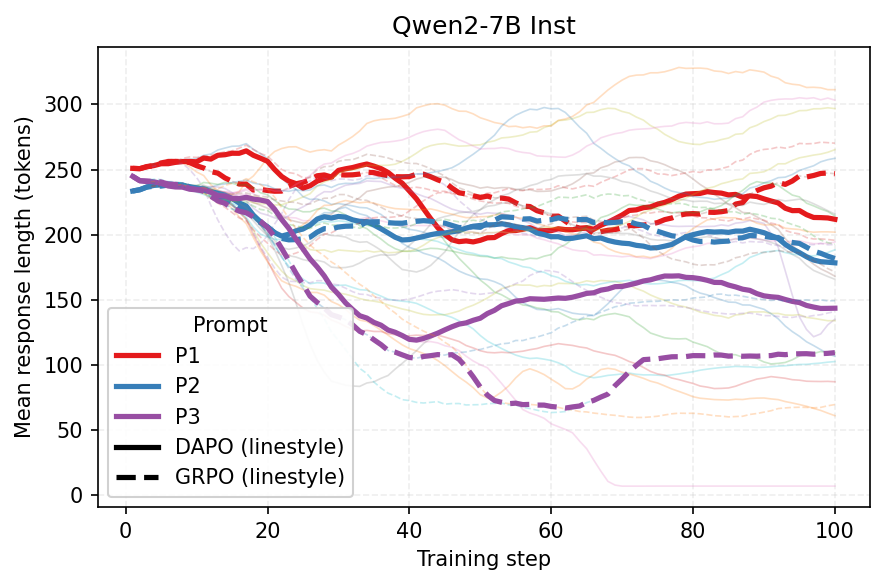}\vspace{2pt}\end{subfigure}\\[6pt]
\begin{subfigure}{0.48\textwidth}\centering\includegraphics[width=\linewidth]{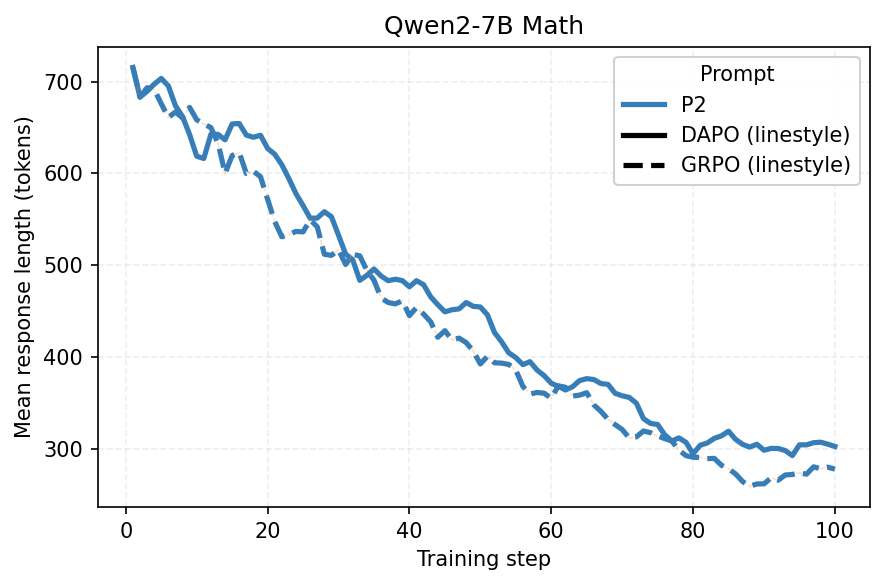}\vspace{2pt}\end{subfigure}
\begin{subfigure}{0.48\textwidth}\centering\includegraphics[width=\linewidth]{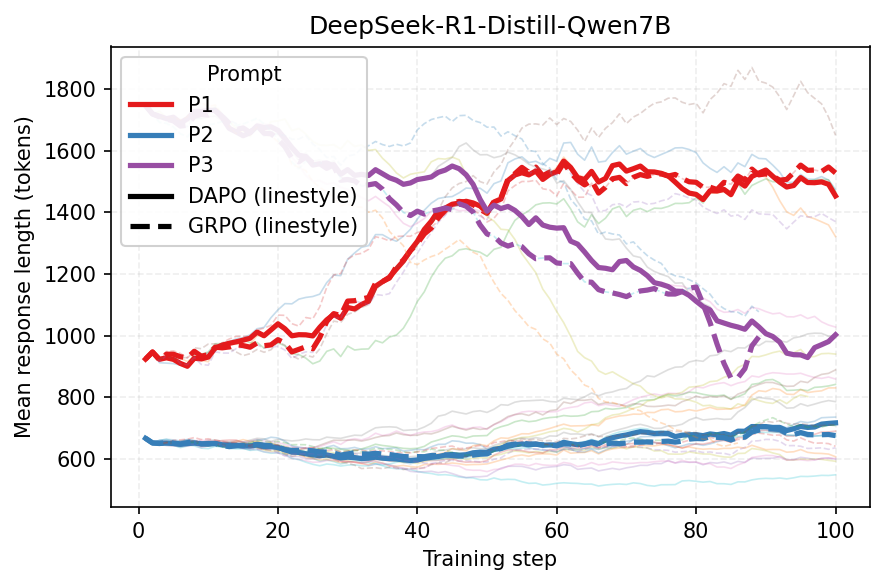}\vspace{2pt}\end{subfigure}
\caption{Training dynamics: mean response length vs.\ step. Thin lines are individual runs; thick lines are per-step medians. Prompts P1/P2/P3 share colors; \emph{linestyle} encodes algo (DAPO solid, GRPO dashed).}
\label{fig:train-resp-len-2x2}
\end{figure}

We report the mean response length (tokens) over RLVR steps in Figure~\ref{fig:train-resp-len-2x2}, stratified by backbone and prompt (\textbf{P1}/\textbf{P2}/\textbf{P3}), with DAPO and GRPO shown separately when behavior differs. Under \textbf{P2}. Initially, at step~0, \texttt{Qwen-Math} is longest at \(\sim720\) tokens, followed by \texttt{DeepSeek-R1-Distill} at \(\sim660\), \texttt{Llama} at \(\sim520\), and \texttt{Qwen-Inst} at \(\sim230\). By the end of training, \texttt{DeepSeek-R1-Distill} becomes the longest (\(\sim600\!-\!900\) tokens), while most other backbones shrink to \(<300\) tokens on average.

\textbf{Prompt effects (\textbf{P1} vs \textbf{P3}).} \texttt{DeepSeek-R1-Distill} departs most from the rest. Under \textbf{P3}, initially \(\sim1700\) tokens on average; converges to \(\sim600\!-\!800\). While under \textbf{P1}, initially \(\sim900\) tokens; increases to \(\sim1400\!-\!1600\) at the end. This \textbf{P1}/\textbf{P3} divergence appears consistently for both \textbf{DAPO} and \textbf{GRPO}. 

\textbf{Interpretation.} \textbf{P3} acts as a concision prior (shorter, more focused completions) even for verbose backbones like \texttt{DeepSeek-R1-Distill}, whereas \textbf{P1} encourages expanded deliberation and tends to lengthen responses over training for that backbone. Under \textbf{P2}, most models learn to compress, but \texttt{DeepSeek-R1-Distill} maintains longer chains, suggesting a backbone–prompt interaction where mathematical reasoning habits resist aggressive shortening unless the prompt explicitly imposes it (\textbf{P3}).

%\textbf{Takeaway.} Choose prompts to control verbosity: \textbf{P3} reliably shortens; \textbf{P1} can lengthen for the \texttt{DeepSeek-R1-Distill} model. For exploratory training on \texttt{DeepSeek-R1-Distill}, \textbf{P1} can be leveraged when longer reasoning is acceptable.

\subsubsection{Accuracy over training steps and time}
\label{sec:train-acc5}

\begin{figure}[!htbp]
\centering
\begin{subfigure}{0.32\textwidth}\centering\includegraphics[width=\linewidth]{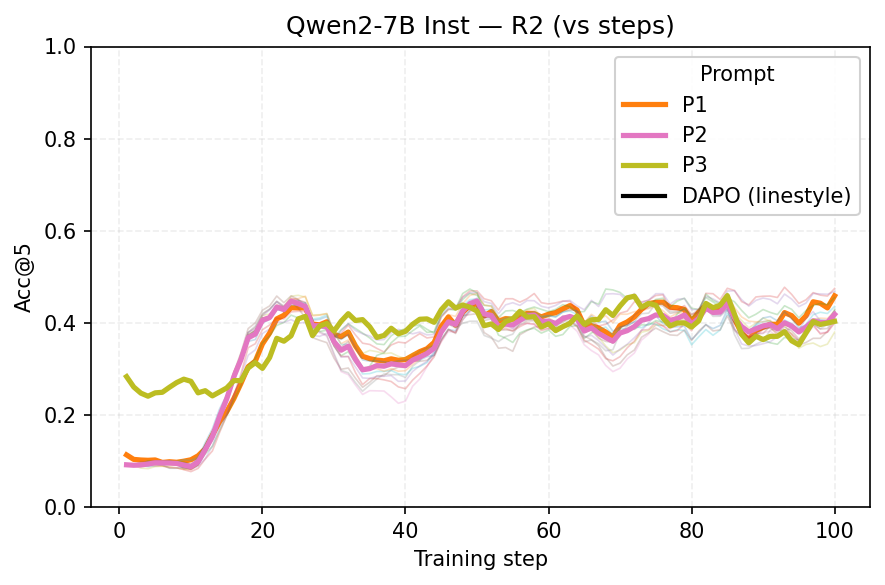}\vspace{2pt}\end{subfigure}
\begin{subfigure}{0.32\textwidth}\centering\includegraphics[width=\linewidth]{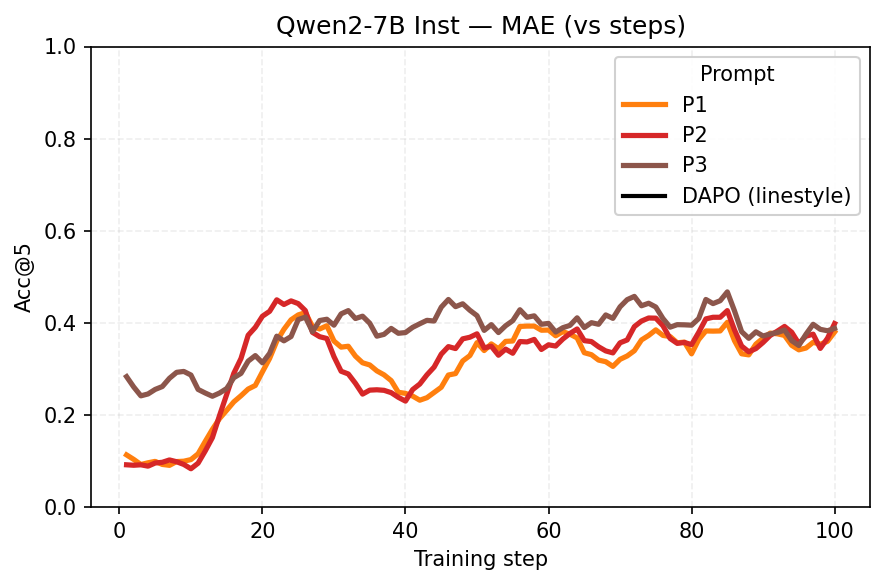}\vspace{2pt}\end{subfigure}
\begin{subfigure}{0.32\textwidth}\centering\includegraphics[width=\linewidth]{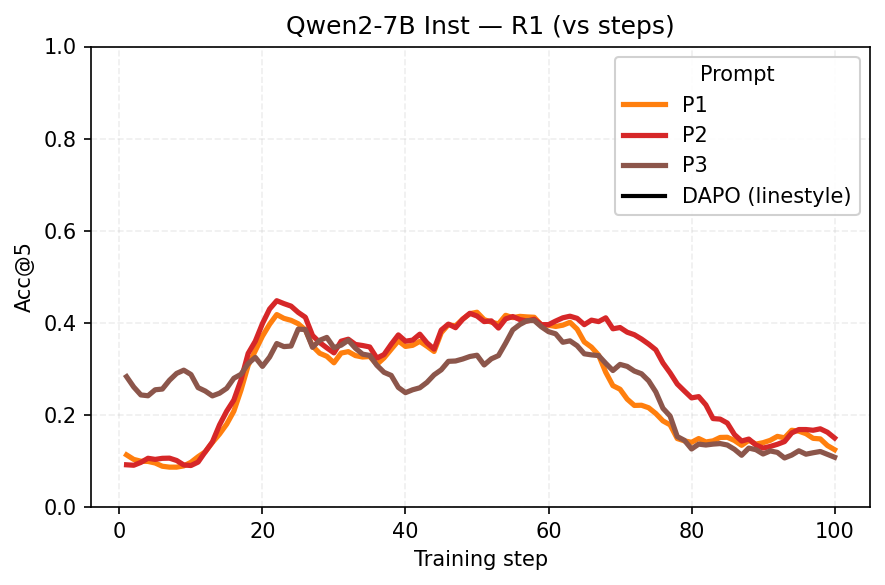}\vspace{2pt}\end{subfigure}\\[6pt]
\begin{subfigure}{0.32\textwidth}\centering\includegraphics[width=\linewidth]{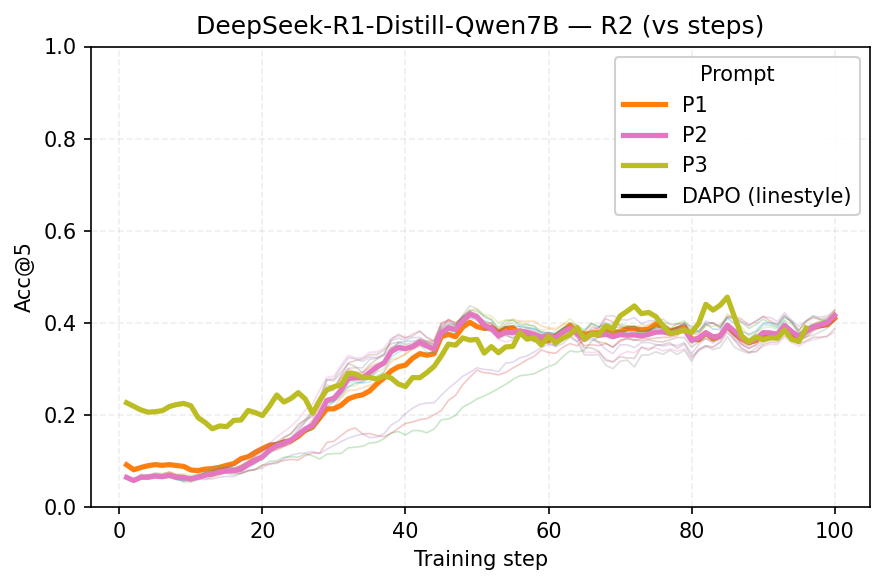}\vspace{2pt}\end{subfigure}
\begin{subfigure}{0.32\textwidth}\centering\includegraphics[width=\linewidth]{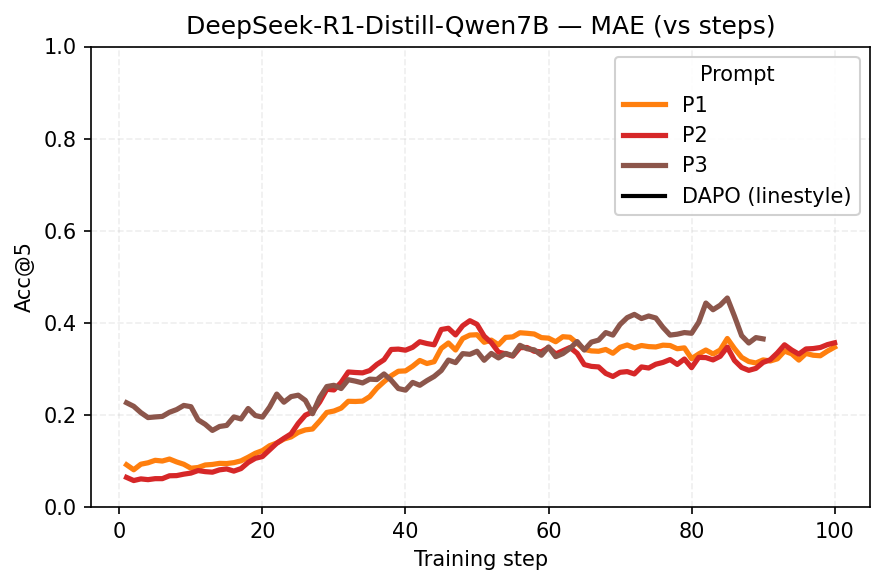}\vspace{2pt}\end{subfigure}
\begin{subfigure}{0.32\textwidth}\centering\includegraphics[width=\linewidth]{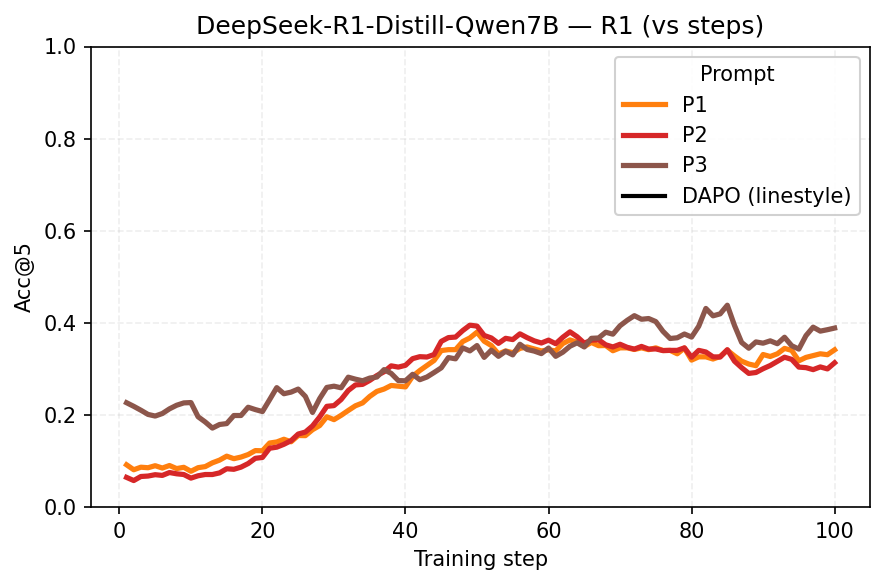}\vspace{2pt}\end{subfigure}
\caption{Acc@5 vs.\ training steps. Each panel shows thin per-run curves and thick per-step medians, colored by prompt (P1/P2/P3). Linestyle encodes algo (DAPO solid, GRPO dashed).}
\label{fig:acc5-steps-domain}
\end{figure}

\begin{figure}[!htbp]
\centering
\begin{subfigure}{0.48\textwidth}\centering\includegraphics[width=\linewidth]{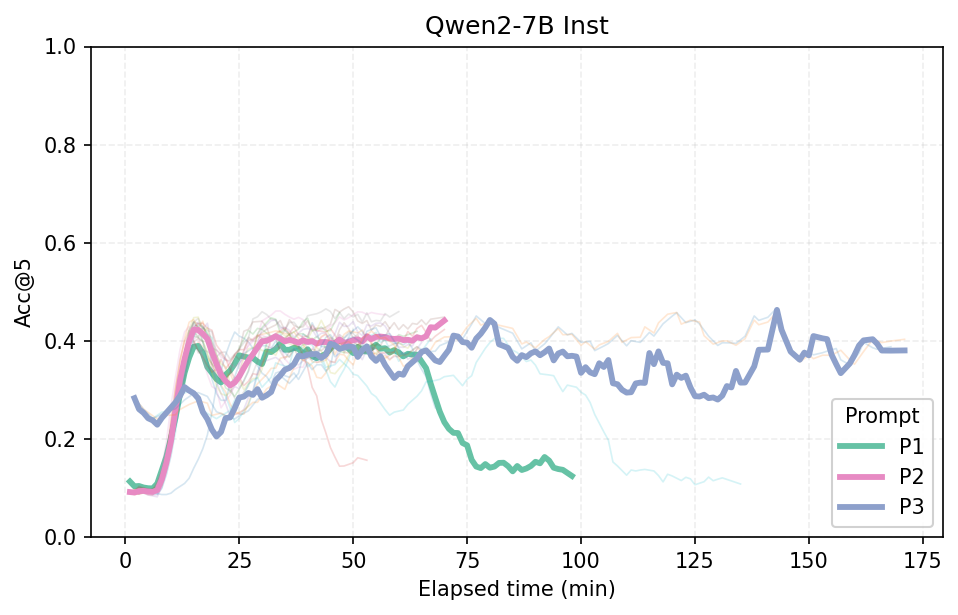}\vspace{2pt}\end{subfigure}
\begin{subfigure}{0.48\textwidth}\centering\includegraphics[width=\linewidth]{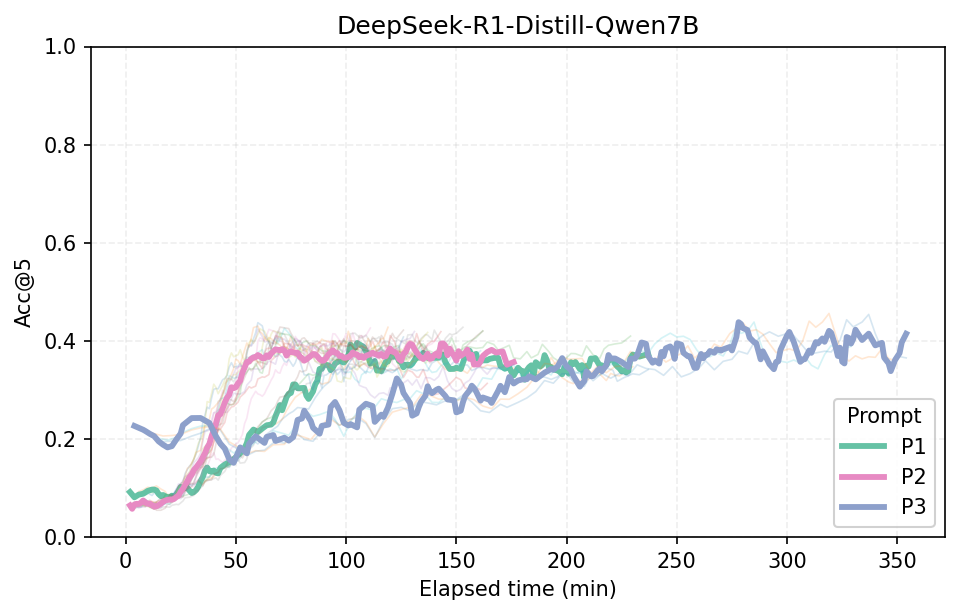}\vspace{2pt}\end{subfigure}
\caption{Acc@5 vs.\ wall-clock time. Colors denote prompts (P1/P2/P3). Thin lines: individual runs; thick lines: per-minute medians (across all rewards).}
\label{fig:acc5-time-domain-2panels}
\end{figure}

We track \textbf{Acc@\!5} both versus training steps (Figure~\ref{fig:acc5-steps-domain}) and versus wall-clock time (Figure~\ref{fig:acc5-time-domain-2panels}), comparing reward designs as described in Eqs.~\ref{eq:r0}--\ref{eq:r2}, and prompts.

\textbf{Reward effects.} Binary Reward (\textbf{R1}) performs worst: several runs decrease in Acc@\!5 near the end of training. Our \textbf{R2} consistently achieves higher final Acc@\!5 and larger relative gains from initialization than either baseline.

\textbf{Prompt effects.} \textbf{P3} starts substantially higher (Acc@\!5 \(\approx 23\!-\!29\)), \textbf{P1} is mid (\(\sim 10\)), and \textbf{P2} is lowest (\(\sim 6\!-\!9\)). Although \textbf{P2} exposes richer categorical hints compared to \textbf{P1} (no domain info), they do not translate into immediate reasoning gains. Relative to the Category-Mean baseline in \S\ref{subsubsec:llm-extracter} (Acc@\!5 \(=22\)), \textbf{P3} already adds zero-shot lift on the same information and outperforms most B0/B1 baselines except SVR.

Despite weak zero-shot, \textbf{P2} ultimately attains the highest Acc@\!5 after training, indicating that its vaguer, uncertain domain cues are effective as learning signals even if they initially distract. 
In contrast, \textbf{R1} is the only setting with runs that fail to surpass the classical baselines by the end.

When plotted against steps, many runs converge to similar Acc@\!5. However, as Figure~\ref{fig:acc5-time-domain-2panels} reveals, when considering the wall-clock time, \textbf{P3} requires significantly more time to reach the same Acc@\!5 level achieved by \textbf{P1}/\textbf{P2}, despite its high starting point.

\textbf{Interpretation.} Binary rewards (\textbf{R1}) provide coarse credit and can misguide late training. (\textbf{R2}) sustains improvement. Prompted priors matter: \textbf{P3} is a strong zero-shot prior but slower to optimize; \textbf{P2} is a weaker prior yet a stronger teacher during RLVR, eventually winning on Acc@\!5.

%\textbf{Takeaway.} Prefer \textbf{R2} for stable gains. Use \textbf{P3} when warm-starting high accuracy is critical, but budget extra time; use \textbf{P2} to maximize final Acc@\!5 under fixed compute. Avoid pure binary reward \textbf{R1}.

\subsubsection{Instruction-following coverage}
\label{sec:train-coverage}

Coverage is the share of responses from which a valid answer can be extracted (e.g., a numeric value in \verb|\boxed{ }|).

\begin{figure}[!htbp]
\centering
\begin{subfigure}{0.48\textwidth}\centering\includegraphics[width=\linewidth]{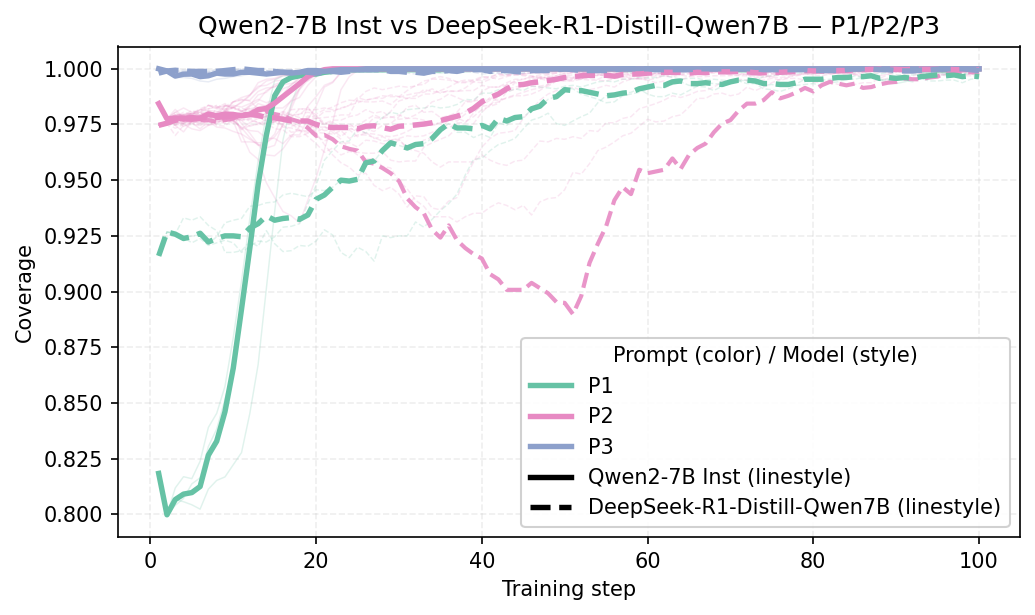}\vspace{2pt}\end{subfigure}
\begin{subfigure}{0.48\textwidth}\centering\includegraphics[width=\linewidth]{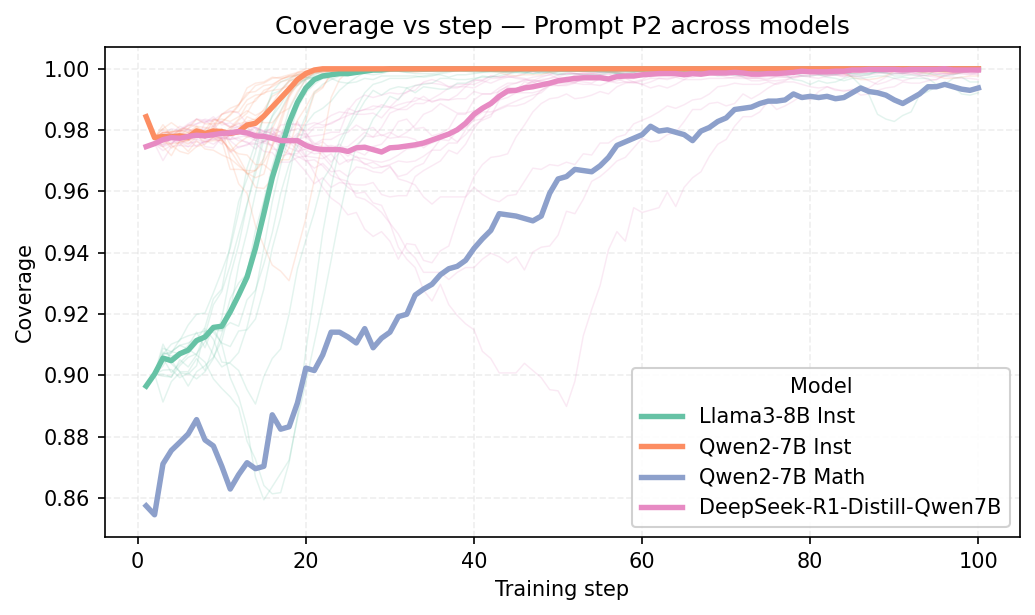}\vspace{2pt}\end{subfigure}
\caption{Coverage vs.\ training step. Left: Qwen2-7B Inst (solid) vs DeepSeek-R1-Distill-Qwen7B (dashed) under prompts P1/P2/P3 (colors). Right: Prompt P2 across all models. Thin lines are individual runs; thick lines are per-step medians. The DeepSeek-R1 run with the lowest coverage at step 50 is highlighted (left). 
% TODO: \sd{change line specs as well, for gray scale version}
}
\label{fig:coverage-two-panels}
\end{figure}

As shown in Figure~\ref{fig:coverage-two-panels}, initially, within each backbone the ordering is consistent: \(\textbf{P3}>\textbf{P2}>\textbf{P1}\). For example, \texttt{Qwen-Inst} has \(\sim80\%\) coverage with \textbf{P1} and \(100\%\) with \textbf{P3}. Across models (illustrated with \textbf{P2}): \texttt{Llama} \(\sim90\%\), \texttt{Qwen-Math} \(\sim85\%\), and \texttt{Qwen-Inst}/\texttt{DeepSeek-R1-Distill} \(\sim98\%\).

Most settings rise steadily to \(100\%\). A notable exception is \texttt{DeepSeek-R1-Distill} under \textbf{P2}, which dips early and then recovers.

\textbf{Prompt effects.} \textbf{P3} strongly implies the desired format (high zero-shot coverage). \textbf{P2} improves adherence relative to the domain-free \textbf{P1}, but is less prescriptive than \textbf{P3}.

\subsection{Limitations}
While we have demonstrated the feasibility of using RLVR-trained LLMs for incident duration prediction, there are several limitations that provide directions for future work.

First, the "ground truth" durations ($y_e$) used as the prediction target are not from a structured, objective source. As detailed in \S\ref{subsec:data-intro}, these durations were inferred using an LLM-assisted, majority-voting procedure on alert sequences to identify a "terminal alert". This process introduces potential label noise from the LLM-annotator. Furthermore, events without a confidently voted terminal alert were excluded, potentially biasing the dataset.

\begin{figure}[t]
\centering
\includegraphics[width=1\linewidth]{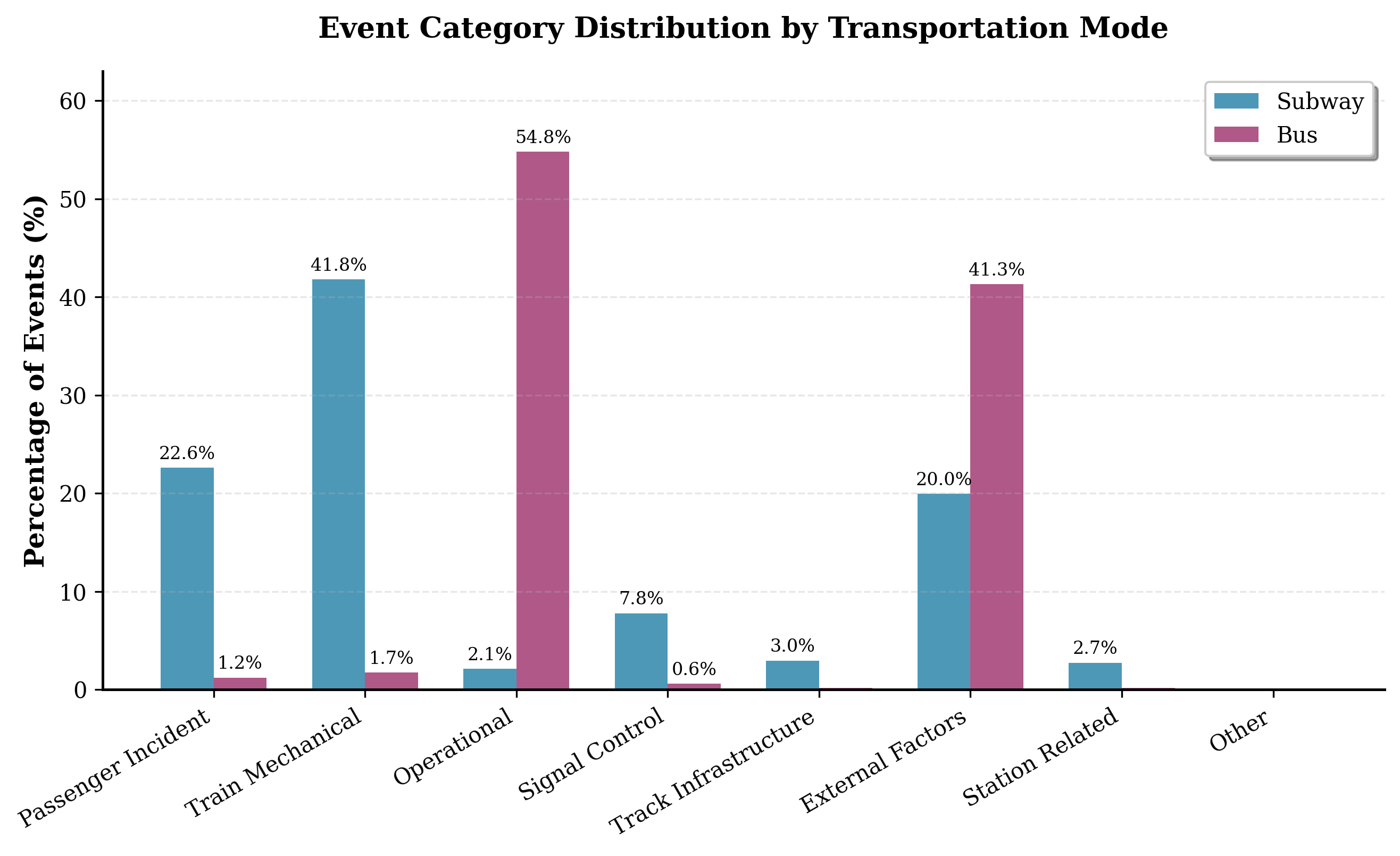}
\caption{\textbf{Event category distribution by transportation mode.} Comparison between subway (N=17,947, 85.1\%) and bus (N=2,609, 12.4\%) systems shows distinct operational patterns. Subway experiences more mechanical and infrastructure issues (39.5\% train mechanical), while bus events are dominated by external factors (40.0\%) and operational issues (28.6\%). 
% \sd{Wouldn't Fig. 1, 6, and this one convey similar information?}
}
\label{fig:mode_comparison}
\end{figure}

Second, Our dataset is sourced exclusively from the New York City MTA system. The specific alert phrasing may not be representative of other transit systems. For instance, NYC is a subway-dominant network, as shown in Figure~\ref{fig:mode_comparison}, which may differ from the systems where bus dominates. 

The third is the limited training scale. The RLVR models are trained for 100 steps, with training jobs capped at six hours. While it is already sufficient to show performance gains over baselines (especially for tight accuracy bands), this limited budget may not reveal the behaviors at large scale. The training dynamics, such as the P2 prompt's low initial accuracy but high final performance, suggest that longer training could yield different results. Thus, it remains an open question for the full potential of RLVR finetuned LLMs in predicting impact of transit incidents.

\section{Conclusion and future works}
\subsection{Conclusion}
This work presents the first, to our knowledge, application of RLVR to the real-world, noisy task of public transit incident duration forecasting from unstructured text alerts. Our central research question was whether the reasoning capabilities of math-focused LLMs could transfer to this uncertain, continuous prediction task. Our findings indicate they do not; specialized math models were outperformed by general-purpose, instruction-tuned models, which were better equipped to handle the noisy and ambiguous nature of real-world alert text. However, we demonstrate that a well-designed reward is highly effective. By adapting the verifier to this continuous, noisy-label domain using a shaped reward, we significantly improved performance. This continuous reward design is critical, as the simple binary reward was unstable and led to performance drops. We argue this highlights the need to design verifiers that reflect the continuous nature of the forecasting problem rather than a binary notion of correctness. We also found that prompt design creates a trade-off between zero-shot performance and training efficacy. Detailed prompts could provide a strong zero-shot baseline, but this prior knowledge acted more like an obstacle during RLVR training. The simpler P2 prompt, despite a weak start, ultimately achieved the highest final accuracy.

The resulting RLVR-trained models have a distinct performance profile compared to classical regression baselines. While regressors like SVR were superior at minimizing overall MAE/MSE, our RLVR approach dominated at strict, tight-band accuracy (e.g., Acc@5 and Acc@10). For instance, our RLVR model achived Acc@5 of 0.471, 35\% higher than the strongest baseline (LLM-as-extractor with Acc@5 of 0.349). This suggests RLVR is the preferred method for applications requiring high-precision, early-stage estimates, whereas classical methods suffice for coarser, hour-scale reporting.

\subsection{Future Works}
Several future research directions can be explored following this study.

The first is Spatio-Temporal Impact. Our current model focuses exclusively on the temporal impact of an incident. A critical next step is to extend this framework to predict the spatial impact, such as identifying the specific network segments, stations, and routes that will be affected and quantifying the delay propagation.

The second is Cross-City Generalizability. The models were trained and evaluated solely on data from the NYC MTA system. Future work will investigate cross-city transferability. Testing the trained models on alerts from other transit agencies would reveal whether the learned linguistic and operational patterns are universal or city-specific, and would motivate research into domain adaptation techniques for this task.

\section*{Acknowledgments}
We gratefully acknowledge use of the research computing resources of the Empire AI Consortium, Inc, with support from the State of New York, the Simons Foundation, and the Secunda Family Foundation.
 
% Future works, and how to address for the future directions

\clearpage
\bibliography{ref_drl}

\begin{thebibliography}{59}
\expandafter\ifx\csname natexlab\endcsname\relax\def\natexlab#1{#1}\fi
\providecommand{\url}[1]{\texttt{#1}}
\providecommand{\href}[2]{#2}
\providecommand{\path}[1]{#1}
\providecommand{\DOIprefix}{doi:}
\providecommand{\ArXivprefix}{arXiv:}
\providecommand{\URLprefix}{URL: }
\providecommand{\Pubmedprefix}{pmid:}
\providecommand{\doi}[1]{\href{http://dx.doi.org/#1}{\path{#1}}}
\providecommand{\Pubmed}[1]{\href{pmid:#1}{\path{#1}}}
\providecommand{\bibinfo}[2]{#2}
\ifx\xfnm\relax \def\xfnm[#1]{\unskip,\space#1}\fi
%Type = Article
\bibitem[{Agarwal and Rambha(2024)}]{agarwal2024scalable}
\bibinfo{author}{Agarwal, P.}, \bibinfo{author}{Rambha, T.}, \bibinfo{year}{2024}.
\newblock \bibinfo{title}{Scalable algorithms for bicriterion trip-based transit routing}.
\newblock \bibinfo{journal}{IEEE Transactions on Intelligent Transportation Systems} \bibinfo{volume}{25}, \bibinfo{pages}{14313--14327}.
%Type = Inproceedings
\bibitem[{Bast et~al.(2010)Bast, Carlsson, Eigenwillig, Geisberger, Harrelson, Raychev and Viger}]{bast2010fast}
\bibinfo{author}{Bast, H.}, \bibinfo{author}{Carlsson, E.}, \bibinfo{author}{Eigenwillig, A.}, \bibinfo{author}{Geisberger, R.}, \bibinfo{author}{Harrelson, C.}, \bibinfo{author}{Raychev, V.}, \bibinfo{author}{Viger, F.}, \bibinfo{year}{2010}.
\newblock \bibinfo{title}{Fast routing in very large public transportation networks using transfer patterns}, in: \bibinfo{booktitle}{European Symposium on Algorithms}, \bibinfo{organization}{Springer}. pp. \bibinfo{pages}{290--301}.
%Type = Incollection
\bibitem[{Bloom et~al.(2025)Bloom, Brumberg, Fisk, Harrison, Hull, Ramasubramanian, Van~Vliet and Wing}]{bloom2025empire}
\bibinfo{author}{Bloom, S.}, \bibinfo{author}{Brumberg, J.}, \bibinfo{author}{Fisk, I.}, \bibinfo{author}{Harrison, R.}, \bibinfo{author}{Hull, R.}, \bibinfo{author}{Ramasubramanian, M.}, \bibinfo{author}{Van~Vliet, K.}, \bibinfo{author}{Wing, J.}, \bibinfo{year}{2025}.
\newblock \bibinfo{title}{Empire ai: A new model for provisioning ai and hpc for academic research in the public good}, in: \bibinfo{booktitle}{Practice and Experience in Advanced Research Computing 2025: The Power of Collaboration}, pp. \bibinfo{pages}{1--4}.
%Type = Article
\bibitem[{Brown et~al.(2020)Brown, Mann, Ryder, Subbiah, Kaplan, Dhariwal, Neelakantan, Shyam, Sastry, Askell et~al.}]{brown2020language}
\bibinfo{author}{Brown, T.}, \bibinfo{author}{Mann, B.}, \bibinfo{author}{Ryder, N.}, \bibinfo{author}{Subbiah, M.}, \bibinfo{author}{Kaplan, J.D.}, \bibinfo{author}{Dhariwal, P.}, \bibinfo{author}{Neelakantan, A.}, \bibinfo{author}{Shyam, P.}, \bibinfo{author}{Sastry, G.}, \bibinfo{author}{Askell, A.}, et~al., \bibinfo{year}{2020}.
\newblock \bibinfo{title}{Language models are few-shot learners}.
\newblock \bibinfo{journal}{Advances in neural information processing systems} \bibinfo{volume}{33}, \bibinfo{pages}{1877--1901}.
%Type = Article
\bibitem[{Cai et~al.(2025)Cai, Wang, Liu, Liu, Niu and Sugiyama}]{cai2025reinforcement}
\bibinfo{author}{Cai, X.Q.}, \bibinfo{author}{Wang, W.}, \bibinfo{author}{Liu, F.}, \bibinfo{author}{Liu, T.}, \bibinfo{author}{Niu, G.}, \bibinfo{author}{Sugiyama, M.}, \bibinfo{year}{2025}.
\newblock \bibinfo{title}{Reinforcement learning with verifiable yet noisy rewards under imperfect verifiers}.
\newblock \bibinfo{journal}{arXiv preprint arXiv:2510.00915} .
%Type = Article
\bibitem[{Chen et~al.(2024)Chen, He, Wang, Chen and Luo}]{chen2024delayptc}
\bibinfo{author}{Chen, C.}, \bibinfo{author}{He, Y.}, \bibinfo{author}{Wang, H.}, \bibinfo{author}{Chen, J.}, \bibinfo{author}{Luo, Q.}, \bibinfo{year}{2024}.
\newblock \bibinfo{title}{Delayptc-llm: Metro passenger travel choice prediction under train delays with large language models}.
\newblock \bibinfo{journal}{arXiv preprint arXiv:2410.00052} .
%Type = Article
\bibitem[{Christiano et~al.(2017)Christiano, Leike, Brown, Martic, Legg and Amodei}]{christiano2017deep}
\bibinfo{author}{Christiano, P.F.}, \bibinfo{author}{Leike, J.}, \bibinfo{author}{Brown, T.}, \bibinfo{author}{Martic, M.}, \bibinfo{author}{Legg, S.}, \bibinfo{author}{Amodei, D.}, \bibinfo{year}{2017}.
\newblock \bibinfo{title}{Deep reinforcement learning from human preferences}.
\newblock \bibinfo{journal}{Advances in neural information processing systems} \bibinfo{volume}{30}.
%Type = Article
\bibitem[{Cobbe et~al.(2021)Cobbe, Kosaraju, Bavarian, Chen, Jun, Kaiser, Plappert, Tworek, Hilton, Nakano et~al.}]{cobbe2021training}
\bibinfo{author}{Cobbe, K.}, \bibinfo{author}{Kosaraju, V.}, \bibinfo{author}{Bavarian, M.}, \bibinfo{author}{Chen, M.}, \bibinfo{author}{Jun, H.}, \bibinfo{author}{Kaiser, L.}, \bibinfo{author}{Plappert, M.}, \bibinfo{author}{Tworek, J.}, \bibinfo{author}{Hilton, J.}, \bibinfo{author}{Nakano, R.}, et~al., \bibinfo{year}{2021}.
\newblock \bibinfo{title}{Training verifiers to solve math word problems}.
\newblock \bibinfo{journal}{arXiv preprint arXiv:2110.14168} .
%Type = Article
\bibitem[{Cui et~al.(2025)Cui, Zhang, Chen, Yuan, Wang, Zuo, Li, Fan, Chen, Chen et~al.}]{cui2025entropy}
\bibinfo{author}{Cui, G.}, \bibinfo{author}{Zhang, Y.}, \bibinfo{author}{Chen, J.}, \bibinfo{author}{Yuan, L.}, \bibinfo{author}{Wang, Z.}, \bibinfo{author}{Zuo, Y.}, \bibinfo{author}{Li, H.}, \bibinfo{author}{Fan, Y.}, \bibinfo{author}{Chen, H.}, \bibinfo{author}{Chen, W.}, et~al., \bibinfo{year}{2025}.
\newblock \bibinfo{title}{The entropy mechanism of reinforcement learning for reasoning language models}.
\newblock \bibinfo{journal}{arXiv preprint arXiv:2505.22617} .
%Type = Inproceedings
\bibitem[{Delling et~al.(2017)Delling, Dibbelt, Pajor and Z{\"u}ndorf}]{delling2017faster}
\bibinfo{author}{Delling, D.}, \bibinfo{author}{Dibbelt, J.}, \bibinfo{author}{Pajor, T.}, \bibinfo{author}{Z{\"u}ndorf, T.}, \bibinfo{year}{2017}.
\newblock \bibinfo{title}{Faster transit routing by hyper partitioning}, in: \bibinfo{booktitle}{17th Workshop on Algorithmic Approaches for Transportation Modelling, Optimization, and Systems (ATMOS 2017)}, \bibinfo{organization}{Schloss Dagstuhl--Leibniz-Zentrum f{\"u}r Informatik}. pp. \bibinfo{pages}{8--1}.
%Type = Article
\bibitem[{Devunuri and Lehe(2025)}]{devunuri2025transitgpt}
\bibinfo{author}{Devunuri, S.}, \bibinfo{author}{Lehe, L.}, \bibinfo{year}{2025}.
\newblock \bibinfo{title}{Transitgpt: a generative ai-based framework for interacting with gtfs data using large language models: S. devunuri, l. lehe}.
\newblock \bibinfo{journal}{Public Transport} , \bibinfo{pages}{1--27}.
%Type = Article
\bibitem[{Fang et~al.(2024)Fang, Yang, Wang and Di}]{fang2024travellm}
\bibinfo{author}{Fang, B.}, \bibinfo{author}{Yang, Z.}, \bibinfo{author}{Wang, S.}, \bibinfo{author}{Di, X.}, \bibinfo{year}{2024}.
\newblock \bibinfo{title}{Travellm: Could you plan my new public transit route in face of a network disruption?}
\newblock \bibinfo{journal}{arXiv preprint arXiv:2407.14926} .
%Type = Article
\bibitem[{Gruver et~al.(2023)Gruver, Finzi, Qiu and Wilson}]{gruver2023large}
\bibinfo{author}{Gruver, N.}, \bibinfo{author}{Finzi, M.}, \bibinfo{author}{Qiu, S.}, \bibinfo{author}{Wilson, A.G.}, \bibinfo{year}{2023}.
\newblock \bibinfo{title}{Large language models are zero-shot time series forecasters}.
\newblock \bibinfo{journal}{Advances in Neural Information Processing Systems} \bibinfo{volume}{36}, \bibinfo{pages}{19622--19635}.
%Type = Inproceedings
\bibitem[{Gubichev et~al.(2010)Gubichev, Bedathur, Seufert and Weikum}]{gubichev2010fast}
\bibinfo{author}{Gubichev, A.}, \bibinfo{author}{Bedathur, S.}, \bibinfo{author}{Seufert, S.}, \bibinfo{author}{Weikum, G.}, \bibinfo{year}{2010}.
\newblock \bibinfo{title}{Fast and accurate estimation of shortest paths in large graphs}, in: \bibinfo{booktitle}{Proceedings of the 19th ACM international conference on Information and knowledge management}, pp. \bibinfo{pages}{499--508}.
%Type = Article
\bibitem[{Guo et~al.(2025a)Guo, Yang, Zhang, Song, Zhang, Xu, Zhu, Ma, Wang, Bi et~al.}]{guo2025deepseek}
\bibinfo{author}{Guo, D.}, \bibinfo{author}{Yang, D.}, \bibinfo{author}{Zhang, H.}, \bibinfo{author}{Song, J.}, \bibinfo{author}{Zhang, R.}, \bibinfo{author}{Xu, R.}, \bibinfo{author}{Zhu, Q.}, \bibinfo{author}{Ma, S.}, \bibinfo{author}{Wang, P.}, \bibinfo{author}{Bi, X.}, et~al., \bibinfo{year}{2025}a.
\newblock \bibinfo{title}{Deepseek-r1: Incentivizing reasoning capability in llms via reinforcement learning}.
\newblock \bibinfo{journal}{arXiv preprint arXiv:2501.12948} .
%Type = Article
\bibitem[{Guo et~al.(2025b)Guo, Yang, Peng, Lu, Zhu and Yang}]{guo2025automating}
\bibinfo{author}{Guo, X.}, \bibinfo{author}{Yang, X.}, \bibinfo{author}{Peng, M.}, \bibinfo{author}{Lu, H.}, \bibinfo{author}{Zhu, M.}, \bibinfo{author}{Yang, H.}, \bibinfo{year}{2025}b.
\newblock \bibinfo{title}{Automating traffic model enhancement with ai research agent}.
\newblock \bibinfo{journal}{Transportation Research Part C: Emerging Technologies} \bibinfo{volume}{178}, \bibinfo{pages}{105187}.
%Type = Article
\bibitem[{Jariyasunant et~al.(2011)Jariyasunant, Mai and Sengupta}]{jariyasunant2011algorithm}
\bibinfo{author}{Jariyasunant, J.}, \bibinfo{author}{Mai, E.}, \bibinfo{author}{Sengupta, R.}, \bibinfo{year}{2011}.
\newblock \bibinfo{title}{Algorithm for finding optimal paths in a public transit network with real-time data}.
\newblock \bibinfo{journal}{Transportation research record} \bibinfo{volume}{2256}, \bibinfo{pages}{34--42}.
%Type = Article
\bibitem[{Jin et~al.(2021)Jin, Wi, Lee, Kang, Kim and Kim}]{jin2021trafficbert}
\bibinfo{author}{Jin, K.}, \bibinfo{author}{Wi, J.}, \bibinfo{author}{Lee, E.}, \bibinfo{author}{Kang, S.}, \bibinfo{author}{Kim, S.}, \bibinfo{author}{Kim, Y.}, \bibinfo{year}{2021}.
\newblock \bibinfo{title}{Trafficbert: Pre-trained model with large-scale data for long-range traffic flow forecasting}.
\newblock \bibinfo{journal}{Expert Systems with Applications} \bibinfo{volume}{186}, \bibinfo{pages}{115738}.
%Type = Article
\bibitem[{Jin et~al.(2023)Jin, Wang, Ma, Chu, Zhang, Shi, Chen, Liang, Li, Pan et~al.}]{jin2023time}
\bibinfo{author}{Jin, M.}, \bibinfo{author}{Wang, S.}, \bibinfo{author}{Ma, L.}, \bibinfo{author}{Chu, Z.}, \bibinfo{author}{Zhang, J.Y.}, \bibinfo{author}{Shi, X.}, \bibinfo{author}{Chen, P.Y.}, \bibinfo{author}{Liang, Y.}, \bibinfo{author}{Li, Y.F.}, \bibinfo{author}{Pan, S.}, et~al., \bibinfo{year}{2023}.
\newblock \bibinfo{title}{Time-llm: Time series forecasting by reprogramming large language models}.
\newblock \bibinfo{journal}{arXiv preprint arXiv:2310.01728} .
%Type = Inproceedings
\bibitem[{Jurayj et~al.(2025)Jurayj, Cheng and Van~Durme}]{jurayj-etal-2025-final}
\bibinfo{author}{Jurayj, W.}, \bibinfo{author}{Cheng, J.}, \bibinfo{author}{Van~Durme, B.}, \bibinfo{year}{2025}.
\newblock \bibinfo{title}{Is that your final answer? test-time scaling improves selective question answering}, in: \bibinfo{editor}{Che, W.}, \bibinfo{editor}{Nabende, J.}, \bibinfo{editor}{Shutova, E.}, \bibinfo{editor}{Pilehvar, M.T.} (Eds.), \bibinfo{booktitle}{Proceedings of the 63rd Annual Meeting of the Association for Computational Linguistics (Volume 2: Short Papers)}, \bibinfo{publisher}{Association for Computational Linguistics}, \bibinfo{address}{Vienna, Austria}. pp. \bibinfo{pages}{636--644}.
\newblock \URLprefix \url{https://aclanthology.org/2025.acl-short.50/}, \DOIprefix\doi{10.18653/v1/2025.acl-short.50}.
%Type = Article
\bibitem[{Kalair and Connaughton(2021)}]{kalair2021dynamic}
\bibinfo{author}{Kalair, K.}, \bibinfo{author}{Connaughton, C.}, \bibinfo{year}{2021}.
\newblock \bibinfo{title}{Dynamic and interpretable hazard-based models of traffic incident durations}.
\newblock \bibinfo{journal}{Frontiers in future transportation} \bibinfo{volume}{2}, \bibinfo{pages}{669015}.
%Type = Article
\bibitem[{Krishnakumari et~al.(2020)Krishnakumari, Cats and van Lint}]{krishnakumari2020estimation}
\bibinfo{author}{Krishnakumari, P.}, \bibinfo{author}{Cats, O.}, \bibinfo{author}{van Lint, H.}, \bibinfo{year}{2020}.
\newblock \bibinfo{title}{Estimation of metro network passenger delay from individual trajectories}.
\newblock \bibinfo{journal}{Transportation Research Part C: Emerging Technologies} \bibinfo{volume}{117}, \bibinfo{pages}{102704}.
%Type = Article
\bibitem[{Kuang et~al.(2025)Kuang, Liu, Qu and Wei}]{kuang2025traffic}
\bibinfo{author}{Kuang, S.}, \bibinfo{author}{Liu, Y.}, \bibinfo{author}{Qu, X.}, \bibinfo{author}{Wei, Y.}, \bibinfo{year}{2025}.
\newblock \bibinfo{title}{Traffic-it: Enhancing traffic scene understanding for multimodal large language models}.
\newblock \bibinfo{journal}{Transportation Research Part C: Emerging Technologies} \bibinfo{volume}{180}, \bibinfo{pages}{105325}.
%Type = Misc
\bibitem[{Kydlíček()}]{Kydlicek_Math-Verify_Math_Verification}
\bibinfo{author}{Kydlíček, H.}, .
\newblock \bibinfo{title}{{Math-Verify: Math Verification Library}}.
\newblock \URLprefix \url{https://github.com/huggingface/math-verify}.
%Type = Article
\bibitem[{Lewkowycz et~al.(2022)Lewkowycz, Andreassen, Dohan, Dyer, Michalewski, Ramasesh, Slone, Anil, Schlag, Gutman-Solo et~al.}]{lewkowycz2022solving}
\bibinfo{author}{Lewkowycz, A.}, \bibinfo{author}{Andreassen, A.}, \bibinfo{author}{Dohan, D.}, \bibinfo{author}{Dyer, E.}, \bibinfo{author}{Michalewski, H.}, \bibinfo{author}{Ramasesh, V.}, \bibinfo{author}{Slone, A.}, \bibinfo{author}{Anil, C.}, \bibinfo{author}{Schlag, I.}, \bibinfo{author}{Gutman-Solo, T.}, et~al., \bibinfo{year}{2022}.
\newblock \bibinfo{title}{Solving quantitative reasoning problems with language models}.
\newblock \bibinfo{journal}{Advances in neural information processing systems} \bibinfo{volume}{35}, \bibinfo{pages}{3843--3857}.
%Type = Article
\bibitem[{Li et~al.(2025)Li, Zhao, Zhang and Gan}]{li2025steering}
\bibinfo{author}{Li, J.}, \bibinfo{author}{Zhao, W.}, \bibinfo{author}{Zhang, Y.}, \bibinfo{author}{Gan, C.}, \bibinfo{year}{2025}.
\newblock \bibinfo{title}{Steering llm thinking with budget guidance}.
\newblock \bibinfo{journal}{arXiv preprint arXiv:2506.13752} .
%Type = Article
\bibitem[{Li et~al.(2018)Li, Pereira and Ben-Akiva}]{li2018overview}
\bibinfo{author}{Li, R.}, \bibinfo{author}{Pereira, F.C.}, \bibinfo{author}{Ben-Akiva, M.E.}, \bibinfo{year}{2018}.
\newblock \bibinfo{title}{Overview of traffic incident duration analysis and prediction}.
\newblock \bibinfo{journal}{European transport research review} \bibinfo{volume}{10}, \bibinfo{pages}{1--13}.
%Type = Inproceedings
\bibitem[{Lightman et~al.(2023)Lightman, Kosaraju, Burda, Edwards, Baker, Lee, Leike, Schulman, Sutskever and Cobbe}]{lightman2023let}
\bibinfo{author}{Lightman, H.}, \bibinfo{author}{Kosaraju, V.}, \bibinfo{author}{Burda, Y.}, \bibinfo{author}{Edwards, H.}, \bibinfo{author}{Baker, B.}, \bibinfo{author}{Lee, T.}, \bibinfo{author}{Leike, J.}, \bibinfo{author}{Schulman, J.}, \bibinfo{author}{Sutskever, I.}, \bibinfo{author}{Cobbe, K.}, \bibinfo{year}{2023}.
\newblock \bibinfo{title}{Let's verify step by step}, in: \bibinfo{booktitle}{The Twelfth International Conference on Learning Representations}.
%Type = Inproceedings
\bibitem[{Liu et~al.(2024)Liu, Yang, Xu, Li, Long, Li and Zhao}]{liu2024spatial}
\bibinfo{author}{Liu, C.}, \bibinfo{author}{Yang, S.}, \bibinfo{author}{Xu, Q.}, \bibinfo{author}{Li, Z.}, \bibinfo{author}{Long, C.}, \bibinfo{author}{Li, Z.}, \bibinfo{author}{Zhao, R.}, \bibinfo{year}{2024}.
\newblock \bibinfo{title}{Spatial-temporal large language model for traffic prediction}, in: \bibinfo{booktitle}{2024 25th IEEE International Conference on Mobile Data Management (MDM)}, \bibinfo{organization}{IEEE}. pp. \bibinfo{pages}{31--40}.
%Type = Article
\bibitem[{Liu et~al.(2023)Liu, Yuan, Fu, Jiang, Hayashi and Neubig}]{liu2023pre}
\bibinfo{author}{Liu, P.}, \bibinfo{author}{Yuan, W.}, \bibinfo{author}{Fu, J.}, \bibinfo{author}{Jiang, Z.}, \bibinfo{author}{Hayashi, H.}, \bibinfo{author}{Neubig, G.}, \bibinfo{year}{2023}.
\newblock \bibinfo{title}{Pre-train, prompt, and predict: A systematic survey of prompting methods in natural language processing}.
\newblock \bibinfo{journal}{ACM computing surveys} \bibinfo{volume}{55}, \bibinfo{pages}{1--35}.
%Type = Article
\bibitem[{Liu et~al.(2025)Liu, Zhou, Gu, Liu, Liu, Zhang, He and Zhang}]{liu2025trip}
\bibinfo{author}{Liu, Z.}, \bibinfo{author}{Zhou, Z.}, \bibinfo{author}{Gu, Z.}, \bibinfo{author}{Liu, S.}, \bibinfo{author}{Liu, P.}, \bibinfo{author}{Zhang, Y.}, \bibinfo{author}{He, Y.}, \bibinfo{author}{Zhang, K.}, \bibinfo{year}{2025}.
\newblock \bibinfo{title}{Trip: Transport reasoning with intelligence progression—a foundation framework}.
\newblock \bibinfo{journal}{Transportation Research Part C: Emerging Technologies} \bibinfo{volume}{179}, \bibinfo{pages}{105260}.
%Type = Article
\bibitem[{Ma et~al.(2025)Ma, Liao, Liu, Jiang, Stanford, Cao and Ma}]{ma5218098learning}
\bibinfo{author}{Ma, H.}, \bibinfo{author}{Liao, X.}, \bibinfo{author}{Liu, Y.}, \bibinfo{author}{Jiang, Q.}, \bibinfo{author}{Stanford, C.}, \bibinfo{author}{Cao, A.}, \bibinfo{author}{Ma, J.}, \bibinfo{year}{2025}.
\newblock \bibinfo{title}{Learning universal human mobility patterns with a foundation model for cross-domain data fusion}.
\newblock \bibinfo{journal}{Available at SSRN 5218098} .
%Type = Article
\bibitem[{Muennighoff et~al.(2025)Muennighoff, Yang, Shi, Li, Fei-Fei, Hajishirzi, Zettlemoyer, Liang, Cand{\`e}s and Hashimoto}]{muennighoff2025s1}
\bibinfo{author}{Muennighoff, N.}, \bibinfo{author}{Yang, Z.}, \bibinfo{author}{Shi, W.}, \bibinfo{author}{Li, X.L.}, \bibinfo{author}{Fei-Fei, L.}, \bibinfo{author}{Hajishirzi, H.}, \bibinfo{author}{Zettlemoyer, L.}, \bibinfo{author}{Liang, P.}, \bibinfo{author}{Cand{\`e}s, E.}, \bibinfo{author}{Hashimoto, T.}, \bibinfo{year}{2025}.
\newblock \bibinfo{title}{s1: Simple test-time scaling}.
\newblock \bibinfo{journal}{arXiv preprint arXiv:2501.19393} .
%Type = Article
\bibitem[{Narayanan et~al.(2024)Narayanan, Makarov and Antoniou}]{narayanan2024graph}
\bibinfo{author}{Narayanan, S.}, \bibinfo{author}{Makarov, N.}, \bibinfo{author}{Antoniou, C.}, \bibinfo{year}{2024}.
\newblock \bibinfo{title}{Graph neural networks as strategic transport modelling alternative-a proof of concept for a surrogate}.
\newblock \bibinfo{journal}{IET Intelligent Transport Systems} \bibinfo{volume}{18}, \bibinfo{pages}{2059--2077}.
%Type = Article
\bibitem[{Ouyang et~al.(2022)Ouyang, Wu, Jiang, Almeida, Wainwright, Mishkin, Zhang, Agarwal, Slama, Ray et~al.}]{ouyang2022training}
\bibinfo{author}{Ouyang, L.}, \bibinfo{author}{Wu, J.}, \bibinfo{author}{Jiang, X.}, \bibinfo{author}{Almeida, D.}, \bibinfo{author}{Wainwright, C.}, \bibinfo{author}{Mishkin, P.}, \bibinfo{author}{Zhang, C.}, \bibinfo{author}{Agarwal, S.}, \bibinfo{author}{Slama, K.}, \bibinfo{author}{Ray, A.}, et~al., \bibinfo{year}{2022}.
\newblock \bibinfo{title}{Training language models to follow instructions with human feedback}.
\newblock \bibinfo{journal}{Advances in neural information processing systems} \bibinfo{volume}{35}, \bibinfo{pages}{27730--27744}.
%Type = Article
\bibitem[{Pedregosa et~al.(2011)Pedregosa, Varoquaux, Gramfort, Michel, Thirion, Grisel, Blondel, Prettenhofer, Weiss, Dubourg, Vanderplas, Passos, Cournapeau, Brucher, Perrot and Duchesnay}]{scikit-learn}
\bibinfo{author}{Pedregosa, F.}, \bibinfo{author}{Varoquaux, G.}, \bibinfo{author}{Gramfort, A.}, \bibinfo{author}{Michel, V.}, \bibinfo{author}{Thirion, B.}, \bibinfo{author}{Grisel, O.}, \bibinfo{author}{Blondel, M.}, \bibinfo{author}{Prettenhofer, P.}, \bibinfo{author}{Weiss, R.}, \bibinfo{author}{Dubourg, V.}, \bibinfo{author}{Vanderplas, J.}, \bibinfo{author}{Passos, A.}, \bibinfo{author}{Cournapeau, D.}, \bibinfo{author}{Brucher, M.}, \bibinfo{author}{Perrot, M.}, \bibinfo{author}{Duchesnay, E.}, \bibinfo{year}{2011}.
\newblock \bibinfo{title}{Scikit-learn: Machine learning in {P}ython}.
\newblock \bibinfo{journal}{Journal of Machine Learning Research} \bibinfo{volume}{12}, \bibinfo{pages}{2825--2830}.
%Type = Article
\bibitem[{Rafailov et~al.(2023)Rafailov, Sharma, Mitchell, Manning, Ermon and Finn}]{rafailov2023direct}
\bibinfo{author}{Rafailov, R.}, \bibinfo{author}{Sharma, A.}, \bibinfo{author}{Mitchell, E.}, \bibinfo{author}{Manning, C.D.}, \bibinfo{author}{Ermon, S.}, \bibinfo{author}{Finn, C.}, \bibinfo{year}{2023}.
\newblock \bibinfo{title}{Direct preference optimization: Your language model is secretly a reward model}.
\newblock \bibinfo{journal}{Advances in neural information processing systems} \bibinfo{volume}{36}, \bibinfo{pages}{53728--53741}.
%Type = Article
\bibitem[{Sarhani et~al.(2025)Sarhani, Nourmohammadzadeh, Vo{\ss} and Amrani}]{sarhani2025predicting}
\bibinfo{author}{Sarhani, M.}, \bibinfo{author}{Nourmohammadzadeh, A.}, \bibinfo{author}{Vo{\ss}, S.}, \bibinfo{author}{Amrani, M.E.}, \bibinfo{year}{2025}.
\newblock \bibinfo{title}{Predicting and analyzing ferry transit delays using open data and machine learning}.
\newblock \bibinfo{journal}{Journal of Public Transportation} \bibinfo{volume}{27}, \bibinfo{pages}{100124}.
%Type = Article
\bibitem[{Sarhani and Vo{\ss}(2024)}]{sarhani2024prediction}
\bibinfo{author}{Sarhani, M.}, \bibinfo{author}{Vo{\ss}, S.}, \bibinfo{year}{2024}.
\newblock \bibinfo{title}{Prediction of rail transit delays with machine learning: How to exploit open data sources}.
\newblock \bibinfo{journal}{Multimodal Transportation} \bibinfo{volume}{3}, \bibinfo{pages}{100120}.
%Type = Article
\bibitem[{Schulman(2020)}]{schulman2020approximating}
\bibinfo{author}{Schulman, J.}, \bibinfo{year}{2020}.
\newblock \bibinfo{title}{Approximating kl divergence}.
\newblock \bibinfo{journal}{John Schulman’s Homepage} .
%Type = Article
\bibitem[{Shao et~al.(2025)Shao, Li, Xin, Geng, Wang, Oh, Du, Lambert, Min, Krishna et~al.}]{shao2025spurious}
\bibinfo{author}{Shao, R.}, \bibinfo{author}{Li, S.S.}, \bibinfo{author}{Xin, R.}, \bibinfo{author}{Geng, S.}, \bibinfo{author}{Wang, Y.}, \bibinfo{author}{Oh, S.}, \bibinfo{author}{Du, S.S.}, \bibinfo{author}{Lambert, N.}, \bibinfo{author}{Min, S.}, \bibinfo{author}{Krishna, R.}, et~al., \bibinfo{year}{2025}.
\newblock \bibinfo{title}{Spurious rewards: Rethinking training signals in rlvr}.
\newblock \bibinfo{journal}{arXiv preprint arXiv:2506.10947} .
%Type = Article
\bibitem[{Shao et~al.(2024)Shao, Wang, Zhu, Xu, Song, Bi, Zhang, Zhang, Li, Wu et~al.}]{shao2024deepseekmath}
\bibinfo{author}{Shao, Z.}, \bibinfo{author}{Wang, P.}, \bibinfo{author}{Zhu, Q.}, \bibinfo{author}{Xu, R.}, \bibinfo{author}{Song, J.}, \bibinfo{author}{Bi, X.}, \bibinfo{author}{Zhang, H.}, \bibinfo{author}{Zhang, M.}, \bibinfo{author}{Li, Y.}, \bibinfo{author}{Wu, Y.}, et~al., \bibinfo{year}{2024}.
\newblock \bibinfo{title}{Deepseekmath: Pushing the limits of mathematical reasoning in open language models}.
\newblock \bibinfo{journal}{arXiv preprint arXiv:2402.03300} .
%Type = Article
\bibitem[{Wang et~al.(2024)Wang, Feng, Qiu, Gu and Zhao}]{wang2024news}
\bibinfo{author}{Wang, X.}, \bibinfo{author}{Feng, M.}, \bibinfo{author}{Qiu, J.}, \bibinfo{author}{Gu, J.}, \bibinfo{author}{Zhao, J.}, \bibinfo{year}{2024}.
\newblock \bibinfo{title}{From news to forecast: Integrating event analysis in llm-based time series forecasting with reflection}.
\newblock \bibinfo{journal}{Advances in Neural Information Processing Systems} \bibinfo{volume}{37}, \bibinfo{pages}{58118--58153}.
%Type = Article
\bibitem[{Wang et~al.(2022)Wang, Wei, Schuurmans, Le, Chi, Narang, Chowdhery and Zhou}]{wang2022self}
\bibinfo{author}{Wang, X.}, \bibinfo{author}{Wei, J.}, \bibinfo{author}{Schuurmans, D.}, \bibinfo{author}{Le, Q.}, \bibinfo{author}{Chi, E.}, \bibinfo{author}{Narang, S.}, \bibinfo{author}{Chowdhery, A.}, \bibinfo{author}{Zhou, D.}, \bibinfo{year}{2022}.
\newblock \bibinfo{title}{Self-consistency improves chain of thought reasoning in language models}.
\newblock \bibinfo{journal}{arXiv preprint arXiv:2203.11171} .
%Type = Article
\bibitem[{Wei et~al.(2021)Wei, Bosma, Zhao, Guu, Yu, Lester, Du, Dai and Le}]{wei2021finetuned}
\bibinfo{author}{Wei, J.}, \bibinfo{author}{Bosma, M.}, \bibinfo{author}{Zhao, V.Y.}, \bibinfo{author}{Guu, K.}, \bibinfo{author}{Yu, A.W.}, \bibinfo{author}{Lester, B.}, \bibinfo{author}{Du, N.}, \bibinfo{author}{Dai, A.M.}, \bibinfo{author}{Le, Q.V.}, \bibinfo{year}{2021}.
\newblock \bibinfo{title}{Finetuned language models are zero-shot learners}.
\newblock \bibinfo{journal}{arXiv preprint arXiv:2109.01652} .
%Type = Article
\bibitem[{Wen et~al.(2025)Wen, Liu, Zheng, Xu, Ye, Wu, Liang, Wang, Li, Miao et~al.}]{wen2025reinforcement}
\bibinfo{author}{Wen, X.}, \bibinfo{author}{Liu, Z.}, \bibinfo{author}{Zheng, S.}, \bibinfo{author}{Xu, Z.}, \bibinfo{author}{Ye, S.}, \bibinfo{author}{Wu, Z.}, \bibinfo{author}{Liang, X.}, \bibinfo{author}{Wang, Y.}, \bibinfo{author}{Li, J.}, \bibinfo{author}{Miao, Z.}, et~al., \bibinfo{year}{2025}.
\newblock \bibinfo{title}{Reinforcement learning with verifiable rewards implicitly incentivizes correct reasoning in base llms}.
\newblock \bibinfo{journal}{arXiv preprint arXiv:2506.14245} .
%Type = Article
\bibitem[{Witt(2016)}]{witt2016trip}
\bibinfo{author}{Witt, S.}, \bibinfo{year}{2016}.
\newblock \bibinfo{title}{Trip-based public transit routing using condensed search trees}.
\newblock \bibinfo{journal}{arXiv preprint arXiv:1607.01299} .
%Type = Article
\bibitem[{Wu et~al.(2025)Wu, Zhang, Dong, Xi, Zhao, Jin, Fan, Zhou, Fu, Liu et~al.}]{wu2025reasoning}
\bibinfo{author}{Wu, M.}, \bibinfo{author}{Zhang, Z.}, \bibinfo{author}{Dong, Q.}, \bibinfo{author}{Xi, Z.}, \bibinfo{author}{Zhao, J.}, \bibinfo{author}{Jin, S.}, \bibinfo{author}{Fan, X.}, \bibinfo{author}{Zhou, Y.}, \bibinfo{author}{Fu, Y.}, \bibinfo{author}{Liu, Q.}, et~al., \bibinfo{year}{2025}.
\newblock \bibinfo{title}{Reasoning or memorization? unreliable results of reinforcement learning due to data contamination}.
\newblock \bibinfo{journal}{arXiv preprint arXiv:2507.10532} .
%Type = Article
\bibitem[{Xue et~al.(2025)Xue, Tan, Ma and Ukkusuri}]{xue2025data}
\bibinfo{author}{Xue, J.}, \bibinfo{author}{Tan, R.}, \bibinfo{author}{Ma, J.}, \bibinfo{author}{Ukkusuri, S.V.}, \bibinfo{year}{2025}.
\newblock \bibinfo{title}{Data mining in transportation networks with graph neural networks: A review and outlook}.
\newblock \bibinfo{journal}{arXiv preprint arXiv:2501.16656} .
%Type = Article
\bibitem[{Yao et~al.(2025)Yao, Chen, Sun, Chen, Zhang, Pan, Li and Ding}]{yao2025group}
\bibinfo{author}{Yao, C.}, \bibinfo{author}{Chen, Y.}, \bibinfo{author}{Sun, Y.}, \bibinfo{author}{Chen, Y.}, \bibinfo{author}{Zhang, W.}, \bibinfo{author}{Pan, X.}, \bibinfo{author}{Li, Y.}, \bibinfo{author}{Ding, B.}, \bibinfo{year}{2025}.
\newblock \bibinfo{title}{Group-relative reinforce is secretly an off-policy algorithm: Demystifying some myths about grpo and its friends}.
\newblock \bibinfo{journal}{arXiv preprint arXiv:2509.24203} .
%Type = Article
\bibitem[{Yao et~al.(2023)Yao, Yu, Zhao, Shafran, Griffiths, Cao and Narasimhan}]{yao2023tree}
\bibinfo{author}{Yao, S.}, \bibinfo{author}{Yu, D.}, \bibinfo{author}{Zhao, J.}, \bibinfo{author}{Shafran, I.}, \bibinfo{author}{Griffiths, T.}, \bibinfo{author}{Cao, Y.}, \bibinfo{author}{Narasimhan, K.}, \bibinfo{year}{2023}.
\newblock \bibinfo{title}{Tree of thoughts: Deliberate problem solving with large language models}.
\newblock \bibinfo{journal}{Advances in neural information processing systems} \bibinfo{volume}{36}, \bibinfo{pages}{11809--11822}.
%Type = Article
\bibitem[{Yap and Cats(2021)}]{yap2021predicting}
\bibinfo{author}{Yap, M.}, \bibinfo{author}{Cats, O.}, \bibinfo{year}{2021}.
\newblock \bibinfo{title}{Predicting disruptions and their passenger delay impacts for public transport stops}.
\newblock \bibinfo{journal}{Transportation} \bibinfo{volume}{48}, \bibinfo{pages}{1703--1731}.
%Type = Article
\bibitem[{Yu et~al.(2025)Yu, Zhang, Zhu, Yuan, Zuo, Yue, Dai, Fan, Liu, Liu et~al.}]{yu2025dapo}
\bibinfo{author}{Yu, Q.}, \bibinfo{author}{Zhang, Z.}, \bibinfo{author}{Zhu, R.}, \bibinfo{author}{Yuan, Y.}, \bibinfo{author}{Zuo, X.}, \bibinfo{author}{Yue, Y.}, \bibinfo{author}{Dai, W.}, \bibinfo{author}{Fan, T.}, \bibinfo{author}{Liu, G.}, \bibinfo{author}{Liu, L.}, et~al., \bibinfo{year}{2025}.
\newblock \bibinfo{title}{Dapo: An open-source llm reinforcement learning system at scale}.
\newblock \bibinfo{journal}{arXiv preprint arXiv:2503.14476} .
%Type = Inproceedings
\bibitem[{Yuan et~al.(2024)Yuan, Ding, Feng, Jin and Li}]{yuan2024unist}
\bibinfo{author}{Yuan, Y.}, \bibinfo{author}{Ding, J.}, \bibinfo{author}{Feng, J.}, \bibinfo{author}{Jin, D.}, \bibinfo{author}{Li, Y.}, \bibinfo{year}{2024}.
\newblock \bibinfo{title}{Unist: A prompt-empowered universal model for urban spatio-temporal prediction}, in: \bibinfo{booktitle}{Proceedings of the 30th ACM SIGKDD Conference on Knowledge Discovery and Data Mining}, pp. \bibinfo{pages}{4095--4106}.
%Type = Article
\bibitem[{Yue et~al.(2025)Yue, Chen, Lu, Zhao, Wang, Song and Huang}]{yue2025does}
\bibinfo{author}{Yue, Y.}, \bibinfo{author}{Chen, Z.}, \bibinfo{author}{Lu, R.}, \bibinfo{author}{Zhao, A.}, \bibinfo{author}{Wang, Z.}, \bibinfo{author}{Song, S.}, \bibinfo{author}{Huang, G.}, \bibinfo{year}{2025}.
\newblock \bibinfo{title}{Does reinforcement learning really incentivize reasoning capacity in llms beyond the base model?}
\newblock \bibinfo{journal}{arXiv preprint arXiv:2504.13837} .
%Type = Article
\bibitem[{Zhang et~al.(2024)Zhang, Fu, Liang, Zhang, Yu, Cai and Yao}]{zhang2024trafficgpt}
\bibinfo{author}{Zhang, S.}, \bibinfo{author}{Fu, D.}, \bibinfo{author}{Liang, W.}, \bibinfo{author}{Zhang, Z.}, \bibinfo{author}{Yu, B.}, \bibinfo{author}{Cai, P.}, \bibinfo{author}{Yao, B.}, \bibinfo{year}{2024}.
\newblock \bibinfo{title}{Trafficgpt: Viewing, processing and interacting with traffic foundation models}.
\newblock \bibinfo{journal}{Transport Policy} \bibinfo{volume}{150}, \bibinfo{pages}{95--105}.
%Type = Article
\bibitem[{Zhao et~al.(2024)Zhao, Ma, Peng and Cheng}]{zhao2024predicting}
\bibinfo{author}{Zhao, Y.}, \bibinfo{author}{Ma, Z.}, \bibinfo{author}{Peng, H.}, \bibinfo{author}{Cheng, Z.}, \bibinfo{year}{2024}.
\newblock \bibinfo{title}{Predicting metro incident duration using structured data and unstructured text logs}.
\newblock \bibinfo{journal}{Transportmetrica A: Transport Science} , \bibinfo{pages}{1--29}.
%Type = Article
\bibitem[{Zheng et~al.(2025)Zheng, Liu, Li, Chen, Yu, Gao, Dang, Liu, Men, Yang et~al.}]{zheng2025group}
\bibinfo{author}{Zheng, C.}, \bibinfo{author}{Liu, S.}, \bibinfo{author}{Li, M.}, \bibinfo{author}{Chen, X.H.}, \bibinfo{author}{Yu, B.}, \bibinfo{author}{Gao, C.}, \bibinfo{author}{Dang, K.}, \bibinfo{author}{Liu, Y.}, \bibinfo{author}{Men, R.}, \bibinfo{author}{Yang, A.}, et~al., \bibinfo{year}{2025}.
\newblock \bibinfo{title}{Group sequence policy optimization}.
\newblock \bibinfo{journal}{arXiv preprint arXiv:2507.18071} .
%Type = Article
\bibitem[{Ziegler et~al.(2019)Ziegler, Stiennon, Wu, Brown, Radford, Amodei, Christiano and Irving}]{ziegler2019fine}
\bibinfo{author}{Ziegler, D.M.}, \bibinfo{author}{Stiennon, N.}, \bibinfo{author}{Wu, J.}, \bibinfo{author}{Brown, T.B.}, \bibinfo{author}{Radford, A.}, \bibinfo{author}{Amodei, D.}, \bibinfo{author}{Christiano, P.}, \bibinfo{author}{Irving, G.}, \bibinfo{year}{2019}.
\newblock \bibinfo{title}{Fine-tuning language models from human preferences}.
\newblock \bibinfo{journal}{arXiv preprint arXiv:1909.08593} .

\end{thebibliography}

% \clearpage
% \section*{Appendix}
% \input{appendix}

\end{document}